\documentclass[letterpaper]{article} % DO NOT CHANGE THIS
\usepackage{aaai2026}  % DO NOT CHANGE THIS
\usepackage{times}  % DO NOT CHANGE THIS
\usepackage{helvet}  % DO NOT CHANGE THIS
\usepackage{courier}  % DO NOT CHANGE THIS
\usepackage[hyphens]{url}  % DO NOT CHANGE THIS
\usepackage{graphicx} % DO NOT CHANGE THIS
\usepackage{natbib}  % DO NOT CHANGE THIS AND DO NOT ADD ANY OPTIONS TO IT
\usepackage{caption} % DO NOT CHANGE THIS AND DO NOT ADD ANY OPTIONS TO IT
\usepackage{algorithm}
\usepackage{algorithmic}

\usepackage[utf8]{inputenc} 
\usepackage[table]{xcolor}
\usepackage{multirow}
\usepackage{adjustbox}
\usepackage{enumitem}
\usepackage{array}
\usepackage{amsthm}
\usepackage{amsmath}
\usepackage{amssymb}
\usepackage{subfigure}
\usepackage{marvosym}
\usepackage{comment}
\usepackage{colortbl}
\usepackage{hhline}
\usepackage{xspace} 
\usepackage{dblfloatfix}
\usepackage{diagbox}
\usepackage{booktabs}
\usepackage{nicefrac}
\usepackage{microtype}

\makeatletter
\DeclareRobustCommand\onedot{\futurelet\@let@token\@onedot}
\def\@onedot{\ifx\@let@token.\else.\null\fi\xspace}
\def\eg{e.g\onedot} 
\def\ie{i.e\onedot} 
 
\def\wrt{w.r.t\onedot}

\newcommand{\@toptitlebar}{
  \hrule height 4\p@
  \vskip 0.25in
}
\newcommand{\@bottomtitlebar}{
  \vskip 0.29in
  \hrule height 1\p@
  \vskip 0.09in%
}

\newcommand{\bestResults}{%
Best results are \textbf{bolded}, second best \underline{underlined}.}

\newcommand{\ILITableCaptionAppendix}[2]{%
ILI rate forecasting results in #1~(#2) with web search frequency time series as exogenous variables. We assess performance on $4$ test seasons (2015/16 to 2018/19) using linear correlation ($r$), MAE, and sMAPE ($\epsilon\%$). \bestResults}

\newcommand{\ILIFigCaptionAppendix}[3]{%
#1-day ahead forecasts for all influenza seasons and models, #2~(#3).}

\newcommand{\WEATableCaptionAppendix}[2]{%
WEA forecasting results (temperature at $850$ hPa) in #1~(#2). We assess performance on $3$ test periods (years 2016, '17, '18) using linear correlation ($r$), MAE, and sMAPE ($\epsilon\%$). \bestResults}

\DeclareUrlCommand\url{\color{black}\def\UrlLeft{}}

\usepackage{newfloat}
\usepackage{listings}
\DeclareCaptionStyle{ruled}{labelfont=normalfont,labelsep=colon,strut=off} % DO NOT CHANGE THIS
\floatstyle{ruled}
\newfloat{listing}{tb}{lst}{}
\floatname{listing}{Listing}

\title{Sonnet: Spectral Operator Neural Network for \\ Multivariable Time Series Forecasting}
\author{
    Yuxuan Shu, Vasileios Lampos
}
\affiliations{
    Centre for Artificial Intelligence \\
    Department of Computer Science \\
    University College London\\
    \{yuxuan.shu.22, v.lampos\}@ucl.ac.uk
}

\usepackage{bibentry}
\nocopyright

\begin{document}

\maketitle

\begin{abstract}
  Multivariable time series forecasting methods can integrate information from exogenous variables, leading to significant prediction accuracy gains. The transformer architecture has been widely applied in various time series forecasting models due to its ability to capture long-range sequential dependencies. However, a na\"ive application of transformers often struggles to effectively model complex relationships among variables over time. To mitigate against this, we propose a novel architecture, termed \textbf{S}pectral \textbf{O}perator \textbf{N}eural \textbf{Net}work (\textbf{Sonnet}). Sonnet applies learnable wavelet transformations to the input and incorporates spectral analysis using the Koopman operator. Its predictive skill relies on the \textbf{M}ulti\textbf{v}ariable \textbf{C}oherence \textbf{A}ttention (\textbf{MVCA}), an operation that leverages spectral coherence to model variable dependencies. Our empirical analysis shows that Sonnet yields the best performance on $34$ out of $47$ forecasting tasks with an average mean absolute error (MAE) reduction of $2.2\%$ against the most competitive baseline. We further show that MVCA can remedy the deficiencies of na\"ive attention in various deep learning models, reducing MAE by $10.7\%$ on average in the most challenging forecasting tasks.
  \footnote{Accepted for oral presentation at AAAI 2026. This is an extended version of the conference paper.}

\end{abstract}

\begin{links}
    \link{Code \& data}{https://github.com/ClaudiaShu/Sonnet}
\end{links}

\section{Introduction}
\label{sec:introduction}

Multivariable time series (MTS) forecasting methods learn from multiple input variables to predict a single target variable~\citep{hidalgo2013multivariate}. MTS models are deployed in many real-life applications, such as financial modelling~\citep{pniemira2004analytic,antunes2018forecasting,wtaylor2003using,fwitt1995forecasting}, numerical weather prediction~\citep{fildes2011validation,cyoung2018data,dalton2022exogenous}, and computational epidemiology~\citep{freyerdugas2013influenza,shaman2012forecasting,silva2020forecasting,morris2023neural}. Recent machine learning models for time series forecasting tend to emphasise multivariate formulations~\citep{nie2023time,zeng2023are,liu2023koopa,ansari2024chronos,das2024decoder}. While such models provide important insights for integrating traditional time series forecasting into deep learning techniques, \eg capturing seasonality~\citep{lin2024cyclenet}, frequency-domain methods~\citep{zhou2022fedformer,zhou2022film}, and autoregression~\citep{zeng2023are,nie2023time}, they often overlook the benefits of leveraging external information through exogenous indicators.

In fact, empirical evidence suggests that capturing dependencies from external variables can increase the risk of overfitting for complex models~\citep{nie2023time}. Moreover, existing methods that capture inter-variable dependencies~\citep{zhang2023crossformer,wang2024timemixer} have, in certain forecasting tasks, been outperformed by models that do not~\citep{lin2024cyclenet,luo2024deformabletst}. However, this can be attributed to the choice of forecasting benchmarks, which, in many occasions, hold strong seasonal patterns~(\citealt{lin2024cyclenet}; see also Appendix~\ref{subsec:when_endo}), that can be effectively modelled using simpler autoregressive methods~\citep{zeng2023are,lin2024cyclenet}. Additionally, models that do not capture inter-variable dependencies benefit from learning using longer look-back windows without significantly increasing modelling complexity and computational cost~\citep{han2023capacity}, making them less likely to overfit. Nevertheless, these empirical outcomes are not necessarily valid for multivariable forecasting scenarios, especially for tasks where exogenous variables have strong predictive power. In such cases, explicitly modelling interactions across variables facilitates better performance~\citep{wang2024timexer,shu2025deformtime}.

Prior work has explored various ways in capturing inter-variable dependencies for multivariate or MTS forecasting tasks. Methods applying attention~\citep{vaswani2017attention} across variables~\citep{liu2024itransformer,ilbert2024samformer} can capture nonlinear dependencies, but disrupt temporal information by embedding sequences along the time dimension. Crossformer~\citep{zhang2023crossformer} attempts to rectify this by proposing a modified transformer structure where input is split into time series patches allowing to capture dependencies across both the time and variable dimensions at the sub-sequence level. ModernTCN~\citep{luo2024moderntcn}, on the other hand, uses a convolutional kernel over both the variable and time dimension to jointly capture the inter- and intra-variable dependencies. However, both Crossformer and ModernTCN suffer from GPU computational overhead as the number of exogenous variables or the length of the input time series increases~\citep{shu2025deformtime,zhou2024efficanet}. TimeXer~\citep{wang2024timexer} models dependencies between exogenous variables and the target variable separately, and then joins their learned embeddings with a cross-attention module. DeformTime~\citep{shu2025deformtime} yields superior results by using deformable attention to incorporate information from exogenous variables at different time steps. However, it only captures inter-variable dependencies within the reception field of a convolutional kernel, which limits the exploration of a wider range of exogenous variables. 

Frequency-based analysis techniques, including those using Fourier~\citep{bochner1953fourier,sorensen1987realvalued} or wavelet transform~\citep{varanini1997spectral,zhang2003detection,arneodo1988wavelet,farge1992wavelet,yu2024adawavenet}, has been widely used in statistical approaches for identifying periodic patterns~\citep{priestley2018spectral} as well as in machine learning methods for forecasting~\citep{lange2021fourier,zhou2022fedformer,zhou2022film,piao2024fredformer}. These methods compress temporal information (due to the Fourier transform) while some of them focus on capturing intra-variable dependencies~\citep{zhou2022fedformer,zhou2022film}. Wavelet transform, albeit dependent on the chosen mother wavelet~\citep{demoortel2004wavelet,ngui2013wavelet}, can maintain both time and frequency information. In the frequency domain, spectral coherence~\citep{white1990cross} serves as a powerful tool for capturing the correlation between variables at different frequencies~\citep{stein1972frequency,mima1999corticomuscular}. To improve input projections into the frequency domain,~\citet{lange2021fourier} used a Koopman operator~\citep{mezic2005spectral,rowley2009spectral,avila2020datadriven,liu2023koopa}, \ie a spectral function that enables linear modelling of nonlinear changes to better address nonlinearities. More recently, AdaWaveNet~\citep{yu2024adawavenet} decomposed the input series into seasonal and long-term components, and used the wavelet transform to capture periodic information. This method accounts for inter-variable dependencies only over the seasonal components. 

Motivated by the aforementioned remarks, we propose \textbf{S}pectral \textbf{O}perator \textbf{N}eural \textbf{Net}work (Sonnet), a model that captures MTS dependencies in the spectral domain using a learnable wavelet transform. Sonnet captures both intra- and inter-variable dependencies using a novel frequency-domain \textbf{M}ulti\textbf{v}ariable \textbf{C}oherence \textbf{A}ttention (MVCA) layer. Furthermore, it deploys a learnable Koopman operator for linearised transitions of temporal states~\citep{li2020learning}. MVCA demonstrates stand-alone effectiveness when integrated into existing architectures, outperforming na\"ive attention and other modified attention mechanisms. Our key contributions are:
\begin{enumerate}[leftmargin=*, topsep=-2pt, itemsep=-1.6pt]
    \item We propose \textbf{Sonnet}, a novel neural network architecture for MTS forecasting that captures inter-variable dependencies via adaptable time-frequency spectral operators while enforcing stability through learnable Koopman dynamics.
    \item We introduce \textbf{MVCA}, an attention mechanism designed to model interactions between variables by leveraging their spectral coherence, a frequency-domain measurement of dependency. Unlike conventional self-attention, which computes pairwise similarity via dot products, MVCA captures temporal relationships through their cross-spectral density with the inclusion of frequency information from all variables, enhancing variable dependency modelling for MTS tasks.
    \item We assess forecasting accuracy on carefully curated MTS data sets, including established benchmarks complemented by tasks on weather forecasting, influenza prevalence, electricity consumption, and energy prices. The weather prediction and influenza prevalence tasks support a more thorough evaluation as they include a substantial amount of exogenous variables, multiple years ($\geq 10$) for training, and multiple test periods ($\geq 3$).
    \item Sonnet reduces mean absolute error (MAE) by $2.2\%$ on average compared to the most performant baseline model (stat. sig., $p < 10^{-3}$). In the more challenging tasks of influenza and weather modelling, MAE is reduced by $3.5\%$ and $2\%$, respectively. Performance gains persist as the forecasting horizon increases, demonstrating the effectiveness of Sonnet in longer-term forecasting. 
\end{enumerate}

\section{MTS Forecasting Task Definition}
\label{sec:task_definition}
We focus exclusively on multi\underline{variable} time series (MTS) forecasting, whereby multiple input variables are used to predict a single target variable. We note that although for some baseline models multiple output variables may be present (multi\underline{variate} forecasting), our evaluation is restricted to the prediction of one specific output. All models are trained and evaluated under a rolling window setup, with a look-back window and a forecasting horizon of $L$ and $H$ time steps, respectively. At each time step $t$, the $C$ observed exogenous variables over $L$ past time steps, $\{t-L+1, \dots, t\}$, are captured in an input matrix $\mathbf{X}_{t} \in \mathbb{R}^{L \times C}$. The autoregressive signal for the target (endogenous) variable is denoted by $\mathbf{y}_{t-\delta} \in \mathbb{R}^{L}$; this encompasses time steps $\{t-\delta-L+1, \dots, t-\delta\}$, where $\delta \in \mathbb{N}_{0}$ is an optional delay applied when the target variable is observed with a temporal lag. We capture the exogenous and endogenous input variables in $ \mathbf{Z}_{t} = [\mathbf{X}_{t}, \mathbf{y}_{t-\delta}] \in \mathbb{R}^{L \times (C + 1)}$. The goal is to predict the target variable at time step $t+H$, where $H$ denotes the forecasting horizon. Hence, the output of a forecasting model is denoted by $\mathbf{y}_{t+H} \in \mathbb{R}^H$ and holds forecasts for time steps $\{t+1, \dots, t+H-1, t+H\}$. For models that conduct multivariate forecasting, the output includes predictions for all covariates and hence is denoted by $\mathbf{Y}_{t+H} = [\mathbf{X}_{t+H}, \mathbf{y}_{t+H}] \in \mathbb{R}^{H \times (C+1)}$. The forecasting task is to learn $f: \mathbf{Z}_t \rightarrow \mathbf{y}_{t+H}$ or $\mathbf{Y}_{t+H}$. Performance is measured based on the endogenous forecast at time step $t+H$, \ie the last (temporally) element of $\mathbf{y}_{t+H}$, $y_{t+H} \in \mathbb{R}$. For notational simplicity, we omit temporal subscripts and use $\mathbf{Z}$ for $\mathbf{Z}_t$, $\mathbf{y}$ for $\mathbf{y}_{t-\delta}$, and $\mathbf{X}$ for $\mathbf{X}_t$.

\section{Spectral Coherence with Sonnet}
\label{sec:method}
In this section, we provide a detailed description of our proposed model, Sonnet. It operates in the spectral domain via a learnable wavelet transform and introduces a \textbf{M}ulti\textbf{v}ariable \textbf{C}oherence \textbf{A}ttention (MVCA) module to capture both inter- and intra-variable dependencies. We further use a Koopman projection layer that enables stable temporal evolution via a learned linear operator.

\subsection{Joint Embedding of Input Variables}
\label{subsec:joint_embedding}
Given an input matrix $\mathbf{Z} \in \mathbb{R}^{L \times (C+1)}$ that consists of the exogenous variables $\mathbf{X} \in \mathbb{R}^{L \times C}$ and endogenous variable $\mathbf{y} \in \mathbb{R}^{L}$, we first obtain the embeddings of $\mathbf{X}$ and $\mathbf{y}$ independently. These are denoted by $\mathbf{E}_{x} \in \mathbb{R}^{L \times \alpha d}$ and $\mathbf{E}_{y} \in \mathbb{R}^{L \times \left( 1-\alpha \right)d}$. They are derived using learnable weight matrices $\mathbf{W}_{x} \in \mathbb{R}^{C \times \alpha d}$ and $\mathbf{W}_{y} \in \mathbb{R}^{1 \times \left( 1-\alpha \right)d}$, as in $\mathbf{E}_{x} = \mathbf{X} \mathbf{W}_{x}$, where $d$ is the embedding dimension and $\alpha \in [0,1]$ a hyperparameter that controls the projected dimensionality of $\mathbf{X}$ and $\mathbf{y}$ in the final embedding.\footnote{We make sure that both products $\alpha d$ and $(1-\alpha)d \in \mathbb{N}$ to avoid dimension mismatch caused by rounding. For $\alpha = 0$, the forecasting is based on the historical values of the target variable only, collapsing to an autoregressive setting. For $\alpha = 1$, forecasting depends entirely on the exogenous variables.} By concatenating along the feature dimension, we obtain the final embedding $\mathbf{E} = [\mathbf{E}_{x}, \mathbf{E}_{y}] \in \mathbb{R}^{L \times d}$.

\subsection{Learnable Wavelet Transform}
\label{subsec:wavelet_transform}
We then transform the time series embedding into the wavelet space that contains both time and frequency information, to capture both fine-grained details and overall trends across various temporal resolutions~\citep{mallat1989theory,daubechies1990wavelet}.
Specifically, after obtaining the input time series embedding $\mathbf{E}$, we first define a set of $K$ learnable wavelet transformations (also referred to as atoms), with the $k$-th atom held in a matrix $\mathbf{M}_k \in \mathbb{R}^{d \times L}$ derived by 
\begin{equation}
    \mathbf{M}_k = \exp \left( -\mathbf{w}_{\alpha} \mathbf{t}^2 \right) \times \cos \left( \mathbf{w}_{\beta}\mathbf{t} + \mathbf{w}_{\gamma} \mathbf{t}^2 \right) \, \, ,
\end{equation}
where $\mathbf{t} \in \mathbb{R}^{L}$ is a row vector capturing normalised time steps with $\mathbf{t}_i = i / (L - 1)$ for $i = 0, \dots, L - 1$, and $\mathbf{w}_{\alpha}$, $\mathbf{w}_{\beta}$, and $\mathbf{w}_{\gamma} \in \mathbb{R}^{d}$ are learnable weight vectors that control the shape of the wavelet, each initialised randomly from a normal distribution. In particular, $\mathbf{w}_{\alpha}$ controls the width of the Gaussian envelope, and $\mathbf{w}_{\beta}$, $\mathbf{w}_{\gamma}$ respectively determine the linear or quadratic frequency modulation of the generated cosine waveforms. This formulation enables the atoms to adapt to localised time-frequency structures in the data. The time series embedding $\mathbf{E} \in \mathbb{R}^{L \times d}$ is then transformed into the wavelet space by projecting it onto each of the wavelet atoms $\mathbf{M}_k \in \mathbb{R}^{d \times L}$ using $\mathbf{P}_k = \mathbf{E} \odot \mathbf{M}_k^{\top}$, where $\mathbf{P}_k \in \mathbb{R}^{L \times d}$ denotes the embedding's projection for the $k$-th atom, and $\odot$ is element-wise multiplication. The transformed wavelet across all $K$ atoms is denoted by $\mathbf{P} \in \mathbb{R}^{K \times L \times d}$. The aforementioned steps help to preserve temporal structure while decomposing the input into multi-resolution time-frequency components using adaptive wavelet transforms that can capture both short- and long-term patterns in the data.

\subsection{Multivariable Coherence Attention (MVCA)}
\label{subsec:MVCA}
In MTS forecasting, input variables can be both auto-correlated (to their own past values) and cross-correlated (to each other). To improve learning from these correlations, we propose MVCA, a module that supplements the standard attention. MVCA can capture inter- and intra-variable dependencies within the frequency domain using spectral density coherence. Its premise is that variables with a higher spectral coherence should contribute more to the attention output. MVCA can be used in place of any na\"ive attention module used in a forecasting method.

Given the obtained input embedding matrix in wavelet space $\mathbf{P} \in \mathbb{R}^{K \times L \times d}$, where $L$ and $d$ are the time and variable dimension, $K$ is the number of wavelets after transformation, we first linearly project it to query, key and value embeddings, denoted as $\mathbf{Q}$, $\mathbf{K}$, and $\mathbf{V} \in \mathbb{R}^{K \times L \times d}$, using weight matrices $\mathbf{W}_{q}$, $\mathbf{W}_{k}$, and $\mathbf{W}_{v} \in \mathbb{R}^{d \times d}$ respectively, as in $\mathbf{Q} = \mathbf{P}\mathbf{W}_{q}$. We then consider the embeddings from different wavelet transformations (indexed by $k$) as separate attention heads. Specifically, for each attention head, we obtain the sub-tensor of $\mathbf{Q}$, $\mathbf{K}$, and $\mathbf{V}$, denoted as $\mathbf{Q}_h$, $\mathbf{K}_h$, and $\mathbf{V}_h \in \mathbb{R}^{L \times d}$, as the query, key, and value embeddings. Therefore, each head captures a distinct subspace of the original embedding, where subspaces correspond to different wavelet transformations of the input. Including a multi-head structure enables the model to learn diverse dependencies in parallel~\citep{vaswani2017attention}. 

We then apply a Fast Fourier Transform (FFT) along the variable dimension~\citep{dudgeon1984multidimensional} of each transformer head to transfer the query and key embeddings to the frequency domain, \ie $\mathbf{Q}_f = \mathrm{FFT}\left(\mathbf{Q}_h\right)$ and $\mathbf{K}_f = \mathrm{FFT}\left(\mathbf{K}_h\right)$, with both $\mathbf{Q}_f$ and $\mathbf{K}_f \in \mathbb{C}^{L \times \ell}$, where $\ell = \lfloor \frac{d}{2} \rfloor + 1$. Each frequency bin in the transformed data ($\mathbf{Q}_f$, $\mathbf{K}_f$) contains information from all original inputs to the FFT~\citep{lee-thorp2022fnet}, capturing both fine-grained (high-frequency) and global (low-frequency) patterns. The cross-spectral density $\mathbf{P}_{qk}\,\in\,\mathbb{C}^{L \times \ell}$ and power-spectral densities $\mathbf{P}_{qq}$,  $\mathbf{P}_{kk}\,\in\,\mathbb{R}^{L \times \ell}$ are obtained as follows:
\begin{equation}
\label{eq:coherence_density}
    \mathbf{P}_{qk} = \mathbf{Q}_{f} \odot \mathbf{K}_{f}^\ast \, , \, \, \,
    \mathbf{P}_{qq} = \mathbf{Q}_{f} \odot \mathbf{Q}_{f}^\ast \, , \, \, \,
    \mathbf{P}_{kk} = \mathbf{K}_{f} \odot \mathbf{K}_{f}^\ast \, , 
\end{equation}
where `$^\ast$' denotes the complex conjugate. We average along the second dimension to obtain $\overline{\mathbf{P}}_{qk} \in \mathbb{C}^{L}$, $\overline{\mathbf{P}}_{qq}$, and $\overline{\mathbf{P}}_{kk} \in \mathbb{R}^{L}$ (see also Appendix~\ref{appsec:further_details}). The normalised spectral coherence $\mathbf{C}_{qk} \in \mathbb{R}^{L}$ is then computed using
\begin{equation}
\label{eq:coherence}
    \mathbf{C}_{qk} = |\overline{\mathbf{P}}_{qk}|^2 / \left(\overline{\mathbf{P}}_{qq} \cdot \overline{\mathbf{P}}_{kk} + \epsilon\right),
\end{equation}
where $\epsilon = 10^{-6}$ mitigates division by $0$.
$\mathbf{C}_{qk}$ captures the linear dependency between sequences across frequency bands. Therefore, the coherence here assigns higher importance to the time steps where the query and key hold more similar averaged values across multiple frequencies.

Following the common design of attention layers~\citep{vaswani2017attention}, we scale, normalise, and regularise $\mathbf{C}_{qk}$, \ie $\mathbf{A}_h = \mathrm{Dropout}(\mathrm{Softmax}(\mathbf{C}_{qk}/\sqrt{d}))$. The attention weights $\mathbf{A}_h \in \mathbb{R}^{L}$ of each head are first broadcast along the feature dimension to form $\mathbf{A} \in \mathbb{R}^{L \times d}$, which is then multiplied with the value representations $\mathbf{V}_h$ (element-wise) to produce head-specific outputs, $\mathbf{O}_h \in \mathbb{R}^{L \times d} = \mathbf{A} \odot \mathbf{V}_h$. We concatenate these outputs across all heads to obtain $\mathbf{O}_r \in \mathbb{R}^{K \times L \times d}$. We then use a $2$-layer perceptron ($\mathrm{MLP}$, $d$-dimensional layers) with Gaussian Error Linear Unit (GELU) activation to further capture nonlinearities. This is connected to a residual layer to form the output $\mathbf{O}_m \in \mathbb{R}^{K \times L \times d} = \mathbf{O}_r + \mathrm{MLP}(\mathbf{O}_r)$. We then multiply $\mathbf{O}_m$ with a weight matrix $\mathbf{W}_{\text{out}} \in \mathbb{R}^{d \times d}$ to obtain the output of MVCA, $\mathbf{O} \in \mathbb{R}^{K \times L \times d} = \mathbf{O}_m \mathbf{W}_{\text{out}}$.

\subsection{Koopman-Guided Spectrum Evolvement}
\label{subsec:koopman}
Motivated by Koopman operator theory~\citep{mezic2005spectral,rowley2009spectral,avila2020datadriven}, which offers a framework for modelling nonlinear dynamics, we introduce a layer to capture the temporal evolution of time-frequency patterns in wavelet space. We aim to learn a Koopman operator with $K$ dimensions in the transformed space. To obtain the operator $\mathbf{K} \in \mathbb{C}^{K \times K}$, we first initialise a learnable complex-valued matrix $\mathbf{S} \in \mathbb{C}^{K \times K}$. At each forward pass, we apply QR decomposition to $\mathbf{S}$, and retain only the resulting unitary matrix $\mathbf{U} \in \mathbb{C}^{K \times K}$, \ie $\mathbf{U} \leftarrow \mathrm{QR} \left( \mathbf{S} \right)$ s.t. $\mathbf{U}^\dagger \mathbf{U} = \mathbf{I}$, where $\mathbf{U}^{\dagger}$ is the conjugate transpose of $\mathbf{U}$. Multiplying with $\mathbf{U}$ therefore prevents data amplification or distortion. 

We then initialise a learnable vector $\mathbf{p} \in \mathbb{R}^K$, where the $k$-th element, $p_k$, controls the temporal evolution (phase transformation angle) of the $k$-th transformation. All elements in the vector are then mapped into complex numbers, obtaining $\mathbf{v} \in \mathbb{C}^K$, where $v_k = e^{i p_k}$. We use $\mathbf{v}$ to form a diagonal matrix $\mathbf{D} \in \mathbb{C}^{K \times K} = \mathrm{diag}(\mathbf{v})$. The Koopman operator, $\mathbf{K}$, is then given by $\mathbf{K} = \mathbf{U} \mathbf{D} \mathbf{U}^\dagger$. We use $\mathbf{K}$ to model the temporal evolution of the MVCA embedding, $\mathbf{O}$, after converting it to a complex form, $\mathbf{O}_c \in \mathbb{C}^{K \times L \times d}$ (the imaginary part is set to $i0$ to perform complex-valued transformations without altering the original embedding), given by $\mathbf{O}_l = \mathbf{K} \times \mathbf{O}_c$, where $\mathbf{O}_l \in \mathbb{C}^{K \times L \times d}$ is the evolved embedding after multiplication with $\mathbf{K}$. A conventional Koopman framework models temporal evolution recursively at each time step~\citep{Lusch2018deep,avila2020datadriven}. We instead choose to apply Koopman transformation in the frequency domain with one forward pass, which is equivalent to learning a direct transformation from the input to the output. This global projection reduces the accumulation of sequential errors while maintaining robustness in training.

\subsection{Sequence Reconstruction from Wavelets}
\label{subsec:reconstruction}
The inverse transformation reconstructs the original sequence by aggregating weights from each wavelet atom. Given the evolved state $\mathbf{O}_l$, we first obtain its real part denoted as $\mathbf{O}_r \in \mathbb{R}^{K \times L \times d}$. For each wavelet atom indexed by $k$, let $\mathbf{O}_k \in \mathbb{R}^{L \times d}$ denote the corresponding slice of $\mathbf{O}_r$. We then multiply it with the wavelet atom $\mathbf{M}_k \in \mathbb{R}^{d \times L}$, \ie $\mathbf{R}_k = \mathbf{O}_k \odot \mathbf{M}_k^{\top}$. $\mathbf{R}_k \in \mathbb{R}^{L \times d}$ denotes the series reconstruction from the $k$-th atom. This operation is conducted over all $K$ atoms. Joining all the reconstructed series forms a $K \times L \times d$ matrix. We sum over its first dimension to obtain the reconstructed embedding, denoted as $\mathbf{R} \in \mathbb{R}^{L \times d}$.

\subsection{Convolutional Decoder}
\label{subsec:decoder}
Finally, a $3$-layer convolutional decoder transforms the learned representation. It comprises $3$ $1$-dimensional convolutional layers with GELU activations between every $2$ layers. Each layer uses kernel sizes $[5, 3, 3]$ and paddings $[2, 1, 1]$ respectively, followed by an adaptive average pooling layer at the end. The dimension of each layer is $[H \times 4, \, H \times 2, \, H]$, producing the final sequence representation $\mathbf{Z}_{\text{out}} \in \mathbb{R}^{H \times H}$, in accordance with the time steps of the target forecasting horizon. The result is then linearly projected using a weight vector $\mathbf{W}_z \in \mathbb{R}^{H}$ to generate the final output, as in $\hat{\mathbf{y}} \in \mathbb{R}^{H} = \mathbf{Z}_{\text{out}}\mathbf{W}_z$, in accordance with the dimensionality of the target variable.

\section{Results}
\label{sec:results}
We assess forecasting accuracy using an expanded collection of data sets and tasks, to overcome potential biases present in the current literature. We first compare Sonnet against other competitive baseline models. We then investigate the role of attention mechanisms in time series forecasting models by removing or replacing na\"ive transformers with more advanced variants, including the proposed MVCA module. We also provide an ablation study of the key components of Sonnet and seed control in Appendix~\ref{appsec:supp_results}.

\begin{table*}[!t]
\centering
\fontsize{9}{10}\selectfont
% \small
\setlength{\tabcolsep}{0.003\linewidth}
\setlength{\aboverulesep}{0.5pt}
\setlength{\belowrulesep}{0.5pt}
% \resizebox{\linewidth}{!}{%
\begin{tabular}{cccccccccccccccccccc}\toprule
\multicolumn{2}{c}{\textbf{Task}} & \multicolumn{2}{c}{\textbf{Sonnet}} & \multicolumn{2}{c}{\textbf{DeformTime}} & \multicolumn{2}{c}{\textbf{ModernTCN}} & \multicolumn{2}{c}{\textbf{Samformer}} & \multicolumn{2}{c}{\textbf{TimeXer}} & \multicolumn{2}{c}{\textbf{PatchTST}} & \multicolumn{2}{c}{\textbf{iTransformer}} & \multicolumn{2}{c}{\textbf{Crossformer}} & \multicolumn{2}{c}{\textbf{DLinear}} \\
 & $H$ & MAE & $\epsilon\%$ & MAE & $\epsilon\%$ & MAE & $\epsilon\%$ & MAE & $\epsilon\%$ & MAE & $\epsilon\%$ & MAE & $\epsilon\%$ & MAE & $\epsilon\%$ & MAE & $\epsilon\%$ & MAE & $\epsilon\%$ \\
\midrule
\multirow{3}{*}{\rotatebox{90}{ELEC}} 
& 12 & \textbf{0.1040} & \textbf{24.95} & \underline{0.1162} & \underline{26.68} & 0.1596 & \cellcolor{gray!20}{36.81} & \cellcolor{gray!20}{0.2336} & \cellcolor{gray!20}{53.84} & 0.1287 & 28.45 & 0.1419 & 30.76 & 0.1468 & 31.91 & 0.1513 & 30.35 & \cellcolor{gray!20}{0.3307} & \cellcolor{gray!20}{71.51} \\
& 24 & \textbf{0.1203} & \textbf{27.70} & \underline{0.1359} & \underline{30.11} & \cellcolor{gray!20}{0.1806} & \cellcolor{gray!20}{39.70} & \cellcolor{gray!20}{0.2520} & \cellcolor{gray!20}{56.39} & 0.1586 & 33.20 & 0.1545 & 34.04 & 0.1588 & \cellcolor{gray!20}{35.40} & \cellcolor{gray!20}{0.1873} & \cellcolor{gray!20}{35.60} & \cellcolor{gray!20}{0.3369} & \cellcolor{gray!20}{73.23} \\
& 36 & \textbf{0.1389} & \textbf{30.21} & \cellcolor{gray!20}{0.1729} & \cellcolor{gray!20}{35.81} & \cellcolor{gray!20}{0.2065} & \cellcolor{gray!20}{43.79} & \cellcolor{gray!20}{0.2866} & \cellcolor{gray!20}{62.40} & \cellcolor{gray!20}{0.1785} & \cellcolor{gray!20}{35.33} & \underline{0.1659} & \underline{33.85} & \cellcolor{gray!20}{0.1791} & \cellcolor{gray!20}{38.18} & \cellcolor{gray!20}{0.2273} & \cellcolor{gray!20}{39.36} & \cellcolor{gray!20}{0.3908} & \cellcolor{gray!20}{82.38} \\
\midrule
\multirow{4}{*}{\rotatebox{90}{ENER}} 
& 24 & \textbf{0.3621} & \cellcolor{gray!20}{\textbf{65.67}} & \cellcolor{gray!20}{0.3717} & \cellcolor{gray!20}{66.92} & \cellcolor{gray!20}{0.3752} & \cellcolor{gray!20}{66.53} & \cellcolor{gray!20}{\underline{0.3703}} & \cellcolor{gray!20}{66.01} & \cellcolor{gray!20}{0.3733} & \cellcolor{gray!20}{69.59} & \cellcolor{gray!20}{0.3717} & \cellcolor{gray!20}{\underline{65.89}} & \cellcolor{gray!20}{0.3838} & \cellcolor{gray!20}{68.36} & \cellcolor{gray!20}{0.3716} & \cellcolor{gray!20}{67.44} & \cellcolor{gray!20}{0.4307} & \cellcolor{gray!20}{76.49} \\
& 48 & \textbf{0.4120} & \textbf{71.07} & 0.4299 & 74.59 & \underline{0.4154} & \underline{71.74} & 0.4294 & 72.34 & 0.4404 & \cellcolor{gray!20}{77.61} & 0.4467 & 75.60 & 0.4386 & 75.63 & 0.4647 & \cellcolor{gray!20}{81.79} & \cellcolor{gray!20}{0.5138} & \cellcolor{gray!20}{89.79} \\
& 72 & \textbf{0.4036} & \textbf{70.52} & 0.4217 & 74.50 & \underline{0.4089} & \underline{70.84} & 0.4353 & 73.40 & 0.4246 & 74.70 & 0.4303 & 73.90 & 0.4276 & 74.19 & 0.4473 & \cellcolor{gray!20}{78.62} & \cellcolor{gray!20}{0.5124} & \cellcolor{gray!20}{89.26} \\
& 168 & \textbf{0.3980} & \textbf{68.63} & 0.4399 & \cellcolor{gray!20}{78.98} & \cellcolor{gray!20}{0.4497} & \cellcolor{gray!20}{74.86} & 0.4243 & \underline{72.28} & \cellcolor{gray!20}{0.4493} & \cellcolor{gray!20}{76.76} & \cellcolor{gray!20}{0.4872} & \cellcolor{gray!20}{81.34} & 0.4352 & \cellcolor{gray!20}{74.11} & \underline{0.4168} & \cellcolor{gray!20}{76.76} & \cellcolor{gray!20}{0.5070} & \cellcolor{gray!20}{87.87} \\
\midrule
\multirow{4}{*}{\rotatebox{90}{ETTh1}} 
& 96 & 0.2145 & 16.01 & \textbf{0.1941} & \textbf{14.96} & 0.2047 & 15.66 & 0.2047 & 15.73 & 0.2135 & 16.03 & \underline{0.2017} & \underline{15.41} & 0.2052 & 15.46 & 0.2126 & 16.52 & \cellcolor{gray!20}{0.2599} & \cellcolor{gray!20}{20.82} \\
& 192 & 0.2330 & 17.61 & \textbf{0.2116} & \textbf{16.08} & 0.2417 & 18.32 & \underline{0.2307} & 17.64 & 0.2322 & \underline{16.96} & 0.2409 & 18.29 & 0.2429 & 18.13 & \cellcolor{gray!20}{0.2820} & \cellcolor{gray!20}{21.63} & \cellcolor{gray!20}{0.3798} & \cellcolor{gray!20}{31.78} \\
& 336 & \underline{0.2392} & 18.91 & \textbf{0.2158} & \textbf{16.27} & 0.2415 & 18.52 & 0.2523 & 18.94 & 0.2414 & \underline{17.34} & 0.2559 & 19.29 & 0.2593 & 19.11 & 0.2947 & 22.65 & \cellcolor{gray!20}{0.6328} & \cellcolor{gray!20}{58.34} \\
& 720 & \underline{0.2768} & \underline{19.73} & 0.2862 & 21.81 & 0.2785 & 20.44 & 0.3026 & 23.21 & \textbf{0.2617} & \textbf{18.58} & 0.3087 & 23.89 & 0.2886 & 22.05 & \cellcolor{gray!20}{0.3350} & 24.84 & \cellcolor{gray!20}{0.7563} & \cellcolor{gray!20}{69.52} \\
\midrule
\multirow{4}{*}{\rotatebox{90}{ETTh2}} 
& 96 & \textbf{0.3098} & \textbf{33.04} & \underline{0.3121} & 40.07 & 0.3199 & 40.68 & 0.3312 & 40.16 & 0.3346 & 41.04 & 0.3145 & \underline{39.25} & 0.3420 & 42.41 & 0.3486 & 40.71 & 0.3349 & 41.68 \\
& 192 & \underline{0.3742} & \underline{39.58} & \textbf{0.3281} & \textbf{37.90} & 0.3887 & 47.08 & 0.3874 & 45.37 & 0.4154 & 47.07 & 0.3839 & 45.45 & 0.4233 & 47.44 & 0.4035 & 43.16 & 0.4084 & \cellcolor{gray!20}{50.67} \\
& 336 & \underline{0.3689} & \underline{39.48} & \textbf{0.3450} & \textbf{37.00} & 0.3904 & 50.54 & 0.4083 & 46.41 & 0.4041 & 42.26 & 0.4018 & 46.77 & 0.4332 & 45.95 & 0.4487 & 49.44 & 0.4710 & \cellcolor{gray!20}{55.53} \\
& 720 & \underline{0.4335} & 51.62 & \textbf{0.3640} & \textbf{34.99} & \cellcolor{gray!20}{0.5728} & \cellcolor{gray!20}{63.04} & 0.5198 & 58.44 & 0.5135 & 56.17 & 0.4960 & 55.27 & 0.4565 & \underline{45.40} & \cellcolor{gray!20}{0.5832} & \cellcolor{gray!20}{61.45} & \cellcolor{gray!20}{0.7981} & \cellcolor{gray!20}{94.67} \\
\midrule
\multirow{4}{*}{\rotatebox{90}{ILI-ENG}} 
& 7 & \textbf{1.4791} & \textbf{21.84} & \underline{1.6417} & \cellcolor{gray!20}{28.60} & 1.9489 & \cellcolor{gray!20}{28.27} & \cellcolor{gray!20}{2.3475} & \cellcolor{gray!20}{28.31} & \cellcolor{gray!20}{2.8084} & \cellcolor{gray!20}{33.66} & \cellcolor{gray!20}{2.3115} & \cellcolor{gray!20}{27.61} & \cellcolor{gray!20}{2.3084} & \cellcolor{gray!20}{26.38} & 1.8698 & \cellcolor{gray!20}{\underline{25.70}} & \cellcolor{gray!20}{2.8214} & \cellcolor{gray!20}{43.02} \\
& 14 & \textbf{1.9225} & \textbf{25.77} & \underline{2.2308} & 33.98 & 2.7050 & \cellcolor{gray!20}{36.01} & 3.0290 & \cellcolor{gray!20}{36.63} & \cellcolor{gray!20}{3.4937} & \cellcolor{gray!20}{41.88} & \cellcolor{gray!20}{3.2547} & \cellcolor{gray!20}{37.76} & \cellcolor{gray!20}{3.2301} & \cellcolor{gray!20}{36.67} & 2.6543 & \underline{30.97} & \cellcolor{gray!20}{3.7922} & \cellcolor{gray!20}{55.29} \\
& 21 & \textbf{2.5101} & \underline{36.53} & \underline{2.6500} & \textbf{32.70} & 3.0400 & 40.02 & \cellcolor{gray!20}{4.4980} & \cellcolor{gray!20}{54.41} & \cellcolor{gray!20}{4.3337} & \cellcolor{gray!20}{51.57} & \cellcolor{gray!20}{4.3192} & \cellcolor{gray!20}{51.11} & \cellcolor{gray!20}{4.2347} & \cellcolor{gray!20}{48.93} & 3.0014 & 40.57 & \cellcolor{gray!20}{4.4739} & \cellcolor{gray!20}{61.25} \\
& 28 & \underline{2.7481} & \textbf{36.95} & \textbf{2.7228} & \underline{40.44} & 3.3611 & 47.87 & \cellcolor{gray!20}{5.1598} & \cellcolor{gray!20}{60.78} & \cellcolor{gray!20}{4.9013} & \cellcolor{gray!20}{61.60} & \cellcolor{gray!20}{4.9964} & \cellcolor{gray!20}{59.60} & \cellcolor{gray!20}{4.8125} & \cellcolor{gray!20}{55.35} & 3.1983 & 46.15 & \cellcolor{gray!20}{5.0347} & \cellcolor{gray!20}{67.75} \\
\midrule
\multirow{4}{*}{\rotatebox{90}{ILI-US2}} 
& 7 & \textbf{0.3806} & \textbf{14.89} & \underline{0.4122} & \underline{16.01} & 0.4398 & 16.55 & \cellcolor{gray!20}{0.6495} & \cellcolor{gray!20}{24.21} & 0.6083 & \cellcolor{gray!20}{23.38} & \cellcolor{gray!20}{0.7097} & \cellcolor{gray!20}{24.51} & \cellcolor{gray!20}{0.6507} & \cellcolor{gray!20}{23.24} & 0.4400 & 16.46 & \cellcolor{gray!20}{0.7355} & \cellcolor{gray!20}{27.94} \\
& 14 & \textbf{0.4491} & \underline{18.38} & \underline{0.4752} & \textbf{17.73} & 0.5279 & 20.22 & 0.7696 & \cellcolor{gray!20}{30.16} & 0.7725 & \cellcolor{gray!20}{29.07} & \cellcolor{gray!20}{0.8635} & \cellcolor{gray!20}{30.11} & 0.7896 & 28.17 & 0.5852 & 20.98 & \cellcolor{gray!20}{0.8435} & \cellcolor{gray!20}{32.22} \\
& 21 & \textbf{0.5326} & \textbf{20.64} & \underline{0.5425} & \underline{22.12} & 0.5781 & 23.85 & 0.8374 & 31.42 & 0.8243 & 31.46 & \cellcolor{gray!20}{1.0286} & \cellcolor{gray!20}{36.70} & 0.8042 & 30.03 & 0.6245 & 22.29 & 0.9124 & \cellcolor{gray!20}{34.93} \\
& 28 & 0.5788 & \textbf{21.15} & \textbf{0.5538} & \underline{22.25} & \underline{0.5710} & 23.66 & 0.9389 & 36.80 & 0.9074 & 34.72 & \cellcolor{gray!20}{1.1525} & \cellcolor{gray!20}{42.61} & 0.9619 & 36.75 & 0.6512 & 23.47 & 0.9999 & 38.46 \\
\midrule
\multirow{4}{*}{\rotatebox{90}{ILI-US9}} 
& 7 & \underline{0.2668} & \underline{12.98} & \textbf{0.2622} & \textbf{12.26} & 0.2899 & 14.17 & 0.4025 & \cellcolor{gray!20}{19.39} & 0.3813 & 18.21 & \cellcolor{gray!20}{0.4116} & \cellcolor{gray!20}{19.34} & \cellcolor{gray!20}{0.4057} & \cellcolor{gray!20}{18.57} & 0.3149 & 14.44 & \cellcolor{gray!20}{0.4675} & \cellcolor{gray!20}{23.47} \\
& 14 & \textbf{0.2806} & \textbf{13.10} & \underline{0.3084} & \underline{13.80} & 0.3417 & 15.29 & \cellcolor{gray!20}{0.5257} & \cellcolor{gray!20}{24.50} & 0.4665 & 22.14 & \cellcolor{gray!20}{0.5020} & \cellcolor{gray!20}{24.09} & 0.4702 & 22.44 & 0.3571 & 17.23 & \cellcolor{gray!20}{0.5467} & \cellcolor{gray!20}{27.35} \\
& 21 & \textbf{0.3179} & \textbf{14.11} & \underline{0.3179} & \underline{14.23} & 0.3710 & 15.43 & 0.5415 & 24.22 & 0.5715 & \cellcolor{gray!20}{27.43} & \cellcolor{gray!20}{0.5935} & \cellcolor{gray!20}{29.40} & 0.5106 & 24.11 & 0.3418 & 15.90 & \cellcolor{gray!20}{0.6001} & \cellcolor{gray!20}{29.66} \\
& 28 & \underline{0.3675} & 16.53 & \textbf{0.3532} & \textbf{15.75} & 0.3940 & 17.19 & 0.6050 & 27.95 & 0.6555 & 31.32 & 0.6665 & \cellcolor{gray!20}{33.35} & 0.6498 & 31.05 & 0.3747 & \underline{16.44} & 0.6564 & \cellcolor{gray!20}{32.16} \\
\midrule
\multirow{4}{*}{\rotatebox{90}{WEA-CT}} 
& 4 & \textbf{1.6240} & \underline{9.63} & 1.7600 & 10.41 & 1.8752 & 11.05 & 2.2265 & 13.09 & 2.2376 & 13.14 & \cellcolor{gray!20}{3.5004} & \cellcolor{gray!20}{20.17} & 2.1906 & 12.83 & \underline{1.6382} & \textbf{9.62} & 3.1483 & 18.22 \\
& 12 & \textbf{3.5432} & \textbf{20.38} & \underline{3.5681} & \underline{20.46} & 3.7761 & 21.67 & 4.0444 & 23.14 & 3.7241 & 21.31 & 4.1910 & 23.99 & 3.9741 & 22.83 & 3.5932 & 20.62 & 4.0265 & 22.94 \\
& 28 & \textbf{3.7277} & \textbf{21.32} & \underline{3.7601} & \underline{21.49} & 3.9399 & 22.36 & 3.9239 & 22.37 & 3.8325 & 21.87 & 3.9405 & 22.48 & 3.9584 & 22.60 & 3.8061 & 21.72 & 3.9254 & 22.37 \\
& 120 & \textbf{3.7373} & \textbf{21.35} & 3.8040 & 21.67 & 4.0889 & 23.54 & 3.9018 & 22.30 & 3.8412 & 21.88 & 4.0547 & 23.02 & 3.9473 & 22.49 & \underline{3.7659} & \underline{21.49} & 4.0570 & 23.05 \\
\midrule
\multirow{4}{*}{\rotatebox{90}{WEA-HK}} 
& 4 & \textbf{0.6389} & \textbf{4.05} & 0.6804 & 4.32 & 0.7004 & 4.42 & 0.8097 & 5.10 & 0.8648 & 5.52 & \cellcolor{gray!20}{1.1488} & \cellcolor{gray!20}{7.15} & 0.8048 & 5.08 & \underline{0.6488} & \underline{4.11} & \cellcolor{gray!20}{0.9898} & \cellcolor{gray!20}{6.21} \\
& 12 & \textbf{1.2355} & \textbf{7.74} & \underline{1.2786} & \underline{7.99} & 1.3555 & 8.43 & 1.4006 & 8.70 & 1.3027 & 8.13 & 1.5825 & 9.77 & 1.4225 & 8.86 & 1.2896 & 8.08 & 1.5464 & 9.62 \\
& 28 & \textbf{1.4135} & \textbf{8.83} & \underline{1.4746} & \underline{9.16} & 1.5866 & 9.82 & 1.6226 & 10.05 & 1.5115 & 9.39 & 1.6441 & 10.16 & 1.6356 & 10.08 & 1.5282 & 9.45 & 1.6232 & 10.06 \\
& 120 & \underline{1.5469} & \underline{9.58} & \textbf{1.5399} & \textbf{9.45} & 1.6326 & 10.22 & 1.8843 & 12.06 & 1.7092 & 10.55 & 2.0084 & 12.84 & 1.6745 & 10.45 & 1.5931 & 9.73 & 1.8329 & 11.49 \\
\midrule
\multirow{4}{*}{\rotatebox{90}{WEA-LD}} 
& 4 & \textbf{1.7231} & \textbf{15.16} & 1.8753 & 16.31 & 1.9456 & 16.88 & 2.1537 & 18.53 & 2.2628 & 19.25 & \cellcolor{gray!20}{2.7602} & \cellcolor{gray!20}{22.93} & 2.1509 & 18.55 & \underline{1.7447} & \underline{15.26} & 2.5065 & 20.94 \\
& 12 & \textbf{2.9589} & \textbf{23.88} & \underline{3.0214} & \underline{24.15} & 3.2056 & 25.58 & 3.3070 & 26.61 & 3.1625 & 25.25 & 3.5406 & 28.33 & 3.3622 & 27.36 & 3.0492 & 24.35 & 3.3927 & 26.70 \\
& 28 & \textbf{3.2161} & \textbf{25.49} & \underline{3.2724} & \underline{25.84} & 3.5067 & 27.48 & 3.5841 & 27.97 & 3.4672 & 27.23 & 3.7365 & 29.42 & 3.6884 & 29.14 & 3.3048 & 25.88 & 3.6073 & 28.16 \\
& 120 & \textbf{3.2464} & \textbf{25.82} & \underline{3.2973} & \underline{26.17} & 3.8434 & 29.92 & 3.8420 & 30.32 & 3.6557 & 28.52 & 4.2344 & 32.40 & 3.8518 & 30.36 & 3.3935 & 26.75 & 3.9640 & 30.35 \\
\midrule
\multirow{4}{*}{\rotatebox{90}{WEA-NY}} 
& 4 & \textbf{1.2716} & \textbf{11.94} & 1.4028 & 13.03 & 1.4154 & 12.95 & 1.6003 & 14.44 & 1.7290 & 15.57 & \cellcolor{gray!20}{2.1644} & \cellcolor{gray!20}{19.24} & 1.6066 & 14.40 & \underline{1.2935} & \underline{12.24} & \cellcolor{gray!20}{1.9782} & \cellcolor{gray!20}{17.88} \\
& 12 & \underline{2.4476} & 21.23 & \textbf{2.4453} & \textbf{21.04} & 2.6221 & 22.93 & 2.7069 & 23.57 & 2.6537 & 22.85 & 2.8592 & 24.81 & 2.6609 & 23.09 & 2.4494 & \underline{21.11} & 2.9507 & 25.29 \\
& 28 & \textbf{2.6744} & \textbf{23.10} & \underline{2.7450} & \underline{23.20} & 2.9336 & 25.37 & 3.0347 & 26.18 & 2.8775 & 24.48 & 3.0956 & 27.01 & 3.0204 & 25.30 & 2.7830 & 23.66 & 3.2099 & 27.07 \\
& 120 & \textbf{2.7135} & \textbf{23.15} & \underline{2.8224} & \underline{23.86} & 3.3000 & 28.05 & 3.5289 & 33.01 & 3.1501 & 26.55 & 3.4086 & 28.78 & 3.1029 & 27.74 & 2.9615 & 24.69 & 3.6129 & 29.75 \\
\midrule
\multirow{4}{*}{\rotatebox{90}{WEA-SG}} 
& 4 & \textbf{0.3444} & \textbf{1.25} & 0.3557 & 1.30 & 0.3624 & 1.32 & 0.3925 & 1.43 & 0.3801 & 1.38 & 0.4238 & 1.54 & 0.3868 & 1.41 & \underline{0.3532} & \underline{1.29} & 0.4048 & 1.47 \\
& 12 & \textbf{0.4160} & \textbf{1.51} & \underline{0.4256} & \underline{1.55} & 0.4493 & 1.64 & 0.4784 & 1.74 & 0.4447 & 1.62 & 0.4992 & 1.82 & 0.4662 & 1.70 & 0.4359 & 1.59 & 0.4764 & 1.73 \\
& 28 & \textbf{0.4653} & \textbf{1.69} & \underline{0.4875} & \underline{1.77} & 0.5196 & 1.89 & 0.5421 & 1.97 & 0.4974 & 1.81 & 0.5542 & 2.02 & 0.5307 & 1.93 & 0.5003 & 1.82 & 0.5284 & 1.92 \\
& 120 & \textbf{0.4830} & \textbf{1.76} & \underline{0.5050} & \underline{1.83} & 0.5321 & 1.94 & 0.5221 & 1.90 & 0.5215 & 1.90 & 0.5357 & 1.95 & 0.5233 & 1.90 & 0.5212 & 1.89 & 0.5328 & 1.94 \\
\bottomrule
\end{tabular}
% }
\caption{Forecasting accuracy results across all tasks, methods, and forecasting horizons ($H$). For the ELEC/ILI/WEA tasks, we report the average performance across $2$, $4$, and $3$ test sets, respectively (detailed breakdowns in Appendix E.1).
% ~\ref{appsubsec:full_results}). 
$\epsilon\%$ denotes sMAPE. \bestResults{} Grey background denotes the model does not outperform persistence.}
\label{tab:results_averaged}
\end{table*}

\subsection{Experiment Settings}
\label{subsec:experiment_settings}
We conduct experiments over $12$ real-world data sets. This includes $2$ established benchmarks from prior papers~\citep{zhou2021informer,zeng2023are}, specifically the ETTh1 and ETTh2 data sets, which contain hourly electricity transformer temperature forecasting. Oil temperature is our target variable, with the remaining indicators considered as exogenous variables, following~\citet{wang2024timexer}. We also use $2$ data sets from the Darts library~\citep{herzen2022darts}, predicting hourly energy prices (ENER) and electricity (low-voltage) consumption (ELEC). In addition, we form weather data sets extracted from the WeatherBench repository~\citep{rasp2020weatherbench} for $5$ cities from diverse geographical locations, namely London (WEA-LD), New York (WEA-NY), Hong Kong (WEA-HK), Cape Town (WEA-CT), and Singapore (WEA-SG), to support a more inclusive analysis. For each city, we sample data from its nearest grid point and obtain $5$ climate indicators. We provide spatial context by including data from its eight surrounding grid points ($3 \times 3$ grid) as additional exogenous variables. Following prior work on global climate forecasting~\citep{verma2024climode}, we resample the data to a temporal resolution of $6$ hours. The forecasting target is the $850$ hPa (T850) temperature, a key indicator for climate modelling~\citep{scherrer2004analysis,hamill2007ensemble}. Finally, we include influenza-like illness (ILI) rate forecasting tasks (as in~\citep{shu2025deformtime}) in $3$ locations, England (ILI-ENG), and in U.S. Health \& Human Services (HHS) Regions 2 (ILI-US2) and 9 (ILI-US9). In the ILI tasks, frequency time series of web searches are included as exogenous predictors. More information about the data sets is provided in Appendix~\ref{appsec:datasets}.

For the ETT tasks, we set the forecasting horizon ($H$) to $\{96,192,336,720\}$ time steps, and use a single test set of consecutive unseen instances adopting the evaluation settings in prior work~\citep{nie2023time,liu2024itransformer}. For ENER we set $H=\{24,48,72,168\}$ hours ahead and use $1$ year (2018) for testing. For ELEC, we use the last $2$ years (2020, '21) as $2$ distinct test seasons, and set $H$ to $\{12,24,36\}$ hours. For the WEA tasks, we form $3$ test sets (years 2016, '17, '18), and set $H$ to $\{4,12,28,120\}$ time steps corresponding to $\{1, 3, 7, 30\}$ days. Following the same setup as in~\citep{shu2025deformtime}, for the ILI forecasting task, we test models on $4$ consecutive influenza seasons (2015/16 to 2018/19), each time training a new model on data from previous seasons. We set $H=\{7,14,21,28\}$ days (more details in Appendix~\ref{appsec:supplementary_setting}).

We compare Sonnet to $8$ competitive forecasting models that, to the best of our knowledge, form the current SOTA methods: DLinear~\citep{zeng2023are}, Crossformer~\citep{zhang2023crossformer}, iTransformer~\citep{liu2024itransformer}, PatchTST~\citep{nie2023time}, TimeXer~\citep{wang2024timexer}, Samformer~\citep{ilbert2024samformer}, ModernTCN~\citep{luo2024moderntcn}, and DeformTime~\citep{shu2025deformtime}, with TimeXer and DeformTime being designed for multivariable forecasting. We also include a na\"ive baseline which can either be a seasonal or a standard persistence model, depending on the presence of strong seasonality. 
For all tasks (except ETT), we conduct hyperparameter tuning for all models. For the ETT tasks, we adopt settings from the official repositories of methods (except for Crossformer, which did not provide this). Appendix~C has further details about the baseline models.

\begin{table*}[!t]
    \renewcommand{\arraystretch}{0.9}
    \centering
    \setlength{\tabcolsep}{5.5pt}
    \setlength{\aboverulesep}{-0.2pt}
    \setlength{\belowrulesep}{1.2pt}
    \small
    % \resizebox{\textwidth}{!}{%
    \begin{tabular}{llcccccccccccc}
      \toprule
\multirow{2}{*}{} & \multirow{2}{*}{\bf Attention} & \multicolumn{3}{c}{$H=7$ days}  & \multicolumn{3}{c}{$H=14$ days}  & \multicolumn{3}{c}{$H=21$ days}  & \multicolumn{3}{c}{$H=28$ days}  \\
 &  & \bf ENG & \bf US2 & \bf US9 & \bf ENG & \bf US2 & \bf US9 & \bf ENG & \bf US2 & \bf US9 & \bf ENG & \bf US2 & \bf US9 \\
% \midrule
\cmidrule(lr){2-5} \cmidrule(lr){6-8} \cmidrule(lr){9-11} \cmidrule(lr){12-14}
\multirow{6}{*}
{\rotatebox[origin=c]{90}{iTransformer}} & --- & 2.3084 & \underline{0.6507} & \underline{0.4057} & \underline{3.2301} & \underline{0.7896} & \underline{0.4702} & 4.2347 & \underline{0.8042} & 0.5106 & \underline{4.8125} & 0.9619 & 0.6498 \\
 & $\neg$ Attn & \underline{2.3011} & 0.7242 & 0.4475 & 3.2557 & 0.8696 & 0.5344 & 4.4729 & 1.0105 & 0.6207 & 5.1864 & 1.1667 & 0.7179 \\
 & FNet & 2.5534 & 0.7665 & 0.4741 & 3.5337 & 0.8746 & 0.5550 & 4.6970 & 1.0006 & 0.6066 & 5.3153 & 1.1118 & 0.7139 \\
 & FED & 3.6150 & 0.9286 & 0.5613 & 4.3098 & 1.0750 & 0.6527 & 5.8596 & 1.4140 & 0.8669 & 6.4196 & 1.5167 & 0.9397 \\
 & VDAB & 2.4287 & 0.6772 & 0.4126 & 3.3093 & 0.8274 & 0.4982 & \textbf{3.9406} & 0.8485 & \underline{0.4992} & \textbf{4.4975} & \underline{0.9068} & \underline{0.5790} \\
 & MVCA & \textbf{2.2707} & \textbf{0.5483} & \textbf{0.3581} & \textbf{3.1233} & \textbf{0.7267} & \textbf{0.4454} & \underline{4.1780} & \textbf{0.7578} & \textbf{0.4880} & 4.8705 & \textbf{0.8885} & \textbf{0.5409} \\
\cmidrule{1-14}
\multirow{6}{*}{\rotatebox[origin=c]{90}{SamFormer}} & --- & 2.3475 & 0.6495 & 0.4025 & \underline{3.0290} & 0.7696 & 0.5257 & 4.4980 & 0.8374 & 0.5415 & 5.1598 & 0.9389 & 0.6050 \\
 & $\neg$ Attn & 2.5010 & 0.8245 & 0.5131 & 3.3015 & 0.9679 & 0.6047 & 4.1830 & 1.1396 & 0.7389 & 5.0136 & 1.2391 & 0.8347 \\
 & FNet & 2.4812 & 0.7765 & 0.4699 & 3.2054 & 0.9202 & 0.5865 & 4.3909 & 0.9266 & 0.6443 & 5.0592 & 1.0983 & 0.7149 \\
 & FED & 3.5217 & 0.9140 & 0.5599 & 4.2109 & 1.0519 & 0.6508 & 5.7800 & 1.3439 & 0.8519 & 5.9809 & 1.4348 & 0.9190 \\
 & VDAB & \textbf{2.0406} & \underline{0.5755} & \underline{0.3651} & \textbf{2.9525} & \underline{0.7281} & \underline{0.4505} & \underline{4.1080} & \underline{0.7883} & \textbf{0.4861} & \underline{4.9015} & \underline{0.8617} & \underline{0.5605} \\
 & MVCA & \underline{2.1365} & \textbf{0.5514} & \textbf{0.3609} & 3.2046 & \textbf{0.6935} & \textbf{0.4379} & \textbf{3.8033} & \textbf{0.7294} & \underline{0.4975} & \textbf{4.7438} & \textbf{0.8159} & \textbf{0.5345} \\
\cmidrule{1-14}
\multirow{6}{*}{\rotatebox[origin=c]{90}{PatchTST}} & --- & 2.3115 & 0.7097 & 0.4116 & 3.2547 & 0.8635 & 0.5020 & 4.3192 & 1.0286 & 0.5935 & 4.9964 & 1.1525 & 0.6665 \\
 & $\neg$ Attn & 2.4723 & 0.7702 & 0.4585 & 3.6415 & 0.9017 & 0.5551 & 4.2998 & 1.0657 & 0.6472 & 4.8538 & 1.1704 & 0.7313 \\
 & FNet & 2.8097 & 0.8033 & 0.4591 & 4.0342 & 0.9322 & 0.5327 & 4.7975 & 1.0341 & 0.5430 & 4.9718 & 1.1087 & 0.6279 \\
 & FED & 3.6158 & 0.9292 & 0.5614 & 4.6572 & 1.0765 & 0.6542 & 5.8725 & 1.4043 & 0.8671 & 6.3728 & 1.4953 & 0.9306 \\
 & VDAB & \underline{2.0799} & \underline{0.5925} & \underline{0.3820} & \textbf{3.0211} & \underline{0.7722} & \underline{0.4413} & \textbf{3.8164} & \underline{0.8009} & \underline{0.5159} & \underline{4.6044} & \textbf{0.8518} & \underline{0.5801} \\
 & MVCA & \textbf{2.0054} & \textbf{0.5824} & \textbf{0.3705} & \underline{3.0411} & \textbf{0.7406} & \textbf{0.4291} & \underline{3.8627} & \textbf{0.7871} & \textbf{0.4746} & \textbf{4.5712} & \underline{0.8765} & \textbf{0.5491} \\
      \bottomrule
    \end{tabular}
    % }%
\caption{Performance (average MAE across $4$ test seasons) of iTransformer, Samformer, and PatchTST on the ILI forecasting tasks (ILI-ENG/US2/US9) with different modifications to the na\"ive attention mechanism. `$\neg$ Attn' denotes the removal of the residual attention module, and FNet / FED / VDAB refer to using the attention modules proposed in FNet~\shortcite{shu2025deformtime}, FEDformer~\shortcite{zhou2022fedformer}, and DeformTime~\shortcite{shu2025deformtime}, respectively. \bestResults{}}
\label{tab:ILI-results-vca}
\end{table*}

\subsection{Forecasting Accuracy of Sonnet}
\label{subsec:forecasting_results_Sonnet}
Prediction accuracy for all tasks is enumerated in Table~\ref{tab:results_averaged}, using Mean Absolute Error (MAE) and symmetric Mean Absolute Percentage Error (sMAPE or $\epsilon\%$) as the main evaluation metrics. For WEA tasks, sMAPE is modified to avoid small values due to Kelvin temperature units (see Appendix~\ref{appsubsec:special_smape}). For ETT and ENER tasks, results are based on a single forecasting season (\ie, one train / test split per forecasting horizon), while we average over multiple seasons for other tasks.

Sonnet exhibits the overall best performance across the explored MTS forecasting tasks. Comparing Sonnet's performance to the best-performing baseline model for each task and forecasting horizon yields an MAE reduction of $1.1\%$ on average. Ranking-wise, Sonnet is the best-performing model on $34$, and the second-best on $9$ out of $47$ forecasting tasks. Sonnet shows relatively inferior performance on the ETT tasks. However, the ETT data sets have only $6$ exogenous variables and cover a limited training time span ($2$ years). Consequently, they offer a limited spectrum to gain from exploring variable dependencies, as well as insufficient historical context, both of which are elements that Sonnet leverages from. Aside from the ETT tasks, Sonnet reduces MAE by $3.3\%$ on average. Therefore, Sonnet is more effective with more covariates and longer time spans for training. Compared to a specific forecasting model, Sonnet consistently improves accuracy, reducing MAE from $2.2\%$ for the best-performing baseline (DeformTime) to $31.1\%$ for the worst-performing one (DLinear). We note that the MAE performance gain over DeformTime is statistically significant based on a paired t-test across all the $47$ forecasting tasks ($p=5e-4$). This further highlights the model's forecasting capacity amongst a wide range of applications. 

Focusing on the more challenging forecasting tasks (ILI and WEA), Sonnet outperforms baselines $25$ or $26$ (out of $32$ tasks in total) based on sMAPE or MAE, respectively. The MAE reduction is $3.5\%$ on average for ILI tasks and $2\%$ for WEA tasks. Within these tasks, we also observe consistent comparative performance patterns in the baseline models. Models that do not capture inter-variable dependencies (DLinear and PatchTST) generally offer lower predictability (Sonnet reduces their MAE by $28.8\%$ and $29.8\%$), and in most ILI tasks, they cannot surpass the performance of a persistence model. In contrast, baseline models that capture inter-variable dependencies tend to perform better. Among these, those that embed the covariates along the temporal dimension, namely Samformer, TimeXer, and iTransformer, exhibit inferior performance (Sonnet reduces their MAE by $23.9\%$, $22.1\%$, and $22.9\%$), whereas models that preserve temporal order, namely DeformTime, ModernTCN, and Crossformer, achieve stronger results (Sonnet reduces their MAE by $3.4\%$, $11.1\%$, and $8\%$). Hence, in tasks with more informative covariates, preserving temporal structure while modelling inter-variable dependencies is a desirable property. We provide ablation study results in Appendix~\ref{subsec:ablation} that quantify the contribution of different modules to the overall forecasting accuracy of Sonnet. Additional performance evaluation results over the entire output sequence are provided in Appendix~\ref{appsubsec:over_sequence}.

\subsection{Effectiveness of Different Attention Modules in MTS Forecasting}
\label{subsec:forecasting_results_vca}
We evaluate the effectiveness of the proposed attention-driven module, MVCA, by integrating it into existing forecasting models (we refer to them as base models) that originally deploy na\"ive transformer attention. Experiments are conducted on the ILI tasks as these are the most representative of the MTS class we are exploring: they not only contain many exogenous predictors, but also frame hard practical epidemiological modelling problems when $H = 21$ or $28$ days ahead. We use Samformer~\citep{ilbert2024samformer}, iTransformer~\citep{liu2024itransformer}, and PatchTST~\citep{nie2023time} as base models. For each one of them, we evaluate $5$ attention configurations: removing the attention module altogether ($\neg$ Attn), attention with Fourier transformer proposed by FNet~\citep{lee-thorp2022fnet}, the Frequency Enhanced Decomposed (FED) attention proposed in FEDformer~\citep{zhou2022fedformer}, the variable deformable attention (VDAB) proposed by DeformTime~\citep{shu2025deformtime}, and ours (MVCA). Further details are provided in Appendices~\ref{appsubsec:select_attn}~and~\ref{appsubsec:setup_mvca}.

Table~\ref{tab:ILI-results-vca} enumerates MAEs averaged across the $4$ test seasons for all forecasting horizons (sMAPEs in Appendix~\ref{appsubsec:mvca_smape}, Table~\ref{tab:ILI-results-vca-smape}). Notably, for the base models, removing the attention module does not lead to significant performance degradation, with some cases even showing improved results, \eg improved accuracy is reported for some horizons for the ILI-ENG task. This indicates that the attention modules in the base models do not always capture useful information. 

\begin{table}[!t]
\centering\fontsize{9}{10}\selectfont
% \small
\setlength{\tabcolsep}{1.6pt}
\setlength{\aboverulesep}{-0.5pt}
\setlength{\belowrulesep}{1.2pt}
% \resizebox{\linewidth}{!}{%
\begin{tabular}{cccccccccccc}
\toprule
\multicolumn{2}{c}{} & \multicolumn{2}{c}{\textbf{Sonnet}} & \multicolumn{2}{c}{\textbf{VDAB}
% ~\shortcite{shu2025deformtime}
} & \multicolumn{2}{c}{\textbf{FED}
% ~\shortcite{zhou2022fedformer}
} & \multicolumn{2}{c}{\textbf{FNet}
% ~\shortcite{lee-thorp2022fnet}
} & \multicolumn{2}{c}{\textbf{Na\"ive}} \\

% \multicolumn{2}{c}{} & \multicolumn{2}{c}{} & \multicolumn{2}{c}{\shortcite{shu2025deformtime}} & \multicolumn{2}{c}{\citet{zhou2022fedformer}} & \multicolumn{2}{c}{\citet{lee-thorp2022fnet}} &  \\

 & $H$ & MAE & $\epsilon\%$ & MAE & $\epsilon\%$ & MAE & $\epsilon\%$ & MAE & $\epsilon\%$ & MAE & $\epsilon\%$ \\
\midrule
\multirow{4}{*}{\rotatebox{90}{ILI-ENG}} 
& 7 & \underline{1.479} & \underline{21.8} & \textbf{1.433} & \textbf{20.3} & 5.991 & 77.7 & 1.591 & 27.7 & 1.717 & 25.3 \\
         
& 14 & \textbf{1.923} & \textbf{25.8} & \underline{1.973} & \underline{29.4} & 6.031 & 78.1 & 2.084 & 32.8 & 2.068 & 31.4 \\
         
& 21 & \textbf{2.510} & \underline{36.5} & 2.774 & \textbf{35.2} & 6.045 & 78.2 & \underline{2.765} & 39.3 & 2.781 & 37.2 \\
         
& 28 & \textbf{2.748} & \textbf{37.0} & \underline{3.060} & \underline{44.3} & 6.051 & 78.4 & 3.240 & 44.5 & 3.078 & 50.0 \\
\midrule
\multirow{4}{*}{\rotatebox{90}{ILI-US2}} 
& 7 & \textbf{0.381} & \textbf{14.9} & 0.429 & \underline{16.6} & 1.168 & 45.0 & \underline{0.415} & 18.5 & 0.436 & 17.5 \\
         
& 14 & \textbf{0.449} & \textbf{18.4} & \underline{0.500} & \underline{19.1} & 1.179 & 45.5 & 0.558 & 23.2 & 0.532 & 20.4 \\
         
& 21 & \textbf{0.533} & \underline{20.6} & \underline{0.575} & \textbf{20.5} & 1.195 & 46.2 & 0.659 & 25.3 & 0.613 & 23.5 \\
         
& 28 & \textbf{0.579} & \textbf{21.1} & \underline{0.643} & \underline{24.3} & 1.193 & 46.1 & 0.705 & 29.1 & 0.651 & 26.1 \\
\midrule
\multirow{4}{*}{\rotatebox{90}{ILI-US9}} 
& 7 & \textbf{0.267} & \textbf{13.0} & \underline{0.277} & \underline{13.4} & 0.856 & 41.0 & 0.286 & 14.4 & 0.301 & 13.9 \\
         
& 14 & \textbf{0.281} & \textbf{13.1} & \underline{0.324} & 15.8 & 0.856 & 41.0 & 0.351 & 16.1 & 0.340 & \underline{14.9} \\
         
& 21 & \textbf{0.318} & \textbf{14.1} & \underline{0.319} & \underline{14.5} & 0.857 & 41.0 & 0.379 & 17.8 & 0.404 & 19.2 \\
         
& 28 & \textbf{0.367} & \textbf{16.5} & \underline{0.375} & \underline{16.8} & 0.857 & 41.0 & 0.379 & 18.2 & 0.411 & 19.1 \\
\midrule
\multirow{4}{*}{\rotatebox{90}{WEA-CT}} 
& 4 & \textbf{1.624} & \textbf{9.6} & 1.714 & 10.1 & 4.359 & 24.6 & \underline{1.629} & \underline{9.6} & 1.726 & 10.2 \\
         
& 12 & \textbf{3.543} & \textbf{20.4} & 3.563 & 20.5 & 4.359 & 24.7 & 3.588 & 20.6 & \underline{3.558} & \underline{20.5} \\
         
& 28 & \textbf{3.728} & \textbf{21.3} & 3.760 & 21.5 & 4.359 & 24.6 & 3.916 & 22.3 & \underline{3.758} & \underline{21.5} \\
         
& 120 & \underline{3.737} & \underline{21.4} & \textbf{3.734} & \textbf{21.3} & 4.356 & 24.6 & 3.815 & 21.7 & 3.763 & 21.5 \\
\midrule
\multirow{4}{*}{\rotatebox{90}{WEA-HK}} 
& 4 & \textbf{0.639} & \textbf{4.1} & 0.645 & 4.1 & 2.801 & 17.7 & \underline{0.639} & \underline{4.1} & 0.676 & 4.3 \\
         
& 12 & \textbf{1.235} & \textbf{7.7} & \underline{1.279} & \underline{8.0} & 2.801 & 17.7 & 1.325 & 8.3 & 1.281 & 8.1 \\
         
& 28 & \textbf{1.413} & \textbf{8.8} & \underline{1.463} & \underline{9.1} & 2.808 & 17.8 & 1.494 & 9.3 & 1.475 & 9.2 \\
         
& 120 & \textbf{1.547} & \textbf{9.6} & \underline{1.547} & \underline{9.6} & 2.810 & 17.8 & 1.639 & 10.1 & 1.621 & 10.0 \\
\midrule
\multirow{4}{*}{\rotatebox{90}{WEA-LD}} 
& 4 & \textbf{1.723} & \textbf{15.2} & 1.814 & 15.8 & 4.668 & 34.9 & \underline{1.740} & \underline{15.3} & 1.811 & 16.0 \\
         
& 12 & \textbf{2.959} & \textbf{23.9} & 2.986 & 24.1 & 4.661 & 34.9 & 3.008 & 24.1 & \underline{2.973} & \underline{24.0} \\
         
& 28 & \textbf{3.216} & \textbf{25.5} & 3.271 & 25.8 & 4.659 & 34.9 & 3.378 & 26.5 & \underline{3.268} & \underline{25.8} \\
         
& 120 & \textbf{3.246} & \textbf{25.8} & \underline{3.319} & \underline{26.2} & 4.668 & 34.9 & 3.667 & 28.5 & 3.407 & 26.7 \\
\midrule
\multirow{4}{*}{\rotatebox{90}{WEA-NY}} 
& 4 & \underline{1.272} & \underline{11.9} & 1.331 & 12.4 & 5.386 & 40.9 & \textbf{1.268} & \textbf{11.8} & 1.366 & 12.9 \\
         
& 12 & \underline{2.448} & \textbf{21.2} & 2.474 & 21.6 & 5.384 & 40.9 & 2.510 & 21.8 & \textbf{2.438} & \underline{21.5} \\
         
& 28 & \textbf{2.674} & \textbf{23.1} & \underline{2.708} & 23.4 & 5.387 & 40.9 & 2.797 & 23.8 & 2.724 & \underline{23.1} \\
         
& 120 & \textbf{2.713} & \textbf{23.2} & \underline{2.788} & \underline{23.5} & 5.389 & 40.9 & 3.063 & 25.6 & 2.911 & 24.6 \\
\midrule
\multirow{4}{*}{\rotatebox{90}{WEA-SG}} 
& 4 & \underline{0.344} & \underline{1.3} & 0.346 & 1.3 & 0.700 & 2.5 & \textbf{0.341} & \textbf{1.2} & 0.346 & 1.3 \\
         
& 12 & \textbf{0.416} & \textbf{1.5} & \underline{0.421} & \underline{1.5} & 0.698 & 2.5 & 0.424 & 1.5 & 0.425 & 1.5 \\
         
& 28 & \textbf{0.465} & \textbf{1.7} & 0.473 & 1.7 & 0.698 & 2.5 & 0.495 & 1.8 & \underline{0.471} & \underline{1.7} \\
         
& 120 & \textbf{0.483} & \textbf{1.8} & 0.496 & 1.8 & 0.697 & 2.5 & 0.512 & 1.9 & \underline{0.485} & \underline{1.8} \\
\bottomrule
\end{tabular}
% }
\caption{Performance of Sonnet across the ILI and WEA tasks with different attention modules. \bestResults{} MAE values are rounded to 3 decimals, sMAPE to 1 for spacing.}
\label{tab:results_other_attn}
\end{table}

By replacing na\"ive attention with the proposed MVCA module, we obtain significant performance gains across all ILI tasks, reducing MAE by $10.7\%$ on average across all base models. The average MAE reduction for PatchTST is greater ($15.1\%$) compared to the other base models ($10.4\%$ for Samformer and $6.7\%$ for iTransformer), indicating that the application of MVCA yields enhanced performance gains for the model that does not capture inter-variable dependencies. When compared to other modified attention modules, MVCA shows competitive performance, reducing MAE by $2.8\%$ on average. Compared to the best-performing baseline (VDAB from DeformTime), it reduces MAE and sMAPE by $3.5\%$ and $3\%$, respectively. While the gains in the ENG region are less pronounced, MVCA achieves consistently the best results across the US regions, and VDAB is the second best; both methods capture inter-variable dependencies while others do not. Comparably, replacing na\"ive attention with the Fourier attention modules from FEDformer and FNet results in worse performance.

\subsection{Sonnet with Different Attention Variants}
\label{subsec:Sonnet_attention_effect}
To understand how different attention mechanisms affect the forecasting accuracy of Sonnet, we replace the MVCA module with the aforementioned modified attention mechanisms (see section~\ref{subsec:forecasting_results_vca}). Experiments are conducted on both the ILI and WEA forecasting tasks across all locations, forecasting horizons, and test periods. Hyperparameters are re-tuned for each attention variant using the same validation procedure. Appendix~\ref{appsubsec:select_attn} provides further details.

Results are enumerated in Table~\ref{tab:results_other_attn}. Overall, using MVCA yields the best accuracy compared to other attention modules, outperforming them in $26$ out of the $32$ tasks (MAE). On average, it reduces the MAE and sMAPE scores by $2.9\%$ and $2\%$, respectively. The most competitive baseline module is VDAB (DeformTime). MVCA outperforms it by $3.6\%$ and $3\%$ on average w.r.t. MAE and sMAPE, respectively. The worst-performing attention module is FED (FEDformer). MVCA reduces its MAE by $51.5\%$. FED focuses on capturing seasonality information using frequency transformation without considering the dependencies between variables. This shows that modelling seasonality alone is insufficient for MTS forecasting, particularly when key predictive information comes from exogenous variables. The performance obtained by using either na\"ive attention or FNet is similar. MVCA reduces their MAE by $7\%$ and $6.6\%$ on average, respectively. This is expected as FNet does not introduce learnable parameters to the na\"ive attention.

\section{Conclusion}
\label{sec:conclusion}
In this paper, we present Sonnet, a novel model for multivariable time series forecasting tasks. Sonnet operates in the spectral dimension using learnable wavelet transforms and captures variable dependencies with frequency-based coherence. It then predicts future temporal dynamics with the Koopman operator. Experiments are conducted on $12$ data sets from different application domains, with $9$ of them containing multiple test seasons for a more comprehensive analysis. Sonnet yields the best performance in $34$ out of $47$ tasks, reducing MAE by $2.2\%$ and $1.1\%$ on average compared to the most performant baseline overall (DeformTime) or per task (can vary), respectively. Additionally, our experiments highlight that replacing vanilla attention, present in various forecasting models, with the proposed MVCA module improves prediction accuracy by a large margin.

\section*{Acknowledgements} The authors would like to thank the RCGP for providing ILI rates for England. V. Lampos would like to acknowledge all levels of support from the EPSRC grant EP/X031276/1.

{\small
\bibliography{reference}
}

\vspace{1in}

% \clearpage

\appendix

% \@toptitlebar
% \section*{Appendix}

\begin{center}
% \LARGE \bfseries Appendix
\LARGE \bfseries Supplementary Materials
\end{center}

% \@bottomtitlebar
\vspace{0.2in}

\setcounter{table}{0}
\renewcommand{\thetable}{S\arabic{table}}%
\setcounter{figure}{0}
\renewcommand{\thefigure}{S\arabic{figure}}%
\setcounter{equation}{0}
\renewcommand{\theequation}{S\arabic{equation}}%

\section{Data sets for multivariable time series forecasting}
\label{appsec:datasets}
In this section, we provide a detailed description of all the data sets used in our MTS forecasting experiments. The weather data set that is formed by us is available in our code and data repository.

\subsection{Hourly electricity transformer temperature prediction}
The Hourly Electricity transformer temperature (ETTh1 and ETTh2) data sets are two widely used benchmarks in time series forecasting, obtained from $2$ different counties in China~\citep{zhou2021informer}.\footnote{ETTh1/h2, \url{github.com/zhouhaoyi/ETDataset}} The target variable of both data sets is the oil temperature. They contain $6$ exogenous variables capturing power load attributes. Both data sets cover the period from July 1, 2016, to June 26, 2018. We use only a subset of $14{,}400$ time steps starting from July 1, 2016, to match prior evaluation setups of the models that we are comparing to~\citep{nie2023time,liu2024itransformer,shu2025deformtime}. Within the subset, the first $8{,}640$ time steps are used for training, the next $2{,}880$ for validation, and the last $2{,}880$ for testing.

\subsection{Electricity consumption prediction}
The Electricity Consumption (ELEC) data set provides the electricity consumption records for Zurich (Switzerland), collected every $15$ minutes.\footnote{Electricity data set, \url{github.com/unit8co/darts}} ELEC contains measurements for both residential households (recorded at low voltage) and commercial businesses (recorded at medium voltage). We choose the electricity consumption of households as our target variable for prediction. External predictors of the electricity consumption include $8$ weather-related covariates recorded hourly by a weather station in the city of Zurich.
 
Note that for some time intervals, certain predictor values may be missing. These are imputed using linear interpolation. We then resample rates to an hourly frequency following prior work in electricity consumption prediction~\citep{møllerandersen2017households,yukseltan2020hourly}.

For the electricity consumption task, we use two test periods, years 2020 ($8{,}784$ time steps, including the leap day) and 2021 ($8{,}760$ time steps). The year preceding the test year is used as the validation ($8{,}760$ and $8{,}784$ time steps, respectively), while the $4$ years before validation form the training set ($35{,}064$ in both evaluations).

\subsection{Energy price prediction}
The energy price (ENER) data set contains $4$ years of electrical consumption, generation, and pricing data, with the main target of predicting the actual energy price in Spain.\footnote{Energy price data set, \url{github.com/unit8co/darts}}
Note that the original data sets have $27$ exogenous variables in total. However, $8$ variables consist entirely of zeros or missing values, which do not provide useful information for forecasting purposes. We therefore exclude these variables from the analysis and use the remaining $19$. Data is sampled on an hourly basis. Data from December 31, 2014 to December 31, 2016 are used for training ($17{,}545$ samples). The subsequent year (January 1 to December 31, 2017; $8{,}760$ samples) is used for validation. For testing, we use data from January 1, 2018, to 22:00, December 31, 2018 ($8{,}759$ samples); the price for the last hour is not available. 

\begin{table*}[!t]
\renewcommand{\arraystretch}{0.92}
\centering
\small
\setlength{\tabcolsep}{2.8pt}
\setlength{\aboverulesep}{0.5pt}
\setlength{\belowrulesep}{0.5pt}
% \resizebox{\linewidth}{!}{
\begin{tabular}{cl ccc ccc ccc ccc ccc}
\toprule
\multirow{2}{*}{\textbf{H}} & \multirow{2}{*}{\textbf{Model}} & \multicolumn{3}{c}{\textbf{2016}} & \multicolumn{3}{c}{\textbf{2017}} & \multicolumn{3}{c}{\textbf{2018}} & \multicolumn{3}{c}{\textbf{2019}} & \multicolumn{3}{c}{\textbf{Average}} \\
 & & $r$ & MAE & $\epsilon\%$ & $r$ & MAE & $\epsilon\%$ & $r$ & MAE & $\epsilon\%$ & $r$ & MAE & $\epsilon\%$ & $r$ & MAE & $\epsilon\%$ \\
\cmidrule{1-1}\cmidrule{2-2}\cmidrule(lr){3-5} \cmidrule(lr){6-8} \cmidrule(lr){9-11} \cmidrule(lr){12-14} \cmidrule(lr){15-17}
    \multirow{9}{*}{7} & DLinear      & 0.8710 & 2.6985 & 38.86 & 0.8769 & 1.6879 & 33.77 & 0.8231 & 4.2734 & 50.14 & 0.8906 & 2.6258 & 49.30 & 0.8654 & 2.8214 & 43.02 \\
        & iTransformer & 0.8773 & 2.0645 & 27.26 & 0.8911 & 1.3734 & 24.06 & 0.8180 & 4.1077 & 31.26 & 0.8911 & 1.6881 & \underline{22.94} & 0.8694 & 2.3084 & 26.38 \\
        & PatchTST     & 0.8832 & 1.9823 & 26.13 & 0.8744 & 1.4307 & 25.39 & 0.8475 & 3.7925 & 31.23 & 0.8526 & 2.0403 & 27.70 & 0.8644 & 2.3115 & 27.61 \\
        & TimeXer      & 0.8927 & 1.9807 & 26.29 & 0.8773 & 1.5327 & 29.94 & 0.6974 & 5.1695 & 43.18 & 0.7564 & 2.5507 & 35.22 & 0.8060 & 2.8084 & 33.66 \\
        & Samformer    & 0.8780 & 2.0244 & 27.60 & 0.8922 & 1.4025 & 25.10 & 0.7718 & 4.1902 & 35.01 & 0.8857 & 1.7729 & 25.55 & 0.8569 & 2.3475 & 28.31 \\
        & ModernTCN    & 0.9490 & 1.6043 & \underline{17.82} & 0.9606 & \underline{1.1988} & 22.75 & \textbf{0.9679} & 3.2340 & 40.55 & 0.9316 & 1.7584 & 31.98 & 0.9523 & 1.9489 & 28.27 \\
        & Crossformer  & 0.8879 & 2.3314 & 36.08 & 0.9591 & \textbf{0.8606} & \textbf{17.82} & 0.9234 & 2.9133 & \underline{27.04} & 0.9436 & \underline{1.3737} & \textbf{21.89} & 0.9285 & 1.8698 & \underline{25.70} \\
        & DeformTime   & \textbf{0.9728} & \underline{1.3786} & 19.57 & \underline{0.9665} & 1.3083 & 29.49 & 0.9585 & \underline{2.6313} & 41.40 & \underline{0.9436} & \textbf{1.2485} & 23.96 & \underline{0.9603} & \underline{1.6417} & 28.60 \\
        & Sonnet       & \underline{0.9621} & \textbf{1.0294} & \textbf{13.44} & \textbf{0.9772} & 1.2003 & \underline{20.79} & \underline{0.9590} & \textbf{2.0369} & \textbf{25.10} & \textbf{0.9590} & 1.6498 & 28.02 & \textbf{0.9643} & \textbf{1.4791} & \textbf{21.84} \\
    \midrule
    \multirow{9}{*}{14} & DLinear      & 0.7810 & 3.6354 & 52.29 & 0.7857 & 2.1872 & 42.47 & 0.6834 & 5.5368 & 60.91 & 0.8351 & 3.8092 & 65.47 & 0.7713 & 3.7922 & 55.29 \\
        & iTransformer & 0.7731 & 2.7620 & 36.46 & 0.7776 & 1.9929 & 34.17 & 0.6576 & 5.7796 & 44.34 & 0.7850 & 2.3862 & 31.71 & 0.7483 & 3.2301 & 36.67 \\
        & PatchTST     & 0.8078 & 2.6267 & 35.59 & 0.7730 & 1.9562 & 33.55 & 0.6722 & 5.6075 & 43.64 & 0.7218 & 2.8284 & 38.25 & 0.7437 & 3.2547 & 37.76 \\
        & TimeXer      & 0.7891 & 2.6580 & 35.27 & 0.7910 & 1.9864 & 36.99 & 0.5503 & 6.2678 & 53.64 & 0.6646 & 3.0626 & 41.61 & 0.6988 & 3.4937 & 41.88 \\
        & Samformer    & 0.7205 & 3.0697 & 41.49 & 0.8113 & 1.7285 & 28.83 & 0.6888 & 4.9426 & 39.40 & 0.8246 & 2.3754 & 36.78 & 0.7613 & 3.0290 & 36.63 \\
        & ModernTCN    & 0.8814 & 2.1727 & 31.93 & 0.8864 & 1.4953 & 28.61 & 0.8905 & 4.6759 & 48.74 & 0.8441 & 2.4760 & 34.76 & 0.8756 & 2.7050 & 36.01 \\
        & Crossformer  & 0.7940 & 2.9124 & 41.75 & 0.8843 & 1.4711 & \underline{23.39} & 0.8470 & 3.8295 & \underline{30.88} & 0.9007 & 2.4044 & \textbf{27.86} & 0.8565 & 2.6543 & \underline{30.97} \\
        & DeformTime   & \textbf{0.9259} & \underline{2.0556} & \underline{25.36} & \underline{0.9400} & \underline{1.3180} & 33.86 & \underline{0.8964} & \underline{3.7631} & 47.33 & \underline{0.9154} & \textbf{1.7863} & \underline{29.36} & \underline{0.9194} & \underline{2.2308} & 33.98 \\
        & Sonnet       & \underline{0.9025} & \textbf{1.5327} & \textbf{22.20} & \textbf{0.9512} & \textbf{1.2453} & \textbf{21.28} & \textbf{0.9233} & \textbf{2.8313} & \textbf{28.62} & \textbf{0.9307} & \underline{2.0808} & 30.99 & \textbf{0.9269} & \textbf{1.9225} & \textbf{25.77} \\
    \midrule
    \multirow{9}{*}{21} & DLinear      & 0.4466 & 4.1958 & 56.02 & 0.6441 & 2.8294 & 52.56 & 0.4920 & 6.9514 & 71.66 & 0.6608 & 3.9192 & 64.75 & 0.5609 & 4.4739 & 61.25 \\
        & iTransformer & 0.5310 & 4.1390 & 55.62 & 0.5987 & 2.6718 & 43.65 & 0.4403 & 7.0816 & 55.17 & 0.6556 & 3.0464 & 41.29 & 0.5564 & 4.2347 & 48.93 \\
        & PatchTST     & 0.6126 & 3.6607 & 48.57 & 0.6214 & 2.6377 & 44.77 & 0.4427 & 7.3828 & 60.54 & 0.5861 & 3.5956 & 50.55 & 0.5657 & 4.3192 & 51.11 \\
        & TimeXer      & 0.5958 & 3.7654 & 49.46 & 0.6981 & 2.4045 & 41.17 & 0.4377 & 7.3686 & 62.71 & 0.5301 & 3.7962 & 52.91 & 0.5654 & 4.3337 & 51.57 \\
        & Samformer    & 0.4919 & 4.2610 & 57.57 & 0.5352 & 2.8996 & 43.99 & 0.5263 & 6.8716 & 60.01 & 0.4793 & 3.9598 & 56.07 & 0.5082 & 4.4980 & 54.41 \\
        & ModernTCN    & 0.7869 & 2.3283 & 27.42 & 0.8590 & \underline{1.6790} & \underline{30.86} & \underline{0.9085} & 5.7489 & 61.11 & 0.8357 & \underline{2.4040} & \underline{40.67} & 0.8475 & 3.0400 & 40.02 \\
        & Crossformer  & 0.8113 & 2.6710 & 36.42 & 0.8014 & 1.8035 & 31.29 & 0.8504 & \underline{4.5615} & 49.08 & 0.7331 & 2.9697 & 45.48 & 0.7991 & 3.0014 & 40.57 \\
        & DeformTime   & \textbf{0.8819} & \underline{2.1700} & \underline{27.01} & \textbf{0.8859} & 1.8980 & 32.65 & \textbf{0.9110} & 4.8984 & \underline{44.78} & \textbf{0.9092} & \textbf{1.6335} & \textbf{26.36} & \textbf{0.8970} & \underline{2.6500} & \textbf{32.70} \\
        & Sonnet       & \underline{0.8537} & \textbf{1.9394} & \textbf{25.77} & \underline{0.8720} & \textbf{1.5679} & \textbf{29.12} & 0.8260 & \textbf{3.9711} & \textbf{43.60} & \underline{0.9042} & 2.5620 & 47.62 & \underline{0.8640} & \textbf{2.5101} & \underline{36.53} \\
    \midrule
    \multirow{9}{*}{28} & DLinear      & 0.3917 & 4.5039 & 60.30 & 0.5078 & 3.3928 & 60.94 & 0.3911 & 7.7738 & 77.58 & 0.5774 & 4.4683 & 72.19 & 0.4670 & 5.0347 & 67.75 \\
        & iTransformer & 0.3492 & 4.9542 & 65.45 & 0.6761 & 2.2745 & 38.38 & 0.2947 & 8.3463 & 69.39 & 0.5286 & 3.6750 & 48.18 & 0.4622 & 4.8125 & 55.35 \\
        & PatchTST     & 0.4271 & 4.5240 & 60.22 & 0.4477 & 3.2658 & 53.01 & 0.3203 & 8.0693 & 65.91 & 0.4486 & 4.1265 & 59.27 & 0.4109 & 4.9964 & 59.60 \\
        & TimeXer      & 0.4978 & 4.2217 & 56.56 & 0.5523 & 2.8584 & 48.91 & 0.3413 & 8.1493 & 71.44 & 0.3337 & 4.3758 & 69.50 & 0.4313 & 4.9013 & 61.60 \\
        & Samformer    & 0.3995 & 4.7015 & 62.97 & 0.3194 & 3.5264 & 55.69 & 0.3127 & 8.1404 & 66.57 & 0.3636 & 4.2709 & 57.89 & 0.3488 & 5.1598 & 60.78 \\
        & ModernTCN    & \underline{0.8017} & \underline{2.6486} & \underline{36.17} & 0.7628 & 2.6508 & 51.35 & 0.7544 & \underline{4.8228} & 49.19 & 0.8152 & 3.3222 & 54.78 & 0.7835 & 3.3611 & 47.87 \\
        & Crossformer  & 0.7299 & 3.1253 & 41.56 & 0.8082 & 1.8197 & \underline{33.57} & \underline{0.7841} & 5.4882 & 55.97 & \underline{0.9349} & \underline{2.3599} & 53.48 & 0.8143 & 3.1983 & 46.15 \\
        & DeformTime   & 0.7919 & 2.6540 & 49.29 & \textbf{0.9229} & \underline{1.8100} & 41.87 & \textbf{0.8493} & 4.8324 & \textbf{37.53} & \textbf{0.9499} & \textbf{1.5947} & \textbf{33.06} & \textbf{0.8785} & \textbf{2.7228} & \underline{40.44} \\
        & Sonnet       & \textbf{0.8975} & \textbf{2.0133} & \textbf{31.59} & \underline{0.8841} & \textbf{1.7490} & \textbf{33.24} & 0.7745 & \textbf{4.4605} & \underline{41.57} & 0.7935 & 2.7698 & \underline{41.41} & \underline{0.8374} & \underline{2.7481} & \textbf{36.95} \\
\bottomrule
\end{tabular}
% }
\caption{\ILITableCaptionAppendix{England}{ILI-ENG}}
\label{tab:results_ili_eng}
\end{table*}

\subsection{Weather forecasting on selected locations}
\label{appsubsec:weather_dataset}
We form the weather forecasting (WEA) data set using climate information extracted from the WeatherBench repository~\citep{rasp2020weatherbench}, covering the period from 1979 to 2018.\footnote{WeatherBench repository, \\ \url{dataserv.ub.tum.de/index.php/s/m1524895}}
Data is sampled from a uniform spatial grid over the globe with a resolution of $5.625^\circ$ in both latitude and longitude. Each grid point represents a distinct geographical location. To capture different climatic conditions, we select five spatially diverse cities from the grid: London (UK), New York (US), Hong Kong (China), Cape Town (South Africa), and Singapore. Among these, London, New York, and Hong Kong are located in the northern hemisphere, with Hong Kong being in the subtropical zone; Singapore lies near the equator, and Cape Town is in the southern hemisphere. The two cities that are closer to the equator (Hong Kong and Singapore) do not have strong seasonal temperature fluctuations over the year. Consequently, we expect models to yield lower prediction errors at these two locations. For each city, climate data is sampled from the nearest grid point to its coordinates.\footnote{We obtain the longitude and latitude of each city using OpenStreetMap (\url{wiki.openstreetmap.org}).} Specifically, we obtain the absolute distance in both latitude and longitude between each city and all grid points, and select the grid point with the smallest distance along both axes. The forecasting target is the $850$ hPa temperature in Kelvin (T850) at the central grid point,\footnote{Further information about T850 is provided at \\ \url{charts.ecmwf.int/products/medium-z500-t850}.} which is an important climate indicator that has been widely used in climate research~\citep{scherrer2004analysis,hamill2007ensemble}. To form exogenous inputs, we include four other climate variables at each selected grid point following prior work~\citep{nguyen2023climax,verma2024climode}. These are the $500$ hPa geopotential (Z500), $2$-meter air temperature (t2m), $10$-meter zonal wind (u10), and $10$-meter meridional wind (v10). We also include these $4$ variables and T850 from the surrounding eight grid points, forming a $3 \times 3$ spatial window centred on each selected grid, resulting in $44$ exogenous variables in total. The data is resampled to maintain a temporal resolution of $6$ hours similarly to~\citet{verma2024climode}.

We consider $3$ test seasons: years 2016, 2017, and 2018 (Jan. 1 to Dec. 31). Each test year contains $1{,}464$, $1{,}460$, and $1{,}460$ time steps, respectively. The year immediately preceding a test year is used for validation (2015, 2016, and 2017). Validation year 2015 contains $1{,}460$ time steps. The training sets include data from January 1, 1980, up to the date preceding the corresponding validation year, resulting in $51{,}136$, $52{,}596$, and $54{,}060$ time steps for the $3$ test seasons.

\begin{table*}[!t]
\renewcommand{\arraystretch}{0.92}
\centering
\small
\setlength{\tabcolsep}{2.8pt}
\setlength{\aboverulesep}{0.5pt}
\setlength{\belowrulesep}{0.5pt}
% \resizebox{\linewidth}{!}{
\begin{tabular}{cl ccc ccc ccc ccc ccc}
\toprule
\multirow{2}{*}{\textbf{H}} & \multirow{2}{*}{\textbf{Model}} & \multicolumn{3}{c}{\textbf{2016}} & \multicolumn{3}{c}{\textbf{2017}} & \multicolumn{3}{c}{\textbf{2018}} & \multicolumn{3}{c}{\textbf{2019}} & \multicolumn{3}{c}{\textbf{Average}} \\
 & & $r$ & MAE & $\epsilon\%$ & $r$ & MAE & $\epsilon\%$ & $r$ & MAE & $\epsilon\%$ & $r$ & MAE & $\epsilon\%$ & $r$ & MAE & $\epsilon\%$ \\
\cmidrule{1-1}\cmidrule{2-2}\cmidrule(lr){3-5} \cmidrule(lr){6-8} \cmidrule(lr){9-11} \cmidrule(lr){12-14} \cmidrule(lr){15-17}
    \multirow{9}{*}{7} & DLinear      & 0.7456 & 0.4982 & 29.23 & 0.7980 & 0.8210 & 31.85 & 0.7133 & 1.0910 & 30.82 & 0.8955 & 0.5318 & 19.84 & 0.7881 & 0.7355 & 27.94 \\
        & iTransformer & 0.7986 & 0.3865 & \underline{21.30} & 0.8171 & 0.7367 & 27.07 & 0.7784 & 0.9080 & 23.22 & 0.8545 & 0.5717 & 21.38 & 0.8122 & 0.6507 & 23.24 \\
        & PatchTST     & 0.7666 & 0.4290 & 23.78 & 0.7973 & 0.7546 & 26.97 & 0.6937 & 1.1018 & 26.52 & 0.8624 & 0.5533 & 20.79 & 0.7800 & 0.7097 & 24.51 \\
        & TimeXer      & 0.8024 & 0.4214 & 24.76 & 0.8480 & 0.6882 & 27.16 & 0.8494 & 0.8233 & 22.90 & 0.8859 & 0.5001 & 18.70 & 0.8464 & 0.6083 & 23.38 \\
        & Samformer    & 0.7549 & 0.4346 & 23.83 & 0.7935 & 0.7784 & 28.57 & 0.8563 & 0.8725 & 25.30 & 0.8834 & 0.5126 & 19.16 & 0.8220 & 0.6495 & 24.21 \\
        & ModernTCN    & 0.8745 & 0.3874 & 21.99 & \underline{0.9359} & 0.5216 & \textbf{17.45} & \underline{0.9569} & \underline{0.4782} & 14.12 & 0.9535 & 0.3722 & 12.64 & \underline{0.9302} & 0.4398 & 16.55 \\
        & Crossformer  & 0.8796 & \underline{0.3382} & 21.39 & 0.9107 & \underline{0.4975} & \underline{18.32} & 0.9009 & 0.6309 & 15.61 & 0.9523 & \underline{0.2936} & \textbf{10.52} & 0.9109 & 0.4400 & 16.46 \\
        & DeformTime   & \underline{0.8887} & 0.3428 & 21.86 & \textbf{0.9463} & \textbf{0.4796} & 18.66 & 0.9008 & 0.5369 & \underline{12.88} & \textbf{0.9622} & \textbf{0.2894} & \underline{10.64} & 0.9245 & \underline{0.4122} & \underline{16.01} \\
        & Sonnet       & \textbf{0.9027} & \textbf{0.2854} & \textbf{16.40} & 0.9336 & 0.5471 & 18.68 & \textbf{0.9683} & \textbf{0.3708} & \textbf{11.52} & \underline{0.9537} & 0.3192 & 12.97 & \textbf{0.9396} & \textbf{0.3806} & \textbf{14.89} \\
    \midrule
    \multirow{9}{*}{14} & DLinear      & 0.6546 & 0.5761 & 33.62 & 0.7335 & 0.9406 & 36.90 & 0.6007 & 1.2213 & 34.62 & 0.8261 & 0.6360 & 23.74 & 0.7037 & 0.8435 & 32.22 \\
        & iTransformer & 0.6864 & 0.4894 & 26.35 & 0.7548 & 0.8934 & 33.26 & 0.6025 & 1.1639 & 30.28 & 0.8294 & 0.6115 & 22.78 & 0.7183 & 0.7896 & 28.17 \\
        & PatchTST     & 0.6628 & 0.5103 & 27.27 & 0.7350 & 0.9267 & 34.44 & 0.5993 & 1.3085 & 31.82 & 0.7922 & 0.7085 & 26.90 & 0.6973 & 0.8635 & 30.11 \\
        & TimeXer      & 0.7178 & 0.4823 & 27.24 & 0.7446 & 0.9211 & 36.47 & 0.7118 & 1.0505 & 28.09 & 0.8164 & 0.6362 & 24.47 & 0.7476 & 0.7725 & 29.07 \\
        & Samformer    & 0.7204 & 0.4929 & 28.49 & 0.7359 & 0.9580 & 37.34 & 0.8198 & 0.9533 & 29.98 & 0.7851 & 0.6741 & 24.81 & 0.7653 & 0.7696 & 30.16 \\
        & ModernTCN    & \underline{0.8093} & 0.4647 & 25.78 & \underline{0.9000} & 0.6482 & 23.18 & \underline{0.9360} & \underline{0.6020} & 17.05 & \underline{0.9572} & 0.3968 & \underline{14.87} & 0.9006 & 0.5279 & 20.22 \\
        & Crossformer  & 0.7872 & \underline{0.4357} & 24.61 & 0.8640 & 0.6502 & 22.95 & 0.8057 & 0.8159 & 20.83 & 0.9241 & 0.4389 & 15.54 & 0.8453 & 0.5852 & 20.98 \\
        & DeformTime   & 0.8050 & 0.4404 & \underline{23.82} & \textbf{0.9271} & \textbf{0.5029} & \textbf{18.53} & 0.9126 & 0.6351 & \underline{16.34} & \textbf{0.9652} & \textbf{0.3226} & \textbf{12.21} & \underline{0.9025} & \underline{0.4752} & \textbf{17.73} \\
        & Sonnet       & \textbf{0.8654} & \textbf{0.3477} & \textbf{20.08} & 0.8832 & \underline{0.6254} & \underline{21.81} & \textbf{0.9731} & \textbf{0.4380} & \textbf{15.30} & 0.9499 & \underline{0.3854} & 16.34 & \textbf{0.9179} & \textbf{0.4491} & \underline{18.38} \\
    \midrule
    \multirow{9}{*}{21} & DLinear      & 0.6016 & 0.6023 & 35.11 & 0.6897 & 0.9671 & 38.06 & 0.4966 & 1.3272 & 38.76 & 0.7936 & 0.7530 & 27.80 & 0.6454 & 0.9124 & 34.93 \\
        & iTransformer & 0.6418 & 0.4790 & \textbf{24.54} & 0.7515 & 0.8784 & 34.22 & 0.8230 & 1.0845 & 31.85 & 0.7321 & 0.7748 & 29.52 & 0.7371 & 0.8042 & 30.03 \\
        & PatchTST     & 0.5775 & 0.5592 & 29.41 & 0.6286 & 1.1661 & 43.41 & 0.5189 & 1.4395 & 38.13 & 0.6478 & 0.9495 & 35.86 & 0.5932 & 1.0286 & 36.70 \\
        & TimeXer      & 0.4856 & 0.5815 & 32.17 & 0.7739 & 0.8696 & 33.97 & 0.6529 & 1.1238 & 32.12 & 0.7482 & 0.7225 & 27.56 & 0.6651 & 0.8243 & 31.46 \\
        & Samformer    & 0.6042 & 0.5472 & 28.91 & 0.7109 & 0.9739 & 35.70 & 0.7263 & 1.1426 & 35.51 & 0.7586 & 0.6860 & 25.58 & 0.7000 & 0.8374 & 31.42 \\
        & ModernTCN    & \textbf{0.8104} & 0.4656 & 26.17 & 0.8746 & 0.6790 & 30.30 & \underline{0.9105} & 0.7258 & \underline{20.75} & \underline{0.9487} & 0.4421 & 18.19 & \textbf{0.8860} & 0.5781 & 23.85 \\
        & Crossformer  & \underline{0.7729} & 0.4634 & 26.54 & 0.8996 & 0.6688 & \underline{25.62} & 0.7621 & 0.8882 & 23.17 & 0.9354 & 0.4777 & \textbf{13.82} & 0.8425 & 0.6245 & 22.29 \\
        & DeformTime   & 0.7414 & \textbf{0.4568} & 25.94 & \underline{0.9028} & \underline{0.6189} & 26.44 & \textbf{0.9313} & \underline{0.6981} & 21.04 & 0.9408 & \underline{0.3963} & \underline{15.09} & \underline{0.8791} & \underline{0.5425} & \underline{22.12} \\
        & Sonnet       & 0.7670 & \underline{0.4571} & \underline{25.22} & \textbf{0.9164} & \textbf{0.6012} & \textbf{23.16} & 0.8722 & \textbf{0.6961} & \textbf{18.91} & \textbf{0.9536} & \textbf{0.3759} & 15.28 & 0.8773 & \textbf{0.5326} & \textbf{20.64} \\
    \midrule
    \multirow{9}{*}{28} & DLinear      & 0.4836 & 0.6512 & 37.96 & 0.6112 & 1.0328 & 41.19 & 0.4125 & 1.4045 & 40.27 & 0.6840 & 0.8337 & 31.08 & 0.5479 & 0.9805 & 37.63 \\
        & iTransformer & 0.5258 & 0.5949 & 32.54 & 0.6567 & 1.0632 & 41.38 & 0.5726 & 1.3053 & 38.82 & 0.6738 & 0.8842 & 34.26 & 0.6072 & 0.9619 & 36.75 \\
        & PatchTST     & 0.4013 & 0.6678 & 36.55 & 0.5274 & 1.2615 & 48.85 & 0.3903 & 1.6234 & 44.67 & 0.5579 & 1.0576 & 40.39 & 0.4692 & 1.1525 & 42.61 \\
        & TimeXer      & 0.3979 & 0.5856 & 32.82 & 0.7017 & 0.9748 & 39.33 & 0.6137 & 1.2462 & 35.29 & 0.7037 & 0.8229 & 31.45 & 0.6042 & 0.9074 & 34.72 \\
        & Samformer    & 0.4934 & 0.5858 & 31.55 & 0.6947 & 1.0083 & 39.01 & 0.6215 & 1.4670 & 50.33 & 0.7698 & 0.6945 & 26.31 & 0.6448 & 0.9389 & 36.80 \\
        & ModernTCN    & \underline{0.7635} & 0.4702 & \underline{27.09} & 0.8667 & 0.6742 & 27.31 & \underline{0.9223} & \underline{0.6618} & \underline{19.83} & \underline{0.9604} & \underline{0.4776} & 20.41 & \underline{0.8782} & \underline{0.5710} & 23.66 \\
        & Crossformer  & 0.7102 & \underline{0.4697} & 27.38 & 0.9064 & 0.7624 & 28.93 & 0.8432 & 0.8253 & 20.69 & 0.8782 & 0.5474 & 18.64 & 0.8345 & 0.6512 & 23.91 \\
        & DeformTime   & \textbf{0.8369} & 0.4716 & 27.21 & \textbf{0.9562} & \textbf{0.6017} & \underline{25.66} & \textbf{0.9253} & \textbf{0.5954} & \textbf{17.62} & 0.9294 & 0.5463 & \underline{18.52} & \textbf{0.9119} & \textbf{0.5538} & \underline{22.25} \\
        & Sonnet       & 0.7578 & \textbf{0.4614} & \textbf{25.04} & \underline{0.9140} & \underline{0.6458} & \textbf{22.49} & 0.8747 & 0.7363 & 21.05 & \textbf{0.9609} & \textbf{0.4717} & \textbf{16.01} & 0.8768 & 0.5788 & \textbf{21.15} \\
\bottomrule
\end{tabular}
% }
\caption{\ILITableCaptionAppendix{US Region 2}{ILI-US2}}
\label{tab:results_ili_us2}
\end{table*}

\subsection{Influenza-like-illness rate prediction}
\label{appsubsec:ili_rate}
We construct an ILI forecasting data set for three regions: England (UK), and US HHS Regions 2 and 9, following the setup in prior work~\citep{shu2025deformtime}. Ground truth ILI rates are obtained from the Royal College of General Practitioners (RCGP) in England and the Centers for Disease Control and Prevention (CDC) for the US regions. Weekly ILI rates are linearly interpolated to a daily resolution. Specifically, we assume the ILI rates are reported corresponding to the middle day of each week, which is Wednesday for the US regions and Thursday for England~\citep{shu2025deformtime}. We also incorporate the ILI rate reporting delay of syndromic surveillance systems; in England there is a lag of $\delta = 7$ days and $\delta = 14$ days for the US regions~\citep{wagner2018added,reich2019accuracy}.

We use web search activity data (daily frequency time series of different search terms), obtained from the Google Health Trends API, as the exogenous indicators. We follow the setup of DeformTime~\citep{shu2025deformtime} to form the set of search queries for our experiments.\footnote{Search queries can be obtained from \url{github.com/ClaudiaShu/DeformTime}.} Note that for the US regions, search frequencies are collected at the state level, and the query data for each region is obtained using weighted averages based on the population of the states inside each region (New Jersey/New York is $0.32/0.68$ for Region 2 and Arizona/California is $0.16/0.84$ for Region 9).

We evaluate forecasting accuracy using $4$ test seasons (2015/16, 2016/17, 2017/18, and 2018/19) in England and the US regions, each corresponding to a 12-month period that encompasses at least one seasonal flu outbreak. For England, each season begins on September $1$ and ends on August $31$ of the next year, while for the US regions, it is from August $1$ to July $31$ of the following year. For each test season, models are trained on the preceding $9$ seasons. Validation sets are extracted from the last three seasons in the training data. Each validation set covers $180$ days, containing three $60$-day segments that capture the onset (from the last season), peak (from the penultimate season), and outset (the $3$rd to last season) phases of an influenza season. The onset is defined as the first point followed by two weeks above a predefined ILI threshold; the peak is the point of the highest ILI rate; and the outset is the last point followed by two weeks above the threshold. These selected points are then considered as the $30$th day in a $60$-day validation period.

\begin{table*}[!t]
\renewcommand{\arraystretch}{0.92}
\centering
\small
\setlength{\tabcolsep}{2.8pt}
\setlength{\aboverulesep}{0.5pt}
\setlength{\belowrulesep}{0.5pt}
% \resizebox{\linewidth}{!}{
\begin{tabular}{cl ccc ccc ccc ccc ccc}
\toprule
\multirow{2}{*}{\textbf{H}} & \multirow{2}{*}{\textbf{Model}} & \multicolumn{3}{c}{\textbf{2016}} & \multicolumn{3}{c}{\textbf{2017}} & \multicolumn{3}{c}{\textbf{2018}} & \multicolumn{3}{c}{\textbf{2019}} & \multicolumn{3}{c}{\textbf{Average}} \\
 & & $r$ & MAE & $\epsilon\%$ & $r$ & MAE & $\epsilon\%$ & $r$ & MAE & $\epsilon\%$ & $r$ & MAE & $\epsilon\%$ & $r$ & MAE & $\epsilon\%$ \\
\cmidrule{1-1}\cmidrule{2-2}\cmidrule(lr){3-5} \cmidrule(lr){6-8} \cmidrule(lr){9-11} \cmidrule(lr){12-14} \cmidrule(lr){15-17}
    \multirow{9}{*}{7} & DLinear      & 0.8071 & 0.4769 & 25.07 & 0.8092 & 0.3727 & 22.02 & 0.6759 & 0.5704 & 22.96 & 0.8925 & 0.4499 & 23.82 & 0.7962 & 0.4675 & 23.47 \\
        & iTransformer & 0.8380 & 0.3736 & 17.55 & 0.8446 & 0.2721 & 16.22 & 0.6664 & 0.6856 & 26.00 & 0.9230 & 0.2915 & 14.52 & 0.8180 & 0.4057 & 18.57 \\
        & PatchTST     & 0.8248 & 0.3993 & 19.30 & 0.8318 & 0.2802 & 16.91 & 0.7068 & 0.6078 & 23.25 & 0.8796 & 0.3592 & 17.91 & 0.8107 & 0.4116 & 19.34 \\
        & TimeXer      & 0.8237 & 0.4071 & 20.04 & 0.8471 & 0.2669 & 15.50 & 0.7638 & 0.5235 & 20.92 & 0.9021 & 0.3277 & 16.36 & 0.8342 & 0.3813 & 18.21 \\
        & Samformer    & 0.8086 & 0.4116 & 19.72 & 0.8045 & 0.3049 & 18.05 & 0.7880 & 0.5042 & 19.90 & 0.8663 & 0.3893 & 19.91 & 0.8169 & 0.4025 & 19.39 \\
        & ModernTCN    & 0.9263 & 0.2615 & 13.08 & 0.8886 & \underline{0.2482} & \underline{14.86} & \underline{0.8706} & 0.4745 & 20.37 & \textbf{0.9770} & \textbf{0.1754} & \textbf{8.37} & 0.9156 & 0.2899 & 14.17 \\
        & Crossformer  & \textbf{0.9305} & \underline{0.2556} & \textbf{12.21} & \textbf{0.9524} & 0.2874 & 14.89 & 0.8213 & 0.4524 & 17.08 & 0.9435 & 0.2640 & 13.58 & 0.9119 & 0.3149 & 14.44 \\
        & DeformTime   & 0.9161 & \textbf{0.2437} & \underline{12.46} & \underline{0.9356} & \textbf{0.2364} & \textbf{11.69} & \textbf{0.8744} & \underline{0.3664} & \underline{13.61} & 0.9675 & 0.2023 & 11.29 & \textbf{0.9234} & \textbf{0.2622} & \textbf{12.26} \\
        & Sonnet       & \underline{0.9301} & 0.2668 & 12.94 & 0.9293 & 0.2692 & 16.42 & 0.8481 & \textbf{0.3376} & \textbf{11.89} & \underline{0.9693} & \underline{0.1935} & \underline{10.67} & \underline{0.9192} & \underline{0.2668} & \underline{12.98} \\
    \midrule
    \multirow{9}{*}{14} & DLinear      & 0.7323 & 0.5767 & 29.99 & 0.7371 & 0.4384 & 25.44 & 0.6105 & 0.6635 & 27.69 & 0.8216 & 0.5085 & 26.27 & 0.7254 & 0.5467 & 27.35 \\
        & iTransformer & 0.7837 & 0.4310 & 20.63 & 0.7629 & 0.3448 & 20.49 & 0.6138 & 0.6853 & 27.21 & 0.8495 & 0.4196 & 21.41 & 0.7525 & 0.4702 & 22.44 \\
        & PatchTST     & 0.7626 & 0.4810 & 23.33 & 0.7411 & 0.3524 & 21.48 & 0.6246 & 0.7164 & 28.11 & 0.8102 & 0.4583 & 23.44 & 0.7346 & 0.5020 & 24.09 \\
        & TimeXer      & 0.7423 & 0.4795 & 23.26 & 0.7872 & 0.3078 & 18.07 & 0.6443 & 0.6440 & 25.49 & 0.8261 & 0.4345 & 21.76 & 0.7500 & 0.4665 & 22.14 \\
        & Samformer    & 0.7297 & 0.4972 & 23.71 & 0.7555 & 0.3416 & 20.14 & 0.6032 & 0.7957 & 30.46 & 0.8117 & 0.4683 & 23.68 & 0.7250 & 0.5257 & 24.50 \\
        & ModernTCN    & \textbf{0.9250} & 0.3582 & 15.44 & 0.8543 & 0.3175 & 17.00 & \underline{0.8696} & 0.4244 & 15.32 & 0.9320 & 0.2666 & 13.40 & 0.8952 & 0.3417 & 15.29 \\
        & Crossformer  & 0.8787 & 0.3333 & 16.05 & 0.8965 & 0.2980 & 16.74 & 0.8339 & 0.4325 & 16.91 & 0.9169 & 0.3647 & 19.21 & 0.8815 & 0.3571 & 17.23 \\
        & DeformTime   & 0.9033 & \underline{0.2962} & \underline{14.39} & \textbf{0.9374} & \underline{0.2893} & \textbf{13.15} & 0.8490 & \underline{0.3897} & \underline{14.62} & \underline{0.9378} & \underline{0.2584} & \underline{13.05} & \underline{0.9069} & \underline{0.3084} & \underline{13.80} \\
        & Sonnet       & \underline{0.9244} & \textbf{0.2807} & \textbf{12.91} & \underline{0.9049} & \textbf{0.2832} & \underline{14.71} & \textbf{0.8771} & \textbf{0.3490} & \textbf{13.24} & \textbf{0.9615} & \textbf{0.2096} & \textbf{11.51} & \textbf{0.9170} & \textbf{0.2806} & \textbf{13.10} \\
    \midrule
    \multirow{9}{*}{21} & DLinear      & 0.6687 & 0.6032 & 30.66 & 0.6920 & 0.4960 & 28.49 & 0.5830 & 0.7252 & 30.15 & 0.7778 & 0.5760 & 29.33 & 0.6804 & 0.6001 & 29.66 \\
        & iTransformer & 0.8659 & 0.3645 & 17.75 & 0.7078 & 0.3702 & 21.73 & 0.4561 & 0.8642 & 33.94 & 0.8298 & 0.4433 & 23.01 & 0.7149 & 0.5106 & 24.11 \\
        & PatchTST     & 0.6899 & 0.5705 & 28.41 & 0.6399 & 0.4255 & 26.14 & 0.5214 & 0.7963 & 32.80 & 0.7137 & 0.5815 & 30.27 & 0.6412 & 0.5935 & 29.40 \\
        & TimeXer      & 0.6203 & 0.5899 & 29.37 & 0.6373 & 0.4132 & 24.19 & 0.5158 & 0.7825 & 31.10 & 0.7460 & 0.5006 & 25.04 & 0.6298 & 0.5715 & 27.43 \\
        & Samformer    & 0.7079 & 0.5086 & 24.07 & 0.8118 & 0.3403 & 18.93 & 0.5685 & 0.7794 & 27.50 & 0.7169 & 0.5378 & 26.37 & 0.7013 & 0.5415 & 24.22 \\
        & ModernTCN    & \underline{0.8695} & \underline{0.3118} & \textbf{12.39} & 0.8775 & 0.3695 & 16.88 & 0.7572 & 0.5057 & 18.36 & 0.8997 & 0.2968 & \underline{14.08} & 0.8510 & 0.3710 & 15.43 \\
        & Crossformer  & \textbf{0.9565} & 0.3267 & 15.49 & 0.8669 & \underline{0.3168} & 15.95 & \underline{0.8777} & \underline{0.3964} & 15.11 & 0.9135 & 0.3273 & 17.07 & \textbf{0.9036} & 0.3418 & 15.90 \\
        & DeformTime   & 0.8331 & 0.3543 & 15.62 & \underline{0.8966} & \textbf{0.3039} & \underline{14.99} & \textbf{0.9141} & \textbf{0.3536} & \textbf{13.06} & \underline{0.9249} & \underline{0.2598} & \textbf{13.27} & 0.8922 & \underline{0.3179} & \underline{14.23} \\
        & Sonnet       & 0.8573 & \textbf{0.2888} & \underline{13.12} & \textbf{0.9619} & 0.3270 & \textbf{14.18} & 0.8359 & 0.4013 & \underline{14.37} & \textbf{0.9382} & \textbf{0.2545} & 14.75 & \underline{0.8983} & \textbf{0.3179} & \textbf{14.11} \\
    \midrule
    \multirow{9}{*}{28} & DLinear      & 0.6083 & 0.6496 & 32.81 & 0.6016 & 0.5329 & 30.18 & 0.5506 & 0.7893 & 32.75 & 0.7438 & 0.6539 & 32.89 & 0.6261 & 0.6564 & 32.16 \\
        & iTransformer & 0.6984 & 0.5531 & 27.80 & 0.5779 & 0.4444 & 25.70 & 0.3211 & 0.9942 & 40.08 & 0.6797 & 0.6077 & 30.60 & 0.5693 & 0.6498 & 31.05 \\
        & PatchTST     & 0.6067 & 0.6457 & 32.75 & 0.5192 & 0.4903 & 30.18 & 0.4424 & 0.8809 & 36.70 & 0.6555 & 0.6492 & 33.78 & 0.5560 & 0.6665 & 33.35 \\
        & TimeXer      & 0.5345 & 0.6372 & 32.39 & 0.6320 & 0.4302 & 23.80 & 0.3998 & 0.9296 & 37.46 & 0.6127 & 0.6248 & 31.64 & 0.5448 & 0.6555 & 31.32 \\
        & Samformer    & 0.6543 & 0.5652 & 27.78 & 0.6646 & 0.3909 & 22.79 & 0.4984 & 0.8367 & 30.81 & 0.5980 & 0.6273 & 30.40 & 0.6038 & 0.6050 & 27.95 \\
        & ModernTCN    & 0.8552 & 0.3611 & 17.38 & \underline{0.9076} & 0.3748 & 17.59 & 0.8068 & 0.5156 & 18.99 & 0.8645 & \underline{0.3245} & \textbf{14.79} & 0.8585 & 0.3940 & 17.19 \\
        & Crossformer  & \textbf{0.8969} & \underline{0.3398} & 16.47 & 0.8346 & \underline{0.3647} & 16.99 & 0.8514 & \underline{0.4365} & \textbf{15.42} & \underline{0.8923} & 0.3578 & 16.87 & 0.8688 & 0.3747 & \underline{16.44} \\
        & DeformTime   & 0.8888 & 0.3718 & \underline{15.55} & 0.9046 & \textbf{0.3153} & \textbf{14.57} & \textbf{0.9037} & \textbf{0.4185} & \underline{16.18} & \textbf{0.9260} & \textbf{0.3069} & \underline{16.68} & \textbf{0.9058} & \textbf{0.3532} & \textbf{15.75} \\
        & Sonnet       & \underline{0.8947} & \textbf{0.3072} & \textbf{15.45} & \textbf{0.9288} & 0.3723 & \underline{16.09} & \underline{0.8850} & 0.4497 & 16.70 & 0.8904 & 0.3406 & 17.90 & \underline{0.8997} & \underline{0.3675} & 16.53 \\
\bottomrule
\end{tabular}
% }
\caption{\ILITableCaptionAppendix{US Region 9}{ILI-US9}}
\label{tab:results_ili_us9}
\end{table*}

\section{Further information about MVCA}
\label{appsec:further_details}

This section supplements Section~\ref{subsec:MVCA} of the main paper. Here, we further clarify complex number operations.

In Equation~\ref{eq:coherence_density}, the cross-spectral and power-spectral density matrices are obtained using the frequency domain embeddings $\mathbf{Q}_f$, $\mathbf{K}_f \in \mathbb{C}^{L \times \ell}$. The obtained matrices $\mathbf{P}_{qq}=\mathbf{Q}_{f} \odot \mathbf{Q}_{f}^\ast$ and $\mathbf{P}_{kk}=\mathbf{K}_{f} \odot \mathbf{K}_{f}^\ast$ are both containing real numbers because multiplying a complex number (\eg $z = a + ib$) by its conjugate ($a - ib$) results in its squared magnitude ($a^2 + b^2$), which is a real number.

In Equation~\ref{eq:coherence}, we obtain the coherence vector $\mathbf{C}_{qk}$ using the averaged cross-spectral density matrix $\mathbf{P}_{qk} \in \mathbf{C}^{L \times \ell}$. The average of this imaginary matrix $\mathbf{P}_{qk}$ is obtained by averaging the real and imaginary components separately. Taking the magnitude $|\overline{\mathbf{P}}_{qk}|$ then results in a real-valued vector, as magnitude is essentially the Euclidean norm (\ie, the square root of the sum of the squares of the real and imaginary parts) of each complex entry. As such, we obtain a real-valued coherence vector $\mathbf{C}_{qk}$.

\section{Baseline forecasting models and attention mechanisms}
\label{appsec:baseline_forecasting}
In this section, we first provide a brief description of the baseline models that we use in our experiments (Section~\ref{appsubsec:baselines_mts}). To the best of our knowledge, they form the current state-of-the-art (SOTA) in MTS forecasting. These also include the base models that use na\"ive attention mechanisms, which we then replace with MVCA (Section~\ref{subsec:forecasting_results_vca}). We also describe the attention modules that have been used in related literature and which we compare to MVCA (Section~\ref{appsubsec:select_attn}).

\subsection{Baseline MTS forecasting models}
\label{appsubsec:baselines_mts}

\textbf{DeformTime}~\citep{shu2025deformtime} is specifically designed for multivariable forecasting. It uses deformable attention to capture the dependencies across time and covariates, reaching competitive performance on MTS forecasting tasks. Based on our evaluation, DeformTime is the best-performing baseline.

\noindent\textbf{ModernTCN}~\citep{luo2024moderntcn} applies convolution along the temporal and variable axes. By leveraging large receptive fields, it effectively captures both intra-variable and inter-variable dependencies.

\noindent\textbf{TimeXer}~\citep{wang2024timexer} is a transformer-based model that focuses on multivariable forecasting. It first embeds the exogenous variables along the sequence dimension, and then captures the exogenous information through a cross-attention mechanism.

\noindent\textbf{Samformer}~\citep{ilbert2024samformer} first embeds the input along the sequence dimension and then applies na\"ive attention along the feature dimension. It conducts forecasts using a linear layer. We also include Samformer as one of the base models that deploy na\"ive attention.

\noindent\textbf{iTransformer}~\citep{liu2024itransformer} embeds the input along the sequence dimension and deploys multiple attention layers along the variable dimension. We also include iTransformer as one of the base models.

\begin{table*}[!t]
\renewcommand{\arraystretch}{0.92}
\centering
\small
\setlength{\tabcolsep}{5.5pt}
\setlength{\aboverulesep}{0.5pt}
\setlength{\belowrulesep}{0.5pt}
% \resizebox{\linewidth}{!}{
\begin{tabular}{cl ccc ccc ccc ccc}
\toprule
\multirow{2}{*}{\textbf{H}} & \multirow{2}{*}{\textbf{Model}} & \multicolumn{3}{c}{\textbf{2016}} & \multicolumn{3}{c}{\textbf{2017}} & \multicolumn{3}{c}{\textbf{2018}} & \multicolumn{3}{c}{\textbf{Average}} \\
 & & $r$ & MAE & $\epsilon\%$ & $r$ & MAE & $\epsilon\%$ & $r$ & MAE & $\epsilon\%$ & $r$ & MAE & $\epsilon\%$ \\
\cmidrule{1-1}\cmidrule{2-2}\cmidrule(lr){3-5} \cmidrule(lr){6-8} \cmidrule(lr){9-11} \cmidrule(lr){12-14}
    \multirow{9}{*}{4} & DLinear      & 0.8341 & 2.3772 & 15.26 & 0.7730 & 2.5122 & 16.47 & 0.8422 & 2.6301 & 31.10 & 0.8164 & 2.5065 & 20.94 \\
        & iTransformer & 0.8746 & 2.0626 & 13.39 & 0.8325 & 2.1816 & 14.51 & 0.8803 & 2.2087 & 27.76 & 0.8625 & 2.1509 & 18.55 \\
        & PatchTST     & 0.7975 & 2.6606 & 17.02 & 0.7233 & 2.7843 & 18.32 & 0.8077 & 2.8359 & 33.44 & 0.7762 & 2.7602 & 22.93 \\
        & TimeXer      & 0.8707 & 2.1636 & 14.07 & 0.8256 & 2.2316 & 14.69 & 0.8681 & 2.3932 & 29.00 & 0.8548 & 2.2628 & 19.25 \\
        & Samformer    & 0.8705 & 2.0649 & 13.40 & 0.8393 & 2.1271 & 14.15 & 0.8752 & 2.2692 & 28.04 & 0.8617 & 2.1537 & 18.53 \\
        & ModernTCN    & 0.8972 & 1.8388 & 11.88 & 0.8560 & 2.0076 & 13.33 & 0.9063 & 1.9904 & 25.41 & 0.8865 & 1.9456 & 16.88 \\
        & Crossformer  & \underline{0.9186} & \textbf{1.6316} & \textbf{10.67} & \underline{0.8788} & \underline{1.8383} & \underline{12.14} & \underline{0.9225} & \underline{1.7641} & \underline{22.97} & \underline{0.9066} & \underline{1.7447} & \underline{15.26} \\
        & DeformTime   & 0.9060 & 1.7647 & 11.64 & 0.8620 & 1.9632 & 12.78 & 0.9096 & 1.8982 & 24.50 & 0.8926 & 1.8753 & 16.31 \\
        & Sonnet       & \textbf{0.9206} & \underline{1.6571} & \underline{10.95} & \textbf{0.8887} & \textbf{1.7880} & \textbf{11.85} & \textbf{0.9269} & \textbf{1.7242} & \textbf{22.67} & \textbf{0.9121} & \textbf{1.7231} & \textbf{15.16} \\
    \midrule
    \multirow{9}{*}{12} & DLinear      & 0.6875 & 3.2620 & 20.14 & 0.5777 & 3.2740 & 20.77 & 0.6782 & 3.6422 & 39.20 & 0.6478 & 3.3927 & 26.70 \\
        & iTransformer & 0.7073 & 3.1987 & 20.42 & 0.6167 & 3.2760 & 21.05 & 0.6916 & 3.6120 & 40.61 & 0.6719 & 3.3622 & 27.36 \\
        & PatchTST     & 0.6801 & 3.4350 & 21.80 & 0.5729 & 3.4499 & 22.13 & 0.6639 & 3.7370 & 41.06 & 0.6390 & 3.5406 & 28.33 \\
        & TimeXer      & 0.7286 & 3.0578 & 19.05 & 0.6602 & 3.0519 & 19.57 & 0.7360 & 3.3778 & 37.11 & 0.7082 & 3.1625 & 25.25 \\
        & Samformer    & 0.7117 & 3.1711 & 20.14 & 0.6352 & 3.1860 & 20.37 & 0.6969 & 3.5638 & 39.31 & 0.6812 & 3.3070 & 26.61 \\
        & ModernTCN    & 0.7266 & 3.0716 & 19.39 & 0.6476 & 3.1372 & 19.95 & 0.7140 & 3.4080 & 37.39 & 0.6961 & 3.2056 & 25.58 \\
        & Crossformer  & \underline{0.7530} & \underline{2.9360} & 18.52 & 0.6326 & 3.1054 & 19.77 & \underline{0.7626} & \underline{3.1061} & \textbf{34.75} & 0.7161 & 3.0492 & 24.35 \\
        & DeformTime   & 0.7403 & 2.9613 & \underline{18.51} & \underline{0.6810} & \underline{2.9529} & \underline{18.83} & 0.7600 & 3.1499 & 35.10 & \underline{0.7271} & \underline{3.0214} & \underline{24.15} \\
        & Sonnet       & \textbf{0.7648} & \textbf{2.8561} & \textbf{17.98} & \textbf{0.6937} & \textbf{2.9154} & \textbf{18.69} & \textbf{0.7652} & \textbf{3.1051} & \underline{34.98} & \textbf{0.7413} & \textbf{2.9589} & \textbf{23.88} \\
    \midrule
    \multirow{9}{*}{28} & DLinear      & 0.6197 & 3.5890 & 21.92 & 0.5479 & 3.3473 & 21.14 & 0.6560 & 3.8855 & 41.41 & 0.6078 & 3.6073 & 28.16 \\
        & iTransformer & 0.6016 & 3.7784 & 23.47 & 0.5710 & 3.4687 & 22.17 & 0.6506 & 3.8182 & 41.77 & 0.6077 & 3.6884 & 29.14 \\
        & PatchTST     & 0.6130 & 3.7670 & 23.35 & 0.5358 & 3.6196 & 23.20 & 0.6436 & 3.8229 & 41.71 & 0.5974 & 3.7365 & 29.42 \\
        & TimeXer      & 0.6561 & 3.5155 & 21.56 & \underline{0.6287} & 3.2048 & 20.43 & 0.6842 & 3.6813 & 39.69 & 0.6563 & 3.4672 & 27.23 \\
        & Samformer    & 0.6239 & 3.6455 & 22.48 & 0.5648 & 3.4061 & 21.46 & 0.6594 & 3.7008 & 39.99 & 0.6160 & 3.5841 & 27.97 \\
        & ModernTCN    & 0.6509 & 3.4012 & 20.79 & 0.5680 & 3.4846 & 22.16 & 0.6841 & 3.6345 & 39.48 & 0.6343 & 3.5067 & 27.48 \\
        & Crossformer  & 0.6791 & 3.2933 & 20.29 & 0.6278 & \textbf{3.1441} & \textbf{19.89} & 0.7001 & 3.4770 & \underline{37.46} & 0.6690 & 3.3048 & 25.88 \\
        & DeformTime   & \underline{0.7076} & \underline{3.2300} & \underline{19.96} & \textbf{0.6311} & \underline{3.1675} & \underline{20.08} & \underline{0.7298} & \underline{3.4197} & 37.48 & \textbf{0.6895} & \underline{3.2724} & \underline{25.84} \\
        & Sonnet       & \textbf{0.7171} & \textbf{3.0995} & \textbf{19.21} & 0.5982 & 3.2051 & 20.27 & \textbf{0.7307} & \textbf{3.3438} & \textbf{37.00} & \underline{0.6820} & \textbf{3.2161} & \textbf{25.49} \\
    \midrule
    \multirow{9}{*}{120} & DLinear      & 0.5178 & 4.0071 & 24.30 & 0.3307 & 3.7275 & 23.27 & 0.6050 & 4.1572 & 43.48 & 0.4845 & 3.9640 & 30.35 \\
        & iTransformer & 0.5382 & 3.9632 & 23.80 & 0.5504 & 3.4798 & 22.72 & 0.5855 & 4.1123 & 44.57 & 0.5581 & 3.8518 & 30.36 \\
        & PatchTST     & 0.3849 & 4.5359 & 27.91 & 0.3942 & 3.9432 & 25.12 & 0.5567 & 4.2240 & 44.16 & 0.4453 & 4.2344 & 32.40 \\
        & TimeXer      & 0.6451 & 3.4743 & 21.42 & 0.4839 & 3.5218 & 22.15 & 0.6744 & 3.9711 & 42.00 & 0.6011 & 3.6557 & 28.52 \\
        & Samformer    & 0.5761 & 3.8839 & 24.10 & 0.4873 & 3.6875 & 23.55 & 0.6481 & 3.9545 & 43.31 & 0.5705 & 3.8420 & 30.32 \\
        & ModernTCN    & 0.5646 & 3.7511 & 22.77 & 0.5422 & 3.5645 & 23.40 & 0.5976 & 4.2146 & 43.60 & 0.5682 & 3.8434 & 29.92 \\
        & Crossformer  & 0.6996 & 3.3054 & 20.39 & \textbf{0.6396} & \textbf{3.0820} & \textbf{19.72} & 0.6691 & 3.7933 & 40.13 & 0.6694 & 3.3935 & 26.75 \\
        & DeformTime   & \underline{0.7277} & \underline{3.1270} & \underline{19.63} & 0.6141 & 3.1808 & 20.31 & \textbf{0.7058} & \underline{3.5842} & \underline{38.58} & \underline{0.6826} & \underline{3.2973} & \underline{26.17} \\
        & Sonnet       & \textbf{0.7281} & \textbf{3.1081} & \textbf{19.38} & \underline{0.6365} & \underline{3.1281} & \underline{20.03} & \underline{0.6899} & \textbf{3.5029} & \textbf{38.04} & \textbf{0.6849} & \textbf{3.2464} & \textbf{25.82} \\
\bottomrule
\end{tabular}
% }
\caption{\WEATableCaptionAppendix{London}{WEA-LD}}
\label{tab:results_wea_ld}
\end{table*}

\noindent\textbf{PatchTST}~\citep{nie2023time} segments time series data into patches of fixed size, treating them as input tokens to a transformer. It processes each variable independently, enabling efficient modelling of long-term dependencies. PatchTST is also used as one of the base models.

\noindent\textbf{Crossformer}~\citep{zhang2023crossformer} captures both inter- and intra-variable dependencies using a two-stage attention mechanism. Each stage is dedicated to one dependency type. The information is captured using time series patches.

\noindent\textbf{DLinear}~\citep{zeng2023are} uses a simpler neural network architecture (feedforward layers). It decomposes each input series into trend and seasonal components, which are then forecasted independently using linear layers. Each variable is modelled separately, using its own historical values for prediction. Based on our evaluation, DLinear is the worst-performing model in MTS forecasting. It does not outperform persistence in most experiments (Table~\ref{tab:results_averaged}). This corroborates findings in related work~\citep{shu2025deformtime}.

\noindent\textbf{Na\"ive forecasting models.} We include simple baseline models as reference points for performance comparison. We use either seasonal persistence or standard persistence, depending on whether a target variable (and task) is characterised by strong seasonality or not. A standard persistence model (na\"ive baseline) uses the last observed value of the target variable as the prediction (forecasting outcome). The seasonal persistence model uses the value from the same point in the previous seasonal cycle. In our case, the electricity consumption data (ELEC) has a strong seasonal trend, with seasonal persistence outperforming the standard one by a large margin (results not shown). Therefore, we use the seasonal persistence model for the ELEC task, while for the other tasks, we retain standard persistence.

\begin{table*}[!t]
\renewcommand{\arraystretch}{0.92}
\centering
\small
\setlength{\tabcolsep}{5.5pt}
\setlength{\aboverulesep}{0.5pt}
\setlength{\belowrulesep}{0.5pt}
% \resizebox{\linewidth}{!}{
\begin{tabular}{cl ccc ccc ccc ccc}
\toprule
\multirow{2}{*}{\textbf{H}} & \multirow{2}{*}{\textbf{Model}} & \multicolumn{3}{c}{\textbf{2016}} & \multicolumn{3}{c}{\textbf{2017}} & \multicolumn{3}{c}{\textbf{2018}} & \multicolumn{3}{c}{\textbf{Average}} \\
 & & $r$ & MAE & $\epsilon\%$ & $r$ & MAE & $\epsilon\%$ & $r$ & MAE & $\epsilon\%$ & $r$ & MAE & $\epsilon\%$ \\
\cmidrule{1-1}\cmidrule{2-2}\cmidrule(lr){3-5} \cmidrule(lr){6-8} \cmidrule(lr){9-11} \cmidrule(lr){12-14}
    \multirow{9}{*}{4} & DLinear      & 0.9285 & 1.9481 & 16.49 & 0.9182 & 2.1349 & 21.52 & 0.9320 & 1.8516 & 15.65 & 0.9262 & 1.9782 & 17.88 \\
        & iTransformer & 0.9476 & 1.5907 & 12.82 & 0.9412 & 1.7162 & 17.87 & 0.9488 & 1.5129 & 12.53 & 0.9459 & 1.6066 & 14.40 \\
        & PatchTST     & 0.9038 & 2.1592 & 17.66 & 0.8927 & 2.3817 & 23.79 & 0.9148 & 1.9524 & 16.26 & 0.9038 & 2.1644 & 19.24 \\
        & TimeXer      & 0.9402 & 1.7540 & 14.49 & 0.9382 & 1.8287 & 18.59 & 0.9460 & 1.6045 & 13.63 & 0.9415 & 1.7290 & 15.57 \\
        & Samformer    & 0.9478 & 1.5838 & 13.06 & 0.9415 & 1.7441 & 17.92 & 0.9520 & 1.4729 & 12.35 & 0.9471 & 1.6003 & 14.44 \\
        & ModernTCN    & 0.9607 & 1.3723 & 11.37 & 0.9539 & 1.5332 & 15.92 & 0.9614 & 1.3408 & 11.57 & 0.9587 & 1.4154 & 12.95 \\
        & Crossformer  & \underline{0.9676} & \textbf{1.2417} & \textbf{10.31} & \underline{0.9631} & \underline{1.3846} & \underline{15.29} & \underline{0.9687} & \underline{1.2541} & \underline{11.14} & \underline{0.9665} & \underline{1.2935} & \underline{12.24} \\
        & DeformTime   & 0.9580 & 1.4031 & 11.30 & 0.9592 & 1.4651 & 15.92 & 0.9638 & 1.3403 & 11.85 & 0.9603 & 1.4028 & 13.03 \\
        & Sonnet       & \textbf{0.9677} & \underline{1.2506} & \underline{10.43} & \textbf{0.9645} & \textbf{1.3370} & \textbf{14.70} & \textbf{0.9691} & \textbf{1.2271} & \textbf{10.69} & \textbf{0.9671} & \textbf{1.2716} & \textbf{11.94} \\
    \midrule
    \multirow{9}{*}{12} & DLinear      & 0.8464 & 2.8374 & 22.86 & 0.8060 & 3.1841 & 29.99 & 0.8487 & 2.8307 & 23.01 & 0.8337 & 2.9507 & 25.29 \\
        & iTransformer & 0.8626 & 2.5160 & 20.10 & 0.8275 & 2.9724 & 29.06 & 0.8638 & 2.4942 & 20.12 & 0.8513 & 2.6609 & 23.09 \\
        & PatchTST     & 0.8405 & 2.7056 & 21.91 & 0.7976 & 3.2096 & 30.96 & 0.8471 & 2.6623 & 21.56 & 0.8284 & 2.8592 & 24.81 \\
        & TimeXer      & 0.8686 & 2.4893 & 19.91 & 0.8404 & 2.9183 & 27.85 & 0.8744 & 2.5534 & 20.79 & 0.8611 & 2.6537 & 22.85 \\
        & Samformer    & 0.8646 & 2.5356 & 20.66 & 0.8293 & 2.9936 & 29.05 & 0.8594 & 2.5914 & 20.98 & 0.8511 & 2.7069 & 23.57 \\
        & ModernTCN    & 0.8677 & 2.4940 & 20.69 & 0.8332 & 2.9126 & 28.23 & 0.8689 & 2.4598 & 19.87 & 0.8566 & 2.6221 & 22.93 \\
        & Crossformer  & \textbf{0.8893} & \textbf{2.3005} & \textbf{18.45} & \textbf{0.8570} & 2.7562 & 26.36 & \textbf{0.8916} & \textbf{2.2916} & \underline{18.52} & \textbf{0.8793} & 2.4494 & \underline{21.11} \\
        & DeformTime   & 0.8852 & \underline{2.3050} & \underline{18.55} & \underline{0.8519} & \textbf{2.7148} & \textbf{26.08} & 0.8864 & 2.3162 & \textbf{18.48} & 0.8745 & \textbf{2.4453} & \textbf{21.04} \\
        & Sonnet       & \underline{0.8858} & 2.3262 & 18.74 & 0.8499 & \underline{2.7186} & \underline{26.25} & \underline{0.8902} & \underline{2.2980} & 18.71 & \underline{0.8753} & \underline{2.4476} & 21.23 \\
    \midrule
    \multirow{9}{*}{28} & DLinear      & 0.8127 & 3.1132 & 24.71 & 0.7533 & 3.4967 & 32.30 & 0.8217 & 3.0198 & 24.19 & 0.7959 & 3.2099 & 27.07 \\
        & iTransformer & 0.8227 & 2.8172 & 21.76 & 0.7710 & 3.5240 & 32.26 & 0.8387 & 2.7201 & 21.89 & 0.8108 & 3.0204 & 25.30 \\
        & PatchTST     & 0.8167 & 2.9402 & 23.82 & 0.7579 & 3.4486 & 33.28 & 0.8256 & 2.8982 & 23.92 & 0.8000 & 3.0956 & 27.01 \\
        & TimeXer      & 0.8333 & 2.8010 & 22.43 & 0.8052 & 3.2027 & 30.03 & 0.8541 & \underline{2.6288} & 20.97 & 0.8309 & 2.8775 & 24.48 \\
        & Samformer    & 0.8221 & 2.8725 & 23.07 & 0.7663 & 3.3700 & 32.19 & 0.8285 & 2.8616 & 23.29 & 0.8056 & 3.0347 & 26.18 \\
        & ModernTCN    & 0.8365 & 2.7179 & 21.53 & 0.7673 & 3.3990 & 32.94 & 0.8474 & 2.6840 & 21.66 & 0.8170 & 2.9336 & 25.37 \\
        & Crossformer  & \underline{0.8602} & \underline{2.5428} & \underline{20.23} & 0.8107 & 3.1558 & 29.80 & \underline{0.8563} & 2.6502 & 20.96 & 0.8424 & 2.7830 & 23.66 \\
        & DeformTime   & \textbf{0.8636} & \textbf{2.5246} & \textbf{19.87} & \underline{0.8166} & \underline{3.0717} & \textbf{28.77} & 0.8548 & 2.6388 & \underline{20.96} & \underline{0.8450} & \underline{2.7450} & \underline{23.20} \\
        & Sonnet       & 0.8545 & 2.5533 & 20.23 & \textbf{0.8342} & \textbf{2.9812} & \underline{29.08} & \textbf{0.8719} & \textbf{2.4887} & \textbf{19.99} & \textbf{0.8535} & \textbf{2.6744} & \textbf{23.10} \\
    \midrule
    \multirow{9}{*}{120} & DLinear      & 0.7883 & 3.4169 & 26.88 & 0.6870 & 3.9611 & 34.99 & 0.7596 & 3.4606 & 27.39 & 0.7450 & 3.6129 & 29.75 \\
        & iTransformer & 0.8418 & 2.7846 & 22.62 & 0.7691 & 3.5547 & 37.41 & 0.8123 & 2.9695 & 23.20 & 0.8078 & 3.1029 & 27.74 \\
        & PatchTST     & 0.7964 & 3.1199 & 24.02 & 0.7451 & 3.6650 & 34.57 & 0.7857 & 3.4409 & 27.76 & 0.7757 & 3.4086 & 28.78 \\
        & TimeXer      & 0.8198 & 2.9056 & 23.44 & 0.8003 & 3.2102 & 30.24 & 0.7830 & 3.3346 & 25.97 & 0.8010 & 3.1501 & 26.55 \\
        & Samformer    & 0.7795 & 3.3766 & 28.57 & 0.7268 & 3.9559 & 42.16 & 0.7942 & 3.2543 & 28.30 & 0.7668 & 3.5289 & 33.01 \\
        & ModernTCN    & 0.7883 & 3.1432 & 24.11 & 0.7353 & 3.5156 & 34.01 & 0.7781 & 3.2413 & 26.02 & 0.7672 & 3.3000 & 28.05 \\
        & Crossformer  & 0.8382 & 2.7770 & 21.74 & \underline{0.8143} & \underline{3.1789} & \underline{29.76} & 0.8129 & 2.9286 & 22.59 & 0.8218 & 2.9615 & 24.69 \\
        & DeformTime   & \textbf{0.8642} & \textbf{2.5570} & \textbf{20.65} & 0.8013 & 3.2431 & 30.09 & \underline{0.8529} & \underline{2.6672} & \underline{20.85} & \underline{0.8395} & \underline{2.8224} & \underline{23.86} \\
        & Sonnet       & \underline{0.8614} & \underline{2.5606} & \underline{20.76} & \textbf{0.8377} & \textbf{2.9614} & \textbf{28.01} & \textbf{0.8597} & \textbf{2.6185} & \textbf{20.68} & \textbf{0.8529} & \textbf{2.7135} & \textbf{23.15} \\
\bottomrule
\end{tabular}
% }
\caption{\WEATableCaptionAppendix{New York}{WEA-NY}}
\label{tab:results_wea_ny}
\end{table*}

\subsection{Additional information on adapting baseline models for Multivariable forecasting tasks}

Amongst the selected baselines, DeformTime~\citep{shu2025deformtime} and TimeXer~\citep{wang2024timexer} are specifically designed for multivariable forecasting tasks. For the rest of the models that were designed for multivariate forecasting, we follow the adaptation strategy from DeformTime and TimeXer, which conduct multivariate forecasting and evaluate the model performance on the target variable. Given that DeformTime and TimeXer have demonstrated their effectiveness across a wide range of forecasting benchmarks, we believe this evaluation provides a fair and consistent comparison against SOTA multivariate forecasting models.

\subsection{Attention mechanisms used for time series forecasting}
\label{appsubsec:select_attn}

\textbf{Variable Deformable Attention Block (VDAB).} Introduced in DeformTime~\citep{shu2025deformtime}, VDAB adaptively attends to relevant information across time and variables by applying cross attention over the original and deformed input. This enables to capture complex inter-variable dependencies in MTS forecasting.

\noindent\textbf{Frequency Enhanced Decomposition (FED).} Employed in FEDformer~\citep{zhou2022fedformer}, FED decomposes time series into trend and seasonal components, applying attention mechanisms to each, thereby capturing both periodic patterns and time-delay dependencies effectively. We note that we do not compare to the complete forecasting model of FEDformer as it has been outperformed by most of our selected baseline models. Crossformer, iTransformer, PatchTST, ModernTCN, DLinear, and Samformer have all provided direct comparison with FEDformer in their original papers. Meanwhile, the ones that do not provide a direct comparison with FEDformer, namely TimeXer and DeformTime, have both outperformed PatchTST and iTransformer on MTS forecasting tasks by a large margin. Given Sonnet's competitive performance over all the aforementioned (baseline) models, we expect it to yield superior performance compared to FEDformer.

\noindent\textbf{FNet.} The attention proposed in FNet~\citep{lee-thorp2022fnet} replaces the commonly used self-attention mechanism in a transformer with a parameter-free Fourier transform. Specifically, it maps the input sequence to the frequency domain and retains only the real part of the result. This approach enables effective mixing between input tokens along the time dimension without adding learnable parameters. FNet is not included as a baseline model in the main results as it is not a model designed for forecasting purposes. Additionally, we show in Sections~\ref{subsec:forecasting_results_vca} and~\ref{subsec:Sonnet_attention_effect} that it gives a similar performance to the vanilla transformer, which has been outperformed.

\noindent\textbf{Na\"ive Attention.} The commonly used attention module was first proposed by~\citet{vaswani2017attention} as a core component of the transformer model. The attention weights are computed using scaled dot product operations between query and key embeddings, facilitating the model's ability to focus on relevant parts of the input sequence. 

\begin{table*}[!t]
\renewcommand{\arraystretch}{0.92}
\centering
\small
\setlength{\tabcolsep}{5.5pt}
\setlength{\aboverulesep}{0.5pt}
\setlength{\belowrulesep}{0.5pt}
% \resizebox{\linewidth}{!}{
\begin{tabular}{cl ccc ccc ccc ccc}
\toprule
\multirow{2}{*}{\textbf{H}} & \multirow{2}{*}{\textbf{Model}} & \multicolumn{3}{c}{\textbf{2016}} & \multicolumn{3}{c}{\textbf{2017}} & \multicolumn{3}{c}{\textbf{2018}} & \multicolumn{3}{c}{\textbf{Average}} \\
 & & $r$ & MAE & $\epsilon\%$ & $r$ & MAE & $\epsilon\%$ & $r$ & MAE & $\epsilon\%$ & $r$ & MAE & $\epsilon\%$ \\
\cmidrule{1-1}\cmidrule{2-2}\cmidrule(lr){3-5} \cmidrule(lr){6-8} \cmidrule(lr){9-11} \cmidrule(lr){12-14}
    \multirow{9}{*}{4} & DLinear      & 0.8955 & 1.0544 & 7.13 & 0.9249 & 0.9475 & 5.29 & 0.9091 & 0.9675 & 6.21 & 0.9098 & 0.9898 & 6.21 \\
        & iTransformer & 0.9343 & 0.8432 & 5.80 & 0.9462 & 0.7692 & 4.25 & 0.9404 & 0.8020 & 5.20 & 0.9403 & 0.8048 & 5.08 \\
        & PatchTST     & 0.8489 & 1.2284 & 8.28 & 0.8955 & 1.0634 & 5.87 & 0.8612 & 1.1546 & 7.31 & 0.8686 & 1.1488 & 7.15 \\
        & TimeXer      & 0.9350 & 0.9353 & 6.53 & 0.9452 & 0.8258 & 4.63 & 0.9515 & 0.8333 & 5.41 & 0.9439 & 0.8648 & 5.52 \\
        & Samformer    & 0.9340 & 0.8454 & 5.82 & 0.9455 & 0.7852 & 4.32 & 0.9414 & 0.7984 & 5.15 & 0.9403 & 0.8097 & 5.10 \\
        & ModernTCN    & 0.9549 & 0.7132 & 4.93 & 0.9589 & 0.6858 & 3.77 & 0.9551 & 0.7023 & 4.57 & 0.9563 & 0.7004 & 4.42 \\
        & Crossformer  & \underline{0.9573} & \underline{0.6755} & \underline{4.67} & \underline{0.9654} & \textbf{0.6245} & \underline{3.46} & \underline{0.9628} & \underline{0.6463} & \underline{4.19} & \underline{0.9618} & \underline{0.6488} & \underline{4.11} \\
        & DeformTime   & 0.9554 & 0.7276 & 5.03 & 0.9626 & 0.6519 & 3.61 & 0.9620 & 0.6617 & 4.31 & 0.9600 & 0.6804 & 4.32 \\
        & Sonnet       & \textbf{0.9617} & \textbf{0.6666} & \textbf{4.63} & \textbf{0.9677} & \underline{0.6253} & \textbf{3.45} & \textbf{0.9673} & \textbf{0.6249} & \textbf{4.08} & \textbf{0.9656} & \textbf{0.6389} & \textbf{4.05} \\
    \midrule
    \multirow{9}{*}{12} & DLinear      & 0.7259 & 1.6528 & 11.11 & 0.8036 & 1.4281 & 7.90 & 0.7237 & 1.5581 & 9.85 & 0.7511 & 1.5464 & 9.62 \\
        & iTransformer & 0.7632 & 1.5487 & 10.49 & 0.8319 & 1.3230 & 7.29 & 0.7864 & 1.3958 & 8.80 & 0.7938 & 1.4225 & 8.86 \\
        & PatchTST     & 0.7110 & 1.6836 & 11.32 & 0.7943 & 1.4387 & 7.87 & 0.6995 & 1.6251 & 10.11 & 0.7349 & 1.5825 & 9.77 \\
        & TimeXer      & 0.7916 & 1.4408 & 9.70 & 0.8623 & 1.2177 & 6.75 & 0.8406 & 1.2495 & 7.94 & 0.8315 & 1.3027 & 8.13 \\
        & Samformer    & 0.7804 & 1.5051 & 10.18 & 0.8340 & 1.3189 & 7.28 & 0.7891 & 1.3778 & 8.65 & 0.8012 & 1.4006 & 8.70 \\
        & ModernTCN    & 0.7819 & 1.4543 & 9.81 & 0.8482 & 1.2875 & 7.12 & 0.8026 & 1.3248 & 8.35 & 0.8109 & 1.3555 & 8.43 \\
        & Crossformer  & 0.7613 & 1.4767 & 9.95 & 0.8716 & 1.1834 & 6.57 & \underline{0.8463} & \underline{1.2087} & \underline{7.73} & 0.8264 & 1.2896 & 8.08 \\
        & DeformTime   & \underline{0.8084} & \underline{1.4218} & \underline{9.50} & \underline{0.8780} & \underline{1.1443} & \underline{6.38} & 0.8351 & 1.2698 & 8.09 & \underline{0.8405} & \underline{1.2786} & \underline{7.99} \\
        & Sonnet       & \textbf{0.8114} & \textbf{1.3921} & \textbf{9.35} & \textbf{0.8820} & \textbf{1.1220} & \textbf{6.25} & \textbf{0.8548} & \textbf{1.1923} & \textbf{7.62} & \textbf{0.8494} & \textbf{1.2355} & \textbf{7.74} \\
    \midrule
    \multirow{9}{*}{28} & DLinear      & 0.6844 & 1.6762 & 11.32 & 0.7547 & 1.5545 & 8.56 & 0.6757 & 1.6389 & 10.30 & 0.7049 & 1.6232 & 10.06 \\
        & iTransformer & 0.6986 & 1.6872 & 11.39 & 0.7499 & 1.6095 & 8.79 & 0.6914 & 1.6101 & 10.07 & 0.7133 & 1.6356 & 10.08 \\
        & PatchTST     & 0.6935 & 1.6950 & 11.50 & 0.7515 & 1.5721 & 8.64 & 0.6758 & 1.6653 & 10.34 & 0.7069 & 1.6441 & 10.16 \\
        & TimeXer      & 0.7459 & 1.5904 & 10.71 & 0.8013 & 1.4330 & 7.92 & 0.7581 & 1.5111 & 9.55 & 0.7684 & 1.5115 & 9.39 \\
        & Samformer    & 0.6904 & 1.6976 & 11.49 & 0.7735 & 1.5271 & 8.43 & 0.6872 & 1.6431 & 10.25 & 0.7170 & 1.6226 & 10.05 \\
        & ModernTCN    & 0.6928 & 1.7091 & 11.51 & 0.7954 & 1.4720 & 8.06 & 0.7131 & 1.5787 & 9.89 & 0.7338 & 1.5866 & 9.82 \\
        & Crossformer  & 0.6899 & 1.7047 & 11.32 & 0.8231 & 1.4059 & 7.75 & 0.7642 & \underline{1.4739} & \underline{9.29} & 0.7590 & 1.5282 & 9.45 \\
        & DeformTime   & \textbf{0.7721} & \underline{1.5560} & \underline{10.36} & \underline{0.8288} & \underline{1.3821} & \underline{7.67} & \textbf{0.7724} & 1.4858 & 9.44 & \underline{0.7911} & \underline{1.4746} & \underline{9.16} \\
        & Sonnet       & \underline{0.7714} & \textbf{1.5099} & \textbf{10.11} & \textbf{0.8518} & \textbf{1.2950} & \textbf{7.22} & \underline{0.7718} & \textbf{1.4355} & \textbf{9.16} & \textbf{0.7983} & \textbf{1.4135} & \textbf{8.83} \\
    \midrule
    \multirow{9}{*}{120} & DLinear      & 0.6432 & 1.9457 & 13.25 & 0.6875 & 1.8548 & 10.28 & 0.7085 & 1.6983 & 10.94 & 0.6797 & 1.8329 & 11.49 \\
        & iTransformer & \underline{0.7564} & \textbf{1.6063} & \underline{11.01} & 0.7150 & 1.6955 & 9.34 & 0.6740 & 1.7218 & 10.98 & 0.7152 & 1.6745 & 10.45 \\
        & PatchTST     & 0.6846 & 2.1676 & 15.78 & 0.6216 & 2.0684 & 11.43 & 0.7092 & 1.7893 & 11.30 & 0.6718 & 2.0084 & 12.84 \\
        & TimeXer      & 0.7208 & 1.7303 & 11.64 & 0.6869 & 1.7013 & 9.33 & 0.7211 & 1.6961 & 10.69 & 0.7096 & 1.7092 & 10.55 \\
        & Samformer    & 0.6898 & 1.9304 & 13.80 & 0.6868 & 1.8838 & 10.46 & 0.7217 & 1.8388 & 11.93 & 0.6995 & 1.8843 & 12.06 \\
        & ModernTCN    & 0.7470 & 1.6455 & 11.27 & \underline{0.7985} & 1.5017 & 8.31 & 0.6635 & 1.7506 & 11.09 & 0.7363 & 1.6326 & 10.22 \\
        & Crossformer  & 0.6789 & 1.7296 & 11.34 & 0.7945 & \underline{1.4771} & \underline{8.09} & 0.7373 & 1.5724 & 9.77 & 0.7369 & 1.5931 & 9.73 \\
        & DeformTime   & \textbf{0.7630} & \underline{1.6101} & \textbf{10.65} & 0.7953 & 1.5073 & 8.23 & \underline{0.7459} & \underline{1.5023} & \underline{9.46} & \underline{0.7681} & \textbf{1.5399} & \textbf{9.45} \\
        & Sonnet       & 0.7563 & 1.7449 & 11.65 & \textbf{0.8106} & \textbf{1.4349} & \textbf{7.92} & \textbf{0.7739} & \textbf{1.4610} & \textbf{9.17} & \textbf{0.7803} & \underline{1.5469} & \underline{9.58} \\
\bottomrule
\end{tabular}
% }
\caption{\WEATableCaptionAppendix{Hong Kong}{WEA-HK}}
\label{tab:results_wea_hk}
\end{table*}

\section{Supplementary experiment settings}
\label{appsec:supplementary_setting}
This section supplements Section~\ref{subsec:experiment_settings} in the main paper. All experiments were conducted using a Linux server (Ubuntu 24.04.2) with 3 NVIDIA L40S GPUs, 2 AMD EPYC 9354 CPUs, and 768GB of DDR5 RAM, except for the ablation experiments which were conducted using a Linux server (Ubuntu 22.04.5) with 2 NVIDIA A40 GPUs, 2 AMD EPYC 7443 CPUs, and 512GB of DDR4 RAM.

\subsection{Hyperparameter settings}
\label{appsubsec:hyperparameters_of_tasks}
The look-back window $L$ is set differently depending on the forecasting task. For the ETTh1, ETTh2 data sets, $L$ is set to $336$ time steps for all $4$ different forecasting horizons $H=\{96,192,336,720\}$, based on prior work~\citep{nie2023time}. In ELEC experiments, we set $L=168$ hours (a week, given the strong weekly periodicity in these time series) for all $H=\{12,24,36\}$. For ENER, $L$ is set equal to the forecasting horizon, \ie $L=H=\{24,48,72,168\}$. For the WEA tasks, we use a look-back window $L=\{28,28,56,240\}$ time steps for $H=\{4,12,28,120\}$, correspondingly. This means that we use the past $7$ days to conduct short-term forecasts ($1$ and $3$ days ahead), and the past $2H$ days for longer-term forecasts. For the ILI tasks, we set $L=\{28,28,56,56\}$ days respectively for forecasting horizons $H=\{7,14,21,28\}$ days ahead following prior work~\citep{shu2025deformtime}.

The batch size $d$ is set to $64$ for all tasks except for the ETT ones, where it is a learnable hyperparameter (see section~\ref{appsubsec:hyper_opt}). We train using a maximum of $100$ epochs (see our early stopping settings in ~\ref{appsubsec:hyper_opt}). Neural networks are optimised with Adam using mean squared error (MSE) loss on all time steps. Note that the output can be a time series (sequence) of one variable (the target) or more variables (target and exogenous), depending on the forecasting method.

\begin{table*}[!t]
\renewcommand{\arraystretch}{0.92}
\centering
\small
\setlength{\tabcolsep}{5.5pt}
\setlength{\aboverulesep}{0.5pt}
\setlength{\belowrulesep}{0.5pt}
% \resizebox{\linewidth}{!}{
\begin{tabular}{cl ccc ccc ccc ccc}
\toprule
\multirow{2}{*}{\textbf{H}} & \multirow{2}{*}{\textbf{Model}} & \multicolumn{3}{c}{\textbf{2016}} & \multicolumn{3}{c}{\textbf{2017}} & \multicolumn{3}{c}{\textbf{2018}} & \multicolumn{3}{c}{\textbf{Average}} \\
 & & $r$ & MAE & $\epsilon\%$ & $r$ & MAE & $\epsilon\%$ & $r$ & MAE & $\epsilon\%$ & $r$ & MAE & $\epsilon\%$ \\
\cmidrule{1-1}\cmidrule{2-2}\cmidrule(lr){3-5} \cmidrule(lr){6-8} \cmidrule(lr){9-11} \cmidrule(lr){12-14}
    \multirow{9}{*}{4} & DLinear      & 0.7760 & 0.4068 & 1.47 & 0.7495 & 0.4174 & 1.52 & 0.7885 & 0.3903 & 1.43 & 0.7713 & 0.4048 & 1.47 \\
        & iTransformer & 0.8033 & 0.3835 & 1.38 & 0.7777 & 0.3985 & 1.45 & 0.8089 & 0.3783 & 1.39 & 0.7967 & 0.3868 & 1.41 \\
        & PatchTST     & 0.7572 & 0.4194 & 1.51 & 0.7266 & 0.4364 & 1.59 & 0.7578 & 0.4156 & 1.53 & 0.7472 & 0.4238 & 1.54 \\
        & TimeXer      & 0.7933 & 0.3910 & 1.41 & 0.8010 & 0.3780 & 1.38 & 0.8147 & 0.3712 & 1.36 & 0.8030 & 0.3801 & 1.38 \\
        & Samformer    & 0.7917 & 0.3940 & 1.42 & 0.7690 & 0.4064 & 1.48 & 0.8109 & 0.3770 & 1.38 & 0.7905 & 0.3925 & 1.43 \\
        & ModernTCN    & 0.8194 & \underline{0.3664} & \underline{1.32} & 0.8135 & 0.3662 & 1.34 & 0.8286 & 0.3546 & 1.30 & 0.8205 & 0.3624 & 1.32 \\
        & Crossformer  & \underline{0.8198} & 0.3695 & 1.33 & \textbf{0.8403} & \underline{0.3441} & \underline{1.26} & \underline{0.8407} & \underline{0.3462} & \underline{1.27} & \underline{0.8336} & \underline{0.3532} & \underline{1.29} \\
        & DeformTime   & 0.8166 & 0.3737 & 1.35 & 0.8356 & 0.3452 & 1.26 & 0.8349 & 0.3481 & 1.28 & 0.8290 & 0.3557 & 1.30 \\
        & Sonnet       & \textbf{0.8313} & \textbf{0.3564} & \textbf{1.29} & \underline{0.8363} & \textbf{0.3431} & \textbf{1.25} & \textbf{0.8469} & \textbf{0.3336} & \textbf{1.22} & \textbf{0.8382} & \textbf{0.3444} & \textbf{1.25} \\
    \midrule
    \multirow{9}{*}{12} & DLinear      & 0.6873 & 0.4789 & 1.72 & 0.6565 & 0.4844 & 1.77 & 0.6639 & 0.4658 & 1.71 & 0.6693 & 0.4764 & 1.73 \\
        & iTransformer & 0.7134 & 0.4571 & 1.65 & 0.6811 & 0.4714 & 1.72 & 0.6759 & 0.4702 & 1.72 & 0.6901 & 0.4662 & 1.70 \\
        & PatchTST     & 0.6632 & 0.4880 & 1.76 & 0.6201 & 0.5099 & 1.86 & 0.6236 & 0.4995 & 1.83 & 0.6356 & 0.4992 & 1.82 \\
        & TimeXer      & 0.7178 & 0.4516 & 1.63 & 0.7203 & 0.4404 & 1.61 & 0.7084 & 0.4421 & 1.62 & 0.7155 & 0.4447 & 1.62 \\
        & Samformer    & 0.6976 & 0.4696 & 1.69 & 0.6613 & 0.4875 & 1.78 & 0.6680 & 0.4779 & 1.75 & 0.6756 & 0.4784 & 1.74 \\
        & ModernTCN    & 0.7328 & 0.4458 & 1.61 & 0.7179 & 0.4445 & 1.62 & 0.7022 & 0.4576 & 1.68 & 0.7177 & 0.4493 & 1.64 \\
        & Crossformer  & 0.7326 & 0.4548 & 1.64 & \underline{0.7615} & \underline{0.4124} & \underline{1.50} & 0.7283 & 0.4406 & 1.61 & \underline{0.7408} & 0.4359 & 1.59 \\
        & DeformTime   & \underline{0.7336} & \underline{0.4319} & \underline{1.56} & 0.7551 & 0.4203 & 1.53 & \underline{0.7325} & \underline{0.4246} & \underline{1.56} & 0.7404 & \underline{0.4256} & \underline{1.55} \\
        & Sonnet       & \textbf{0.7456} & \textbf{0.4302} & \textbf{1.55} & \textbf{0.7752} & \textbf{0.4009} & \textbf{1.46} & \textbf{0.7437} & \textbf{0.4169} & \textbf{1.53} & \textbf{0.7548} & \textbf{0.4160} & \textbf{1.51} \\
    \midrule
    \multirow{9}{*}{28} & DLinear      & 0.6046 & 0.5229 & 1.88 & 0.5531 & 0.5403 & 1.97 & 0.5513 & 0.5220 & 1.92 & 0.5697 & 0.5284 & 1.92 \\
        & iTransformer & 0.5986 & 0.5261 & 1.90 & 0.5899 & 0.5339 & 1.95 & 0.5652 & 0.5322 & 1.95 & 0.5846 & 0.5307 & 1.93 \\
        & PatchTST     & 0.5614 & 0.5468 & 1.97 & 0.5414 & 0.5593 & 2.04 & 0.5210 & 0.5566 & 2.04 & 0.5413 & 0.5542 & 2.02 \\
        & TimeXer      & \underline{0.6816} & \underline{0.4966} & \underline{1.79} & 0.6231 & 0.5027 & 1.83 & 0.6152 & 0.4930 & 1.81 & 0.6399 & 0.4974 & 1.81 \\
        & Samformer    & 0.5995 & 0.5262 & 1.90 & 0.5566 & 0.5498 & 2.00 & 0.5384 & 0.5503 & 2.02 & 0.5648 & 0.5421 & 1.97 \\
        & ModernTCN    & 0.6151 & 0.5183 & 1.87 & 0.6013 & 0.5186 & 1.89 & 0.5871 & 0.5220 & 1.92 & 0.6012 & 0.5196 & 1.89 \\
        & Crossformer  & 0.6660 & 0.5212 & 1.88 & \underline{0.6690} & 0.5088 & 1.85 & \underline{0.6692} & 0.4711 & 1.73 & \underline{0.6681} & 0.5003 & 1.82 \\
        & DeformTime   & 0.6705 & 0.5058 & 1.82 & 0.6664 & \underline{0.4869} & \underline{1.78} & 0.6638 & \underline{0.4698} & \underline{1.72} & 0.6669 & \underline{0.4875} & \underline{1.77} \\
        & Sonnet       & \textbf{0.7098} & \textbf{0.4605} & \textbf{1.66} & \textbf{0.6890} & \textbf{0.4721} & \textbf{1.72} & \textbf{0.6856} & \textbf{0.4633} & \textbf{1.70} & \textbf{0.6948} & \textbf{0.4653} & \textbf{1.69} \\
    \midrule
    \multirow{9}{*}{120} & DLinear      & 0.6260 & 0.5535 & 1.99 & 0.4837 & 0.5678 & 2.07 & 0.6341 & \underline{0.4773} & \underline{1.75} & 0.5812 & 0.5328 & 1.94 \\
        & iTransformer & 0.6079 & 0.5563 & 2.00 & 0.6114 & 0.5013 & 1.83 & 0.5811 & 0.5124 & 1.88 & 0.6001 & 0.5233 & 1.90 \\
        & PatchTST     & 0.6435 & \underline{0.5017} & \underline{1.81} & 0.4324 & 0.6084 & 2.22 & 0.6099 & 0.4972 & 1.82 & 0.5619 & 0.5357 & 1.95 \\
        & TimeXer      & 0.6139 & 0.5558 & 2.00 & 0.6150 & 0.5044 & 1.84 & 0.5935 & 0.5042 & 1.85 & 0.6075 & 0.5215 & 1.90 \\
        & Samformer    & 0.5653 & 0.5537 & 2.00 & 0.5719 & 0.5212 & 1.90 & 0.6136 & 0.4914 & 1.80 & 0.5836 & 0.5221 & 1.90 \\
        & ModernTCN    & 0.6442 & 0.5410 & 1.95 & 0.5708 & 0.5321 & 1.94 & 0.5764 & 0.5233 & 1.92 & 0.5971 & 0.5321 & 1.94 \\
        & Crossformer  & \underline{0.6500} & 0.5573 & 2.00 & 0.6383 & 0.5133 & 1.87 & \underline{0.6498} & 0.4929 & 1.81 & \underline{0.6461} & 0.5212 & 1.89 \\
        & DeformTime   & 0.6204 & 0.5384 & 1.94 & \textbf{0.6661} & \underline{0.4815} & \underline{1.75} & 0.6308 & 0.4951 & 1.81 & 0.6391 & \underline{0.5050} & \underline{1.83} \\
        & Sonnet       & \textbf{0.6609} & \textbf{0.5005} & \textbf{1.80} & \underline{0.6583} & \textbf{0.4794} & \textbf{1.75} & \textbf{0.6568} & \textbf{0.4692} & \textbf{1.72} & \textbf{0.6587} & \textbf{0.4830} & \textbf{1.76} \\
\bottomrule
\end{tabular}
% }
\caption{\WEATableCaptionAppendix{Singapore}{WEA-SG}}
\label{tab:results_wea_sg}
\end{table*}

\subsection{Hyperparameter settings specific to Sonnet}
\label{appsubsec:sonnet_hp_setup}
For Sonnet, the parameter $\alpha$ that controls the dimensionality of the embeddings of the endo- and exogenous variables is chosen from $\alpha\in\{0, .1, .25, .75\}$. The number of atoms, $K$, is selected from the set $K\in\{8, 16, 32\}$. The hidden dimension $d$ of Sonnet is set to be $64$.

\subsection{Hyperparameter settings specific to the ILI rate forecasting tasks}
\label{appsubsec:hyp_settings_ili}
In the ILI rate forecasting task, the number of exogenous variables is decided using a learnable threshold $\tau$ based on linear correlation (with the endogenous/target variable in the training data), following DeformTime~\citep{shu2025deformtime}. We select $\tau$ from $\{.3, .4, .5\}$ for the US regions, and from $\{.05, .1, .2, .3, .4, .5\}$ for England. An increase in the correlation threshold leads to a reduction in the number of selected queries. Consequently, the number of search queries selected by the models ranges from $13$ to $51$ for US regions, and from $80$ to $752$ for England. For the models that require an excessive amount of GPU memory for larger sets of exogenous variables, \ie ModernTCN, PatchTST and Crossformer, we restricted $\tau\in\{.3, .4, .5\}$ for all regions.

\subsection{Hyperparameter validation}
\label{appsubsec:hyper_opt}
For the ETT forecasting tasks, the batch size and the hidden dimension $d$ are both selected from $\{16,32,64\}$. For all tasks, Adam's initial learning rate is selected from $\{2, 1, .5, .2, .1, .05\}\times10^{-3}$. The learning rate is decayed to $0$ with a linear decay scheduler applied after every epoch. If a model structure contains a dropout parameter, its value is selected from $\{0,.1,.2\}$. Hyperparameter values are determined using grid search. Training stops early when the validation error of the target variable over time $\{t+1,\cdots,t+H\}$ does not decrease further (compared to its lowest value) for $5$ consecutive epochs. We then use the model with the lowest validation loss. 

\begin{table*}[!t]
\renewcommand{\arraystretch}{0.92}
\centering
\small
\setlength{\tabcolsep}{5.5pt}
\setlength{\aboverulesep}{0.5pt}
\setlength{\belowrulesep}{0.5pt}
% \resizebox{\linewidth}{!}{
\begin{tabular}{cl ccc ccc ccc ccc}
\toprule
\multirow{2}{*}{\textbf{H}} & \multirow{2}{*}{\textbf{Model}} & \multicolumn{3}{c}{\textbf{2016}} & \multicolumn{3}{c}{\textbf{2017}} & \multicolumn{3}{c}{\textbf{2018}} & \multicolumn{3}{c}{\textbf{Average}} \\
 & & $r$ & MAE & $\epsilon\%$ & $r$ & MAE & $\epsilon\%$ & $r$ & MAE & $\epsilon\%$ & $r$ & MAE & $\epsilon\%$ \\
\cmidrule{1-1}\cmidrule{2-2}\cmidrule(lr){3-5} \cmidrule(lr){6-8} \cmidrule(lr){9-11} \cmidrule(lr){12-14}
    \multirow{9}{*}{4} & DLinear      & 0.6899 & 3.0696 & 17.69 & 0.6677 & 3.3553 & 19.03 & 0.6757 & 3.0199 & 17.93 & 0.6778 & 3.1483 & 18.22 \\
        & iTransformer & 0.8466 & 2.1083 & 12.25 & 0.8325 & 2.3752 & 13.70 & 0.8464 & 2.0884 & 12.54 & 0.8418 & 2.1906 & 12.83 \\
        & PatchTST     & 0.5931 & 3.3676 & 19.33 & 0.5788 & 3.7192 & 21.07 & 0.5372 & 3.4144 & 20.12 & 0.5697 & 3.5004 & 20.17 \\
        & TimeXer      & 0.8464 & 2.1310 & 12.48 & 0.8412 & 2.4309 & 13.92 & 0.8407 & 2.1510 & 13.03 & 0.8428 & 2.2376 & 13.14 \\
        & Samformer    & 0.8495 & 2.0911 & 12.19 & 0.8248 & 2.4305 & 14.03 & 0.8338 & 2.1580 & 13.04 & 0.8361 & 2.2265 & 13.09 \\
        & ModernTCN    & 0.8955 & 1.7858 & 10.42 & 0.8852 & 1.9753 & 11.44 & 0.8806 & 1.8644 & 11.30 & 0.8871 & 1.8752 & 11.05 \\
        & Crossformer  & \underline{0.9114} & \underline{1.5785} & \textbf{9.13} & \textbf{0.9070} & \underline{1.7736} & \underline{10.34} & \underline{0.9135} & \underline{1.5626} & \textbf{9.38} & \underline{0.9106} & \underline{1.6382} & \textbf{9.62} \\
        & DeformTime   & 0.9044 & 1.7005 & 9.98 & 0.8880 & 1.9208 & 11.12 & 0.9057 & 1.6589 & 10.15 & 0.8994 & 1.7600 & 10.41 \\
        & Sonnet       & \textbf{0.9169} & \textbf{1.5732} & \underline{9.20} & \underline{0.9068} & \textbf{1.7550} & \textbf{10.26} & \textbf{0.9160} & \textbf{1.5439} & \underline{9.42} & \textbf{0.9133} & \textbf{1.6240} & \underline{9.63} \\
    \midrule
    \multirow{9}{*}{12} & DLinear      & 0.3597 & 3.9731 & 22.60 & 0.3874 & 4.1974 & 23.32 & 0.3239 & 3.9091 & 22.89 & 0.3570 & 4.0265 & 22.94 \\
        & iTransformer & 0.4321 & 3.8589 & 22.11 & 0.4092 & 4.2382 & 23.80 & 0.3857 & 3.8253 & 22.58 & 0.4090 & 3.9741 & 22.83 \\
        & PatchTST     & 0.3525 & 4.1364 & 23.67 & 0.3792 & 4.3989 & 24.52 & 0.3214 & 4.0377 & 23.78 & 0.3511 & 4.1910 & 23.99 \\
        & TimeXer      & 0.4979 & 3.6755 & 20.94 & 0.4804 & 3.9628 & 22.14 & 0.4763 & 3.5340 & 20.86 & 0.4849 & 3.7241 & 21.31 \\
        & Samformer    & 0.4245 & 3.9482 & 22.49 & 0.4137 & 4.2929 & 23.97 & 0.3626 & 3.8921 & 22.95 & 0.4003 & 4.0444 & 23.14 \\
        & ModernTCN    & 0.4861 & 3.6737 & 21.11 & 0.4560 & 4.0876 & 22.78 & 0.4537 & 3.5671 & 21.10 & 0.4653 & 3.7761 & 21.67 \\
        & Crossformer  & \underline{0.5485} & \underline{3.5536} & \textbf{20.27} & 0.5238 & 3.8275 & 21.42 & 0.5246 & 3.3985 & 20.16 & 0.5323 & 3.5932 & 20.62 \\
        & DeformTime   & 0.5377 & 3.5610 & 20.29 & \underline{0.5394} & \underline{3.7613} & \underline{21.00} & \underline{0.5267} & \underline{3.3821} & \underline{20.10} & \underline{0.5346} & \underline{3.5681} & \underline{20.46} \\
        & Sonnet       & \textbf{0.5527} & \textbf{3.5423} & \underline{20.27} & \textbf{0.5452} & \textbf{3.7261} & \textbf{20.89} & \textbf{0.5389} & \textbf{3.3613} & \textbf{19.99} & \textbf{0.5456} & \textbf{3.5432} & \textbf{20.38} \\
    \midrule
    \multirow{9}{*}{28} & DLinear      & 0.3869 & 3.9030 & 22.17 & 0.4155 & 4.1227 & 22.99 & 0.4047 & 3.7504 & 21.94 & 0.4024 & 3.9254 & 22.37 \\
        & iTransformer & 0.3651 & 4.0092 & 22.90 & 0.4111 & 4.1308 & 23.10 & 0.4143 & 3.7352 & 21.79 & 0.3969 & 3.9584 & 22.60 \\
        & PatchTST     & 0.3824 & 3.9671 & 22.57 & 0.4173 & 4.1099 & 23.01 & 0.4097 & 3.7444 & 21.86 & 0.4031 & 3.9405 & 22.48 \\
        & TimeXer      & 0.4448 & 3.8264 & 21.80 & \textbf{0.5026} & 3.9775 & 22.20 & 0.4052 & 3.6936 & 21.60 & 0.4509 & 3.8325 & 21.87 \\
        & Samformer    & 0.3968 & 3.9169 & 22.27 & 0.4159 & 4.1299 & 23.09 & 0.4219 & 3.7249 & 21.76 & 0.4116 & 3.9239 & 22.37 \\
        & ModernTCN    & 0.3972 & 3.9191 & 22.28 & 0.3855 & 4.2644 & 23.63 & \underline{0.4480} & 3.6363 & 21.18 & 0.4102 & 3.9399 & 22.36 \\
        & Crossformer  & 0.4491 & 3.8314 & 21.83 & 0.4861 & \underline{3.9413} & \underline{22.02} & 0.4135 & 3.6457 & 21.32 & 0.4495 & 3.8061 & 21.72 \\
        & DeformTime   & \underline{0.4704} & \underline{3.7845} & \underline{21.60} & 0.4990 & 3.9476 & 22.06 & \textbf{0.4510} & \textbf{3.5481} & \textbf{20.82} & \underline{0.4735} & \underline{3.7601} & \underline{21.49} \\
        & Sonnet       & \textbf{0.4829} & \textbf{3.7430} & \textbf{21.39} & \underline{0.5017} & \textbf{3.8840} & \textbf{21.73} & 0.4436 & \underline{3.5560} & \underline{20.84} & \textbf{0.4761} & \textbf{3.7277} & \textbf{21.32} \\
    \midrule
    \multirow{9}{*}{120} & DLinear      & 0.2564 & 4.1508 & 23.51 & 0.4413 & 4.0823 & 22.74 & 0.2488 & 3.9379 & 22.90 & 0.3155 & 4.0570 & 23.05 \\
        & iTransformer & 0.3760 & 4.0470 & 22.88 & 0.4466 & 3.9902 & 22.30 & 0.3580 & 3.8046 & 22.30 & 0.3935 & 3.9473 & 22.49 \\
        & PatchTST     & 0.3485 & 4.0798 & 23.27 & 0.4448 & 4.0213 & 22.37 & 0.2282 & 4.0629 & 23.42 & 0.3405 & 4.0547 & 23.02 \\
        & TimeXer      & 0.4334 & 3.8432 & 21.84 & 0.4717 & 4.0044 & 22.36 & \underline{0.4079} & 3.6759 & 21.44 & 0.4377 & 3.8412 & 21.88 \\
        & Samformer    & 0.3890 & 4.0904 & 23.47 & 0.4684 & 3.9402 & 21.96 & 0.3858 & 3.6749 & 21.47 & 0.4144 & 3.9018 & 22.30 \\
        & ModernTCN    & 0.4450 & 3.9583 & 22.94 & 0.4212 & 4.1221 & 23.01 & 0.2811 & 4.1861 & 24.68 & 0.3824 & 4.0889 & 23.54 \\
        & Crossformer  & \underline{0.4738} & 3.7648 & 21.45 & \textbf{0.5259} & \textbf{3.8475} & \textbf{21.54} & 0.3787 & 3.6852 & 21.49 & \underline{0.4595} & \underline{3.7659} & \underline{21.49} \\
        & DeformTime   & \textbf{0.4758} & \underline{3.7615} & \underline{21.44} & 0.4537 & 3.9905 & 22.23 & 0.3999 & \underline{3.6602} & \underline{21.34} & 0.4431 & 3.8040 & 21.67 \\
        & Sonnet       & 0.4718 & \textbf{3.7594} & \textbf{21.41} & \underline{0.5126} & \underline{3.9088} & \underline{21.86} & \textbf{0.4389} & \textbf{3.5438} & \textbf{20.79} & \textbf{0.4744} & \textbf{3.7373} & \textbf{21.35} \\
\bottomrule
\end{tabular}
% }
\caption{\WEATableCaptionAppendix{Cape Town}{WEA-CT}}
\label{tab:results_wea_ct}
\end{table*}

\subsection{Data normalisation}
\label{appsubsec:data_normalisation}
We apply z-score normalisation to the data of all forecasting tasks. This standardises endogenous and exogenous variables in a training set to have a zero mean and unit standard deviation. Normalisation is transferred from the training data to the test data. Data is reverted back to the original scale when computing prediction errors. Additionally, for the ETT, ENER, and ELEC data sets, we adopt variable normalisation~\citep{kim2022reversible} over the target variable within the input's look-back window following prior work~\citep{nie2023time,shu2025deformtime}. 

\begin{table*}[!t]
    \renewcommand{\arraystretch}{0.92}
    \centering
    \setlength{\tabcolsep}{6pt}
    \setlength{\aboverulesep}{1.2pt}
    \setlength{\belowrulesep}{1.2pt}
    \small
    % \resizebox{\textwidth}{!}{%
    \begin{tabular}{llcccccccccccc}
      \toprule
\multirow{2}{*}{Model} & \multirow{2}{*}{Variant} & \multicolumn{3}{c}{$H=7$}  & \multicolumn{3}{c}{$H=14$}  & \multicolumn{3}{c}{$H=21$}  & \multicolumn{3}{c}{$H=28$}  \\
 &  & ENG & US2 & US9 & ENG & US2 & US9 & ENG & US2 & US9 & ENG & US2 & US9 \\
% \midrule
\cmidrule(lr){3-5} \cmidrule(lr){6-8} \cmidrule(lr){9-11} \cmidrule(lr){12-14}
\multirow{6}{*}
{\rotatebox[origin=c]{90}{iTransformer}} & --- & \underline{26.38} & \underline{23.24} & \underline{18.57} & \textbf{36.67} & \underline{28.17} & \underline{22.44} & \underline{48.93} & \underline{30.03} & 24.11 & \underline{55.35} & 36.75 & 31.05 \\
 & $\neg$ Attn & \textbf{26.22} & 25.52 & 21.22 & \underline{37.10} & 30.54 & 25.77 & 54.52 & 34.77 & 29.89 & 61.97 & 41.18 & 34.97 \\
 & FNet~\shortcite{lee-thorp2022fnet} & 30.16 & 26.99 & 22.27 & 41.28 & 31.12 & 26.29 & 57.16 & 36.48 & 29.69 & 64.26 & 41.06 & 35.00 \\
 & FED~\shortcite{zhou2022fedformer} & 40.21 & 32.58 & 26.25 & 48.37 & 37.86 & 30.68 & 67.76 & 50.94 & 40.97 & 73.86 & 54.82 & 44.33 \\
 & VDAB~\shortcite{shu2025deformtime} & 27.46 & 24.55 & 19.21 & 38.10 & 30.39 & 23.29 & \textbf{48.19} & 32.37 & \underline{23.07} & \textbf{54.16} & \underline{35.39} & \underline{26.97} \\
 & MVCA & 26.87 & \textbf{20.27} & \textbf{16.34} & 37.71 & \textbf{26.01} & \textbf{20.77} & 50.64 & \textbf{27.94} & \textbf{22.19} & 57.64 & \textbf{33.23} & \textbf{24.89} \\
\cmidrule{1-14}
\multirow{6}{*}{\rotatebox[origin=c]{90}{SamFormer}} & --- & 28.31 & 24.21 & 19.39 & \textbf{36.63} & 30.16 & 24.50 & 54.41 & 31.42 & 24.22 & 60.78 & 36.8 & 27.95 \\
 & $\neg$ Attn & 29.99 & 29.26 & 24.13 & 40.52 & 34.58 & 28.73 & 52.00 & 41.77 & 36.09 & 62.22 & 46.08 & 40.87 \\
 & FNet~\shortcite{lee-thorp2022fnet} & 30.01 & 27.94 & 21.95 & 40.33 & 33.06 & 27.33 & 53.16 & 36.21 & 31.30 & 61.64 & 42.28 & 34.68 \\
 & FED~\shortcite{zhou2022fedformer} & 40.18 & 32.35 & 26.34 & 48.91 & 37.65 & 30.84 & 67.79 & 49.99 & 41.26 & 73.17 & 53.91 & 44.85 \\
 & VDAB~\shortcite{shu2025deformtime} & \textbf{25.98} & \underline{21.88} & \underline{17.05} & \underline{36.84} & \underline{27.48} & \textbf{20.95} & \textbf{46.52} & \underline{30.78} & \textbf{22.29} & \underline{57.09} & \underline{33.49} & \underline{25.63} \\
 & MVCA & \underline{26.13} & \textbf{20.85} & \textbf{16.90} & 38.14 & \textbf{26.14} & \underline{21.06} & \underline{49.15} & \textbf{28.55} & \underline{22.57} & \textbf{55.27} & \textbf{32.39} & \textbf{24.66} \\
\cmidrule{1-14}
\multirow{6}{*}{\rotatebox[origin=c]{90}{PatchTST}} & --- & 27.61 & 24.51 & 19.34 & \textbf{37.76} & \underline{30.11} & 24.09 & 51.11 & 36.70 & 29.40 & \underline{59.60} & 42.61 & 33.35 \\
 & $\neg$ Attn & 28.54 & 26.99 & 21.51 & 41.64 & 31.82 & 26.36 & 53.64 & 38.75 & 31.75 & 60.11 & 43.25 & 36.17 \\
 & FNet~\shortcite{lee-thorp2022fnet} & 31.93 & 28.81 & 22.14 & 44.15 & 33.77 & 25.64 & 58.17 & 38.11 & 26.46 & 60.51 & 41.48 & 29.92 \\
 & FED~\shortcite{zhou2022fedformer} & 40.21 & 32.59 & 26.25 & 50.26 & 37.88 & 30.76 & 67.85 & 50.89 & 41.16 & 74.08 & 54.61 & 44.42 \\
 & VDAB~\shortcite{shu2025deformtime} & \underline{26.88} & \underline{22.99} & \underline{17.99} & 38.08 & 30.99 & \underline{20.64} & \underline{50.49} & \textbf{32.45} & \underline{23.90} & 60.16 & \textbf{34.89} & \textbf{25.61} \\
 & MVCA & \textbf{25.77} & \textbf{22.90} & \textbf{17.50} & \underline{37.97} & \textbf{28.33} & \textbf{20.48} & \textbf{49.10} & \underline{32.45} & \textbf{22.05} & \textbf{59.48} & \underline{36.14} & \underline{25.97} \\
      \bottomrule
    \end{tabular}
    % }%
\caption{Performance (average sMAPE across the $4$ test seasons) of iTransformer, Samformer, and PatchTST on the ILI forecasting tasks (ILI-ENG/US2/US9) with different modifications to the na\"ive attention mechanism. `$\neg$ Attn' denotes the removal of the residual attention module, and FNet / FED / VDAB refer to using the attention modules proposed in FNet, FEDformer, and DeformTime, respectively. This table supplements Table~\ref{tab:ILI-results-vca}.
% that reports MAEs. 
\bestResults}
\label{tab:ILI-results-vca-smape}
\end{table*}

\subsection{sMAPE definition for the weather (temperature) prediction tasks}
\label{appsubsec:special_smape}
We note that for the WEA tasks, the unit of the temperature time series sequence is originally captured in Kelvin.\footnote{$0$ degrees in Celsius is equal to $273.15$ Kelvin.} Directly computing the sMAPE using Kelvin units will lead to a relatively small error (MAEs are not affected, but sMAPEs will be significantly lower and harder to display in an accessible way). To address this, we subtract the minimum value of the target variable across the test set ($\xi = \min(\mathbf{y})$), and then add a constant number $a=30$ to both the ground truth and the predicted value to avoid division by $0$. This alters the formula of sMAPE as follows:
\begin{equation}
    \mathrm{sMAPE}(\mathbf{y}, \hat{\mathbf{y}}) = \frac{100}{L} \sum_{t=1}^{L} \frac{2|y_t - \hat{y}_t|}{\left( |y_t - \xi| + |\hat{y}_t - \xi| + 2a \right)} \, .
\end{equation}

\subsection{Random seed initialisation}
\label{appsubsec:random_seed_setup}
We use a fixed seed equal to `\texttt{42}' during training for all forecasting tasks, except for the ETTh1 and ETTh2 data sets. For these data sets, we follow prior work~\citep{nie2023time,zeng2023are} and use `\texttt{2021}', with the exception of iTransformer, which uses `\texttt{2023}' in line with its official configuration.

\begin{table}[!b]
\renewcommand{\arraystretch}{0.92}
\centering\fontsize{9}{10}\selectfont
% \small
\setlength{\tabcolsep}{0.5pt}
\setlength{\aboverulesep}{1.2pt}
\setlength{\belowrulesep}{1.2pt}
% \resizebox{\linewidth}{!}{%
\begin{tabular}{cccccccccccc}
\toprule
\multicolumn{2}{c}{} & \multicolumn{2}{c}{\bf$\neg$ Coher} & \multicolumn{2}{c}{\bf$\neg$ MLP} & \multicolumn{2}{c}{\bf$\neg$ MVCA} & \multicolumn{2}{c}{\bf$\neg$ Embed} & \multicolumn{2}{c}{\bf$\neg$ Koop} \\
 & $H$ & MAE & $\Delta\%$ & MAE & $\Delta\%$ & MAE & $\Delta\%$ & MAE & $\Delta\%$ & MAE & $\Delta\%$ \\
\midrule
\multirow{4}{*}{\rotatebox{90}{ILI-ENG}}
& 7 & 1.422 & --3.8 & 1.531 & \cellcolor{gray!20}{+3.5} & 1.591 & \cellcolor{gray!20}{+7.5} & 1.633 & \cellcolor{gray!20}{\textbf{+10.4}} & 1.486 & \cellcolor{gray!20}{+0.4} \\
         
& 14 & 2.110 & \cellcolor{gray!20}{+9.7} & 2.171 & \cellcolor{gray!20}{+12.9} & 2.235 & \cellcolor{gray!20}{\textbf{+16.2}} & 2.066 & \cellcolor{gray!20}{+7.5} & 2.079 & \cellcolor{gray!20}{+8.2} \\
         
& 21 & 2.880 & \cellcolor{gray!20}{\textbf{+14.7}} & 2.614 & \cellcolor{gray!20}{+4.1} & 2.794 & \cellcolor{gray!20}{+11.3} & 2.647 & \cellcolor{gray!20}{+5.4} & 2.630 & \cellcolor{gray!20}{+4.8} \\
         
& 28 & 3.420 & \cellcolor{gray!20}{\textbf{+24.4}} & 3.114 & \cellcolor{gray!20}{+13.3} & 3.274 & \cellcolor{gray!20}{+19.2} & 3.091 & \cellcolor{gray!20}{+12.5} & 3.062 & \cellcolor{gray!20}{+11.4} \\
\midrule
\multirow{4}{*}{\rotatebox{90}{ILI-US2}}
& 7 & 0.414 & \cellcolor{gray!20}{+8.8} & 0.411 & \cellcolor{gray!20}{+8.0} & 0.432 & \cellcolor{gray!20}{\textbf{+13.4}} & 0.416 & \cellcolor{gray!20}{+9.4} & 0.420 & \cellcolor{gray!20}{+10.5} \\
         
& 14 & 0.512 & \cellcolor{gray!20}{+14.0} & 0.507 & \cellcolor{gray!20}{+13.0} & 0.528 & \cellcolor{gray!20}{+17.5} & 0.530 & \cellcolor{gray!20}{\textbf{+17.9}} & 0.496 & \cellcolor{gray!20}{+10.5} \\
         
& 21 & 0.609 & \cellcolor{gray!20}{+14.3} & 0.596 & \cellcolor{gray!20}{+12.0} & 0.671 & \cellcolor{gray!20}{\textbf{+26.0}} & 0.573 & \cellcolor{gray!20}{+7.6} & 0.590 & \cellcolor{gray!20}{+10.8} \\
         
& 28 & 0.623 & \cellcolor{gray!20}{+7.7} & 0.662 & \cellcolor{gray!20}{+14.4} & 0.666 & \cellcolor{gray!20}{\textbf{+15.0}} & 0.606 & \cellcolor{gray!20}{+4.8} & 0.634 & \cellcolor{gray!20}{+9.5} \\
\midrule
\multirow{4}{*}{\rotatebox{90}{ILI-US9}}
& 7 & 0.272 & \cellcolor{gray!20}{+1.9} & 0.302 & \cellcolor{gray!20}{\textbf{+13.1}} & 0.277 & \cellcolor{gray!20}{+4.0} & 0.278 & \cellcolor{gray!20}{+4.4} & 0.280 & \cellcolor{gray!20}{+4.8} \\
         
& 14 & 0.318 & \cellcolor{gray!20}{+13.4} & 0.334 & \cellcolor{gray!20}{\textbf{+19.0}} & 0.331 & \cellcolor{gray!20}{+17.8} & 0.315 & \cellcolor{gray!20}{+12.1} & 0.315 & \cellcolor{gray!20}{+12.3} \\
         
& 21 & 0.358 & \cellcolor{gray!20}{+12.5} & 0.326 & \cellcolor{gray!20}{+2.7} & 0.393 & \cellcolor{gray!20}{\textbf{+23.6}} & 0.311 & --2.1 & 0.319 & \cellcolor{gray!20}{+0.4} \\
         
& 28 & 0.428 & \cellcolor{gray!20}{\textbf{+16.4}} & 0.345 & --6.2 & 0.401 & \cellcolor{gray!20}{+9.2} & 0.359 & --2.4 & 0.348 & --5.4 \\
\midrule
\multirow{4}{*}{\rotatebox{90}{WEA-CT}}
& 4 & 1.663 & \cellcolor{gray!20}{+2.4} & 1.643 & \cellcolor{gray!20}{+1.2} & 1.636 & \cellcolor{gray!20}{+0.7} & 1.678 & \cellcolor{gray!20}{\textbf{+3.3}} & 1.650 & \cellcolor{gray!20}{+1.6} \\
         
& 12 & 3.591 & \cellcolor{gray!20}{\textbf{+1.3}} & 3.574 & \cellcolor{gray!20}{+0.9} & 3.589 & \cellcolor{gray!20}{+1.3} & 3.572 & \cellcolor{gray!20}{+0.8} & 3.568 & \cellcolor{gray!20}{+0.7} \\
         
& 28 & 3.783 & \cellcolor{gray!20}{+1.5} & 3.760 & \cellcolor{gray!20}{+0.9} & 3.862 & \cellcolor{gray!20}{\textbf{+3.6}} & 3.765 & \cellcolor{gray!20}{+1.0} & 3.754 & \cellcolor{gray!20}{+0.7} \\
         
& 120 & 3.769 & \cellcolor{gray!20}{\textbf{+0.9}} & 3.764 & \cellcolor{gray!20}{+0.7} & 3.768 & \cellcolor{gray!20}{+0.8} & 3.766 & \cellcolor{gray!20}{+0.8} & 3.767 & \cellcolor{gray!20}{+0.8} \\

\midrule

\multirow{4}{*}{\rotatebox{90}{WEA-HK}}
& 4 & 0.639 & \cellcolor{gray!20}{+0.0} & 0.639 & \cellcolor{gray!20}{+0.1} & 0.631 & --1.2 & 0.655 & \cellcolor{gray!20}{\textbf{+2.6}} & 0.639 & \cellcolor{gray!20}{+0.1} \\
         
& 12 & 1.290 & \cellcolor{gray!20}{+4.4} & 1.261 & \cellcolor{gray!20}{+2.0} & 1.302 & \cellcolor{gray!20}{\textbf{+5.4}} & 1.278 & \cellcolor{gray!20}{+3.4} & 1.273 & \cellcolor{gray!20}{+3.1} \\
         
& 28 & 1.475 & \cellcolor{gray!20}{+4.3} & 1.467 & \cellcolor{gray!20}{+3.8} & 1.477 & \cellcolor{gray!20}{\textbf{+4.5}} & 1.467 & \cellcolor{gray!20}{+3.8} & 1.463 & \cellcolor{gray!20}{+3.5} \\
         
& 120 & 1.635 & \cellcolor{gray!20}{\textbf{+5.7}} & 1.627 & \cellcolor{gray!20}{+5.2} & 1.620 & \cellcolor{gray!20}{+4.7} & 1.599 & \cellcolor{gray!20}{+3.4} & 1.604 & \cellcolor{gray!20}{+3.7} \\
\midrule
\multirow{4}{*}{\rotatebox{90}{WEA-LD}}
& 4 & 1.754 & \cellcolor{gray!20}{+1.8} & 1.767 & \cellcolor{gray!20}{+2.6} & 1.742 & \cellcolor{gray!20}{+1.1} & 1.787 & \cellcolor{gray!20}{\textbf{+3.7}} & 1.755 & \cellcolor{gray!20}{+1.9} \\
         
& 12 & 2.991 & \cellcolor{gray!20}{+1.1} & 3.007 & \cellcolor{gray!20}{+1.6} & 3.001 & \cellcolor{gray!20}{+1.4} & 3.019 & \cellcolor{gray!20}{\textbf{+2.0}} & 2.993 & \cellcolor{gray!20}{+1.1} \\
         
& 28 & 3.313 & \cellcolor{gray!20}{+3.0} & 3.281 & \cellcolor{gray!20}{+2.0} & 3.328 & \cellcolor{gray!20}{\textbf{+3.5}} & 3.326 & \cellcolor{gray!20}{+3.4} & 3.268 & \cellcolor{gray!20}{+1.6} \\
         
& 120 & 3.385 & \cellcolor{gray!20}{\textbf{+4.3}} & 3.347 & \cellcolor{gray!20}{+3.1} & 3.375 & \cellcolor{gray!20}{+4.0} & 3.359 & \cellcolor{gray!20}{+3.5} & 3.345 & \cellcolor{gray!20}{+3.1} \\
\midrule
\multirow{4}{*}{\rotatebox{90}{WEA-NY}}
& 4 & 1.296 & \cellcolor{gray!20}{+1.9} & 1.294 & \cellcolor{gray!20}{+1.8} & 1.281 & \cellcolor{gray!20}{+0.7} & 1.316 & \cellcolor{gray!20}{\textbf{+3.5}} & 1.284 & \cellcolor{gray!20}{+1.0} \\
         
& 12 & 2.447 & --0.0 & 2.464 & \cellcolor{gray!20}{+0.7} & 2.468 & \cellcolor{gray!20}{\textbf{+0.9}} & 2.467 & \cellcolor{gray!20}{+0.8} & 2.460 & \cellcolor{gray!20}{+0.5} \\
         
& 28 & 2.751 & \cellcolor{gray!20}{+2.9} & 2.708 & \cellcolor{gray!20}{+1.3} & 2.764 & \cellcolor{gray!20}{\textbf{+3.4}} & 2.739 & \cellcolor{gray!20}{+2.4} & 2.723 & \cellcolor{gray!20}{+1.8} \\
         
& 120 & 2.856 & \cellcolor{gray!20}{+5.3} & 2.777 & \cellcolor{gray!20}{+2.4} & 2.938 & \cellcolor{gray!20}{\textbf{+8.3}} & 2.840 & \cellcolor{gray!20}{+4.7} & 2.801 & \cellcolor{gray!20}{+3.2} \\
\midrule
\multirow{4}{*}{\rotatebox{90}{WEA-SG}}
& 4 & 0.342 & --0.8 & 0.345 & \cellcolor{gray!20}{+0.1} & 0.343 & --0.5 & 0.348 & \cellcolor{gray!20}{\textbf{+1.1}} & 0.343 & --0.4 \\
         
& 12 & 0.423 & \cellcolor{gray!20}{+1.6} & 0.419 & \cellcolor{gray!20}{+0.8} & 0.426 & \cellcolor{gray!20}{\textbf{+2.4}} & 0.423 & \cellcolor{gray!20}{+1.6} & 0.422 & \cellcolor{gray!20}{+1.4} \\
         
& 28 & 0.489 & \cellcolor{gray!20}{\textbf{+5.2}} & 0.476 & \cellcolor{gray!20}{+2.2} & 0.474 & \cellcolor{gray!20}{+1.9} & 0.475 & \cellcolor{gray!20}{+2.2} & 0.476 & \cellcolor{gray!20}{+2.3} \\
         
& 120 & 0.500 & \cellcolor{gray!20}{+3.5} & 0.494 & \cellcolor{gray!20}{+2.3} & 0.503 & \cellcolor{gray!20}{\textbf{+4.1}} & 0.497 & \cellcolor{gray!20}{+2.8} & 0.495 & \cellcolor{gray!20}{+2.5} \\

\bottomrule
\end{tabular}
% }
\caption{Ablation study results across the ILI (4 test seasons) and WEA tasks (3 test years). MAE is shown for each ablated variation ($\neg$ denotes removal of a component). $\Delta\%$ is the MAE percentage of difference compared to Sonnet's MAE (which can be seen in Table~\ref{tab:results_averaged}). MAE values are rounded to 3 decimals for spacing. Grey background denotes a performance drop (increased MAE) compared to Sonnet.}
\label{tab:ablation}
\end{table}

\subsection{Replacing attention modules with MVCA}
\label{appsubsec:setup_mvca}
To use MVCA in place of vanilla attention in the $3$ base models for experiments in Section~\ref{subsec:forecasting_results_vca}, some modifications are needed as MVCA initially expects a $3$-dimensional input in $\mathbb{R}^{K \times L \times d}$ with $K$ as the number of wavelet transformations (which we use as the number of heads in multi-head MVCA). In contrast, the attention module in the three base models takes input embeddings with only $2$ dimensions $\in\mathbb{R}^{u \times v}$, which represent time and variables. To address this discrepancy, we split the embedding into multiple attention heads, as MVCA is essentially a variant similar to multihead attention. Specifically, given an embedding in shape $\mathbb{R}^{u \times v}$, we first split it along the feature dimension into $K$ heads, resulting in an input dimension of $v/K$ for each head, where the head embeddings are in shape $\mathbb{R}^{K \times u \times (v/K)}$. The remaining operations then follow the standard MVCA process, with $u$ and $v/K$ corresponding to $L$ and $d$ respectively. The output of different heads is also joined along the feature dimension to form the final output in shape $\mathbb{R}^{u \times v}$. In our experiment, the number of heads is fixed to $8$ for all base models when replacing the attention with MVCA.

\section{Supplementary results}
\label{appsec:supp_results}
This section supplements section~\ref{sec:results} of the main paper.

\subsection{Detailed forecasting results with the ILI and weather forecasting tasks}
\label{appsubsec:full_results}
Complete results, broken down across all test periods, models, and forecasting horizons for the ILI rate prediction tasks in England (ILI-ENG), US Region 2 (ILI-US2), and US Region 9 (ILI-US9) are enumerated in Tables~\ref{tab:results_ili_eng},~\ref{tab:results_ili_us2}, and~\ref{tab:results_ili_us9}, respectively. Similarly, Tables~\ref{tab:results_wea_ld}--\ref{tab:results_wea_ct} present the corresponding results for weather forecasting in London (WEA-LD), New York (WEA-NY), Singapore (WEA-SG), Hong Kong (WEA-HK), and Cape Town (WEA-CT). For both tasks, we report MAE, sMAPE ($\epsilon\%$), and the linear correlation coefficient (denoted by $r$) between the model predictions and the target values over each test period.

\subsection{Supplementary experiment results with sMAPE score}
\label{appsubsec:mvca_smape}

Results enumerated in Table~\ref{tab:ILI-results-vca-smape} complement the ones presented in Section~\ref{subsec:forecasting_results_vca}, Table~\ref{tab:ILI-results-vca} by providing sMAPE estimates ($\epsilon\%$). On average, replacing the attention in the base models with MVCA reduces sMAPE by $9\%$. Moreover, compared to the best-performing variant (VDAB), our proposed MVCA reduces sMAPE by $2.3\%$. This reduction in relative error further consolidates our argument that MVCA provides better prediction accuracy in MTS forecasting.

\begin{table}[!b]
    \renewcommand{\arraystretch}{0.92}
    \centering
    \small
    \setlength{\tabcolsep}{2.5pt}
    \setlength{\aboverulesep}{0pt}
    \setlength{\belowrulesep}{1.5pt}
    % \resizebox{\linewidth}{!}{%
    \begin{tabular}{cr cc cc c}
    \toprule
    $H$ &\bf seed &
    {\bf 2015/16} & 
    {\bf 2016/17} & 
    {\bf 2017/18} & 
    {\bf 2018/19} & 
    {\bf Average} \\
    \midrule
    \multirow{6}{*}{7}
    & \tt 10 & 1.088 & 1.194 & 1.756 & 1.490 & 1.382 \\
    & \tt 42 & 1.029 & 1.200 & 2.037 & 1.650 & 1.479 \\
    & \tt 111 & 1.065 & 1.061 & 2.070 & 1.497 & 1.423 \\
    & \tt 1111 & 1.188 & 1.182 & 1.815 & 1.220 & 1.351 \\
    & \tt 1234 & 1.095 & 1.200 & 1.974 & 1.478 & 1.437 \\
    \cmidrule(lr){2-7}
    % & $\mu~(\sigma)$ & 1.09 (0.06) & 1.17 (0.06) & 1.93 (0.14) & 1.47 (0.16) & 1.41 (0.05) \\
    & $\mu$ & 1.09 & 1.17 & 1.93 & 1.47 & 1.41 \\
    & $\sigma$ & 0.06 & 0.06 & 0.14 & 0.16 & 0.05 \\
    \midrule
    \multirow{6}{*}{14}
    & \tt 10 & 1.539 & 1.418 & 3.022 & 1.916 & 1.974 \\
    & \tt 42 & 1.533 & 1.245 & 2.831 & 2.081 & 1.923 \\
    & \tt 111 & 1.672 & 1.623 & 2.905 & 2.257 & 2.114 \\
    & \tt 1111 & 1.590 & 1.492 & 2.937 & 2.049 & 2.017 \\
    & \tt 1234 & 1.668 & 1.545 & 2.928 & 1.727 & 1.967 \\
    \cmidrule(lr){2-7}
    % & $\mu~(\sigma)$ & 1.60 (0.07) & 1.46 (0.14) & 2.92 (0.07) & 2.01 (0.20) & 2.00 (0.07) \\
    & $\mu$ & 1.60 & 1.46 & 2.92 & 2.01 & 2.00 \\
    & $\sigma$ & 0.07 & 0.14 & 0.07 & 0.20 & 0.07 \\
    % \bottomrule 
    % \end{tabular}%
    % \hspace{0.36in}
    % \begin{tabular}{cr cc cc c}
    % \toprule
    % $H$ &\bf seed &
    % {\bf 2015/16} & 
    % {\bf 2016/17} & 
    % {\bf 2017/18} & 
    % {\bf 2018/19} & 
    % {\bf Average} \\
    \midrule
    \multirow{6}{*}{21}
    & \tt 10 & 1.813 & 1.524 & 3.999 & 2.569 & 2.476 \\
    & \tt 42 & 1.939 & 1.568 & 3.971 & 2.562 & 2.510 \\
    & \tt 111 & 1.945 & 1.329 & 3.907 & 2.541 & 2.431 \\
    & \tt 1111 & 1.966 & 1.487 & 4.001 & 2.469 & 2.481 \\
    & \tt 1234 & 2.107 & 1.320 & 4.006 & 2.542 & 2.494 \\
    \cmidrule(lr){2-7}
    % & $\mu~(\sigma)$ & 1.95 (0.10) & 1.45 (0.11) & 3.98 (0.04) & 2.54 (0.04) & 2.48 (0.03) \\
    & $\mu$ & 1.95 & 1.45 & 3.98 & 2.54 & 2.48 \\
    & $\sigma$ & 0.10 & 0.11 & 0.04 & 0.04 & 0.03 \\
    \midrule
    \multirow{6}{*}{28}
    & \tt 10 & 2.324 & 1.746 & 4.417 & 2.986 & 2.869 \\
    & \tt 42 & 2.013 & 1.749 & 4.460 & 2.770 & 2.748 \\
    & \tt 111 & 2.142 & 1.596 & 4.346 & 2.898 & 2.745 \\
    & \tt 1111 & 2.322 & 1.599 & 4.606 & 2.955 & 2.870 \\
    & \tt 1234 & 2.342 & 1.740 & 4.191 & 2.781 & 2.764 \\
    \cmidrule(lr){2-7}
    % & $\mu~(\sigma)$ & 2.23 (0.15) & 1.69 (0.08) & 4.40 (0.15) & 2.88 (0.10) & 2.80 (0.06) \\
    & $\mu$ & 2.23 & 1.69 & 4.40 & 2.88 & 2.80 \\
    & $\sigma$ & 0.15 & 0.08 & 0.15 & 0.10 & 0.06 \\
    \bottomrule 
    \end{tabular}%
    % }
\caption{Seed control ($5$ seeds) for Sonnet using the ILI-ENG data set across all forecasting horizons ($H$).  $\mu$ and $\sigma$ denote the mean and standard deviation of the $5$ obtained MAEs per test period. The results in the main paper were obtained for seed `\texttt{42}'.}
\label{tab:seed_control}
\end{table}

\subsection{Ablation study on Sonnet}
\label{subsec:ablation}
An ablation study is performed to better understand the contribution of Sonnet's components to the overall performance. We ablate over the following modules: the coherence mechanism (within MVCA), the MLP head for inter-variable dependency modelling (within MVCA), the complete MVCA module (section~\ref{subsec:MVCA}), the joint embedding module (section~\ref{subsec:joint_embedding}), and the Koopman evolvement module (section~\ref{subsec:koopman}). 

\begin{table*}[!b]
\renewcommand{\arraystretch}{0.92}
  \centering
  \small
  \setlength{\tabcolsep}{1.5pt}
  % \resizebox{\linewidth}{!}{%
  \begin{tabular}{lcccccccccccccccccc}
    \toprule
    \multirow{3}{*}{\textbf{$\alpha$}} & \multicolumn{9}{c}{\textbf{2020}} & \multicolumn{9}{c}{\textbf{2021}} \\
    \cmidrule(lr){2-10}\cmidrule(lr){11-19}
      & \multicolumn{3}{c}{\textbf{$H=12$}} & \multicolumn{3}{c}{\textbf{$H=24$}} & \multicolumn{3}{c}{\textbf{$H=36$}} & \multicolumn{3}{c}{\textbf{$H=12$}} & \multicolumn{3}{c}{\textbf{$H=24$}} & \multicolumn{3}{c}{\textbf{$H=36$}} \\
      & $r$ & MAE & $\epsilon\%$ & $r$ & MAE & $\epsilon\%$ & $r$ & MAE & $\epsilon\%$ & $r$ & MAE & $\epsilon\%$ & $r$ & MAE & $\epsilon\%$ & $r$ & MAE & $\epsilon\%$ \\
    \cmidrule(lr){2-4}\cmidrule(lr){5-7}\cmidrule(lr){8-10}\cmidrule(lr){11-13}\cmidrule(lr){14-16}\cmidrule(lr){17-19}
    0.75 & 0.9793 & 0.1085 & \underline{25.62} & 0.9748 & \underline{0.1214} & \underline{27.44} & 0.9614 & 0.1514 & 33.14 & 0.9780 & \underline{0.1073} & \underline{24.87} & 0.9733 & 0.1192 & 26.85 & 0.9634 & 0.1457 & \underline{30.39} \\
    0.5 & 0.9807 & \underline{0.1082} & 25.73 & 0.9747 & 0.1247 & 28.25 & 0.9671 & 0.1456 & \underline{31.41} & 0.9782 & 0.1108 & 25.82 & \underline{0.9747} & \underline{0.1185} & \underline{26.82} & 0.9640 & \underline{0.1421} & 30.88 \\
    0.25 & \underline{0.9813} & 0.1093 & 25.93 & \underline{0.9749} & 0.1250 & 29.17 & \underline{0.9680} & \underline{0.1430} & 31.69 & \underline{0.9788} & 0.1111 & 26.03 & 0.9735 & 0.1209 & \textbf{26.65} & \underline{0.9654} & 0.1434 & 30.56 \\
    0 & \textbf{0.9823} & \textbf{0.1036} & \textbf{25.30} & \textbf{0.9772} & \textbf{0.1179} & \textbf{26.96} & \textbf{0.9694} & \textbf{0.1401} & \textbf{31.25} & \textbf{0.9806} & \textbf{0.1042} & \textbf{24.56} & \textbf{0.9753} & \textbf{0.1170} & 27.09 & \textbf{0.9679} & \textbf{0.1308} & \textbf{28.91} \\
    \midrule
    Na\"ive & 0.9461 & 0.1675 & 34.30 & 0.9461 & 0.1675 & 34.30 & 0.9461 & 0.1675 & 34.30 & 0.9412 & 0.1736 & 35.14 & 0.9412 & 0.1736 & 35.14 & 0.9412 & 0.1736 & 35.14 \\
    \bottomrule
  \end{tabular}
  % }%
\caption{Effect of different $\alpha$ values with Sonnet on the ELEC data set. $r$ and $\epsilon\%$ denote the correlation and sMAPE scores, respectively. \bestResults}
\label{tab:sonnet-alpha-comparison}
\end{table*}

% seasonal persistence: 
% 2020: MAE: 0.1675  sMAPE: 34.30  Correlation: 0.9461
% 2021: MAE: 0.1736  sMAPE: 35.14 Correlation: 0.9412

A detailed description of each ablation variant is provided as follows:

\begin{itemize}[leftmargin=*, topsep=-2pt, itemsep=-1.6pt]
    \item \textbf{$\neg$ Coher}: This denotes that in MVCA, we do not compute the coherence attention. Instead of multiplying the attention with the value embedding ($\mathbf{O}_h = \mathbf{A} \odot \mathbf{V}_h$), the output embedding of each head $\mathbf{O}_h$ is equal to the value embedding, \ie $\mathbf{O}_h = \mathbf{V}_h$. In this case, the attention highlighting temporal importance is not included; only the MLP that captures the variable dependency is included.
    \item \textbf{$\neg$ MLP}: This denotes that we do not add the residual connection with the two-layer perceptron to obtain the attention output. Instead of computing $\mathbf{O}_m = \mathbf{O}_r + \mathrm{MLP}\left(\mathbf{O}_r\right)$, we directly have $\mathbf{O}_m = \mathbf{O}_r$. In this case, the inter-variable dependency via the MLP is omitted. Although some inter-variable interactions are partially (no learnable parameter is included in this process) captured when transforming the input into the frequency space along the feature dimension.
    \item \textbf{$\neg$ MVCA}: This denotes that we remove the MVCA from the modelling structure of Sonnet. After projecting the embedding into the wavelet space using the learnable wavelet transform, we directly feed the input into the Koopman evolvement layer.
    \item \textbf{$\neg$ Embed}: This denotes that instead of embedding the endo- and exogenous variables separately with $2$ different weight matrices, we use $1$ single weight matrix, $\mathbf{W} \in \mathbb{R}^{(C+1) \times d}$, over all input variables, $\mathbf{Z} \in \mathbb{R}^{L \times (C+1)}$, to obtain the input embedding $\mathbf{E} \in \mathbb{R}^{L \times d} = \mathbf{Z}\mathbf{W}$. 
    \item \textbf{$\neg$ Koop}: This denotes that we do not predict the next step of the temporal state using the Koopman operator. Instead, after capturing the dependencies using MVCA, the time series is directly reconstructed from the output of the attention module.
\end{itemize}

\begin{figure*}[!t]
    \centering
    \includegraphics[width=0.8\linewidth]{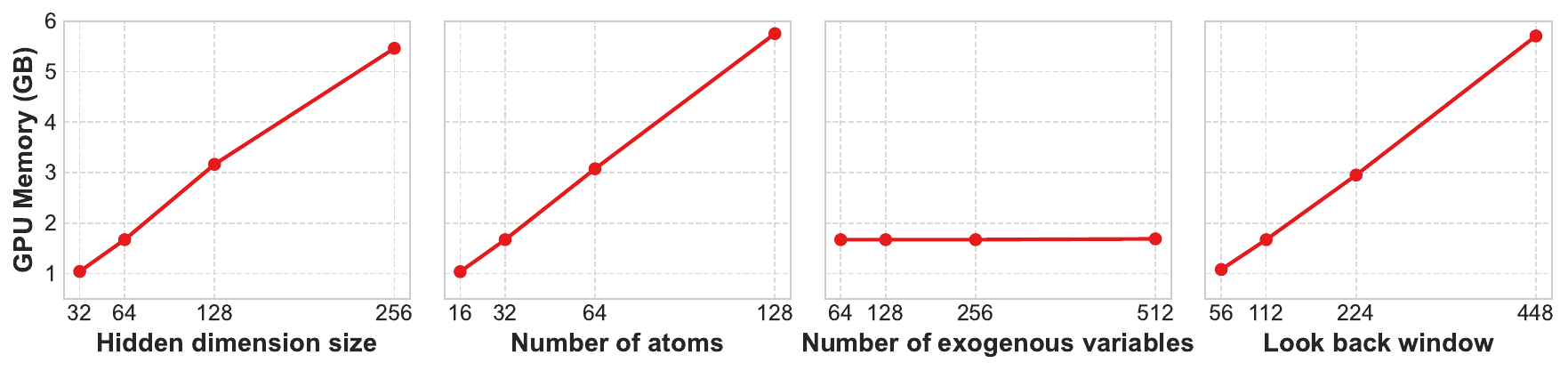}
    \caption{Computational cost (GPU memory) of Sonnet regarding the hidden dimension size $d$, the number of atoms $K$, the number of exogenous variables $C$, and the look back window length $L$.}
    \label{appfig:gpu}
\end{figure*}

We re-tune Sonnet's hyperparameters for each ablation variant across all locations, test seasons, and forecasting horizons (for the ILI and WEA tasks).

Results are enumerated in Table~\ref{tab:ablation}. We report the average MAE across all test seasons and its percentage of difference compared to the Sonnet model (enumerated in Table~\ref{tab:results_averaged}). Among all the ablated components, the MVCA module has the greatest impact. Removing MVCA results in an increase of MAE by $6.3\%$ on average across the ILI and WEA tasks. Breaking down MVCA further, we observe that removing its two internal components: coherency modelling and the MLP residual connection, leads to notable drops in performance ($5.2\%$ and $4.1\%$ increase in MAE, respectively), but less pronounced than removing the entire MVCA block. This indicates that while capturing coherence over time and enhancing variable interactions with residual connected MLP are both important modules, they also work in tandem within the full MVCA design.

Other components within Sonnet also contribute to the model's predictive ability, although to a lesser extent. Removing the joint embedding module leads to a $4\%$ increase in MAE, 
highlighting the importance of separately encoding the target and exogenous variables. Additionally, removing the Koopman operator module increases MAE by $3.2\%$, demonstrating the effectiveness of modelling the temporal evolution. Interestingly, as the forecasting horizon $H$ increases, the performance degradation becomes greater when removing coherence in MVCA, from an MAE increase of $2\%$ and $1.1\%$ (shortest $H$) for the ILI and WEA tasks, respectively, to $13.6\%$ and $3.7\%$ (longest $H$). This indicates that spectral coherence is more important for longer-term forecasting.

\subsection{Sensitivity to random seed initialisation}
\label{appsubsec:seed_control}

Results that assess the sensitivity of Sonnet to different random seed initialisations are enumerated in Table~\ref{tab:seed_control}. We have examined $5$ different seeds (`\texttt{10}', `\texttt{42}', `\texttt{111}', `\texttt{1111}', `\texttt{1234}') while conducting experiments on the ILI-ENG data set for all test periods and forecasting horizons. The obtained results highlight that the proposed model provides consistent results across all regions and forecasting horizons. Results do not deviate significantly from the ones presented in the main paper (Table~\ref{tab:results_averaged}) which are obtained using seed `\texttt{42}'. Standard deviations in the averaged MAEs (all test periods) over the $5$ seeds for each forecasting horizon do not exceed $0.07$ which in real terms (using the units of the actual data) represents $7$ people in a population of $10$ million. This highlights the robustness of the mean estimate.

\subsection{Analysis of model performance for tasks with strong seasonality}
\label{subsec:when_endo}

\begin{table*}[!t]
\renewcommand{\arraystretch}{0.92}
\centering\fontsize{9}{10}\selectfont
% \small
\setlength{\tabcolsep}{4pt}
\setlength{\aboverulesep}{0.5pt}
\setlength{\belowrulesep}{1.2pt}
% \resizebox{\linewidth}{!}{%
\begin{tabular}{ccccccccccc}
\toprule
\textbf{Task} & $H$ & \textbf{Sonnet} & \textbf{DeformTime} & \textbf{ModernTCN} & \textbf{Samformer} & \textbf{TimeXer} & \textbf{PatchTST} & \textbf{Itransformer} & \textbf{Crossformer} & \textbf{DLinear} \\
\midrule
     \multirow{3}{*}{\rotatebox{90}{ELEC}} & 12 & \textbf{0.0885} & \underline{0.1033} & 0.1440 & 0.1801 & 0.1114 & 0.1267 & 0.1261 & 0.1330 & 0.2630 \\
     & 24 & \textbf{0.1034} & \underline{0.1223} & 0.1649 & 0.2198 & 0.1423 & 0.1368 & 0.1394 & 0.1703 & 0.2997 \\
     & 36 & \textbf{0.1138} & 0.1507 & 0.1788 & 0.2441 & 0.1546 & \underline{0.1431} & 0.1554 & 0.1969 & 0.3259 \\
     \midrule
     \multirow{4}{*}{\rotatebox{90}{ENER}} & 24 & \underline{0.3152} & 0.3474 & \textbf{0.3133} & 0.3256 & 0.3178 & 0.3262 & 0.3220 & 0.3205 & 0.3691 \\
     & 48 & \textbf{0.3588} & 0.3962 & \underline{0.3692} & 0.3715 & 0.4186 & 0.3791 & 0.3808 & 0.4253 & 0.4246 \\
     & 72 & \textbf{0.3754} & 0.4071 & \underline{0.3792} & 0.3814 & 0.4227 & 0.3867 & 0.3973 & 0.4123 & 0.4458 \\
     & 168 & \textbf{0.3796} & 0.4620 & 0.4175 & \underline{0.3986} & 0.4196 & 0.4352 & 0.3991 & 0.4772 & 0.4604 \\
     \midrule
     \multirow{4}{*}{\rotatebox{90}{ETTh1}}
     & 96 & 0.1972 & 0.1840 & \underline{0.1774} & 0.1905 & 0.1807 & \bf 0.1757 & 0.1825 & 0.1989 & 0.2045 \\
     & 192 & 0.2187 & \bf 0.1988 & \underline{0.2013} & 0.2203 & 0.2043 & 0.1999 & 0.2051 & 0.2149 & 0.2834 \\
     & 336 & 0.2462 & \bf 0.2024 & \underline{0.2123} & 0.2433 & 0.2226 & 0.2248 & 0.2322 & 0.2577 & 0.3877 \\
     & 720 & 0.2661 & 0.2390 & \bf 0.2288 & 0.2648 & \underline{0.2292} & 0.2500 & 0.2449 & 0.3000 & 0.5007 \\
    \midrule
    \multirow{4}{*}{\rotatebox{90}{ETTh2}}
     & 96 & \bf 0.2743 & 0.2924 & 0.2790 & 0.2986 & 0.2797 & \underline{0.2765} & 0.2991 & 0.3254 & 0.2840 \\
     & 192 & 0.3185 & \bf 0.3010 & 0.3183 & 0.3254 & 0.3331 & \underline{0.3174} & 0.3463 & 0.3534 & 0.3313 \\
     & 336 & 0.3463 & \bf 0.3172 & \underline{0.3350} & 0.3622 & 0.3769 & 0.3417 & 0.3857 & 0.3891 & 0.4056 \\
     & 720 & \underline{0.3729} & \bf 0.3577 & 0.4259 & 0.4239 & 0.4069 & 0.3887 & 0.3945 & 0.4248 & 0.5391 \\
     \midrule
     \multirow{4}{*}{\rotatebox{90}{ILI-ENG}}
     & 7 & \underline{1.3204} & \bf 1.2802 & 1.7351 & 1.7701 & 1.7086 & 1.9402 & 1.7746 & 1.3243 & 2.1892 \\
     & 14 & \bf 1.5964 & \underline{1.6521} & 2.2656 & 2.1201 & 1.9703 & 2.4857 & 2.3660 & 1.9326 & 2.7452 \\
     & 21 & \bf 2.1226 & \underline{2.3246} & 2.3331 & 2.8860 & 2.8852 & 2.8458 & 3.0904 & 2.8733 & 3.2983 \\
     & 28 & \bf 2.4956 & 2.7829 & \underline{2.5948} & 3.3530 & 2.7894 & 3.2161 & 3.4784 & 3.0740 & 3.6805 \\
    \midrule
    \multirow{4}{*}{\rotatebox{90}{ILI-US2}}
     & 7 & \bf 0.3795 & \underline{0.4142} & 0.4523 & 0.5140 & 0.4931 & 0.6400 & 0.5455 & 0.4444 & 0.7214 \\
     & 14 & \bf 0.4186 & \underline{0.4544} & 0.5501 & 0.5784 & 0.5402 & 0.7302 & 0.6357 & 0.4695 & 0.7899 \\
     & 21 & \bf 0.4914 & \underline{0.5374} & 0.5576 & 0.6004 & 0.6190 & 0.7489 & 0.6554 & 0.5720 & 0.8642 \\
     & 28 & \underline{0.5740} & \bf 0.5554 & 0.5892 & 0.6561 & 0.6808 & 0.8450 & 0.7370 & 0.5842 & 0.8920 \\
    \midrule
    \multirow{4}{*}{\rotatebox{90}{ILI-US9}}
     & 7 & \underline{0.2727} & \bf 0.2614 & 0.2945 & 0.3396 & 0.3435 & 0.3882 & 0.3725 & 0.3222 & 0.4348 \\
     & 14 & \bf 0.2925 & \underline{0.3040} & 0.3398 & 0.3965 & 0.3715 & 0.4334 & 0.4049 & 0.3254 & 0.4780 \\
     & 21 & \bf 0.2955 & \underline{0.3187} & 0.3558 & 0.3811 & 0.4430 & 0.4664 & 0.4504 & 0.3285 & 0.5315 \\
     & 28 & \bf 0.3671 & 0.3913 & \underline{0.3786} & 0.4202 & 0.4774 & 0.5002 & 0.5535 & 0.4018 & 0.5622 \\
     \midrule
     \multirow{4}{*}{\rotatebox{90}{WEA-CT}} & 4 & \underline{1.1748} & 1.3217 & 1.3716 & 1.6459 & 1.9038 & 2.8087 & 1.5908 & \textbf{1.1470} & 2.2270 \\
     & 12 & \underline{2.4586} & 2.5082 & 2.6260 & 2.9155 & 2.8796 & 3.6686 & 2.8671 & \textbf{2.3716} & 3.3320 \\
     & 28 & \textbf{3.2135} & 3.3330 & 3.4816 & 3.4502 & 3.4469 & 3.6743 & 3.4602 & \underline{3.2349} & 3.6682 \\
     & 120 & \underline{3.6790} & 3.7547 & 3.9605 & 3.8332 & 3.7892 & 3.9108 & 3.8934 & \textbf{3.6567} & 3.9094 \\
     \midrule
     \multirow{4}{*}{\rotatebox{90}{WEA-HK}} & 4 & \underline{0.5456} & 0.6001 & 0.5729 & 0.6577 & 0.7984 & 0.9391 & 0.6449 & \textbf{0.5435} & 0.7752 \\
     & 12 & \textbf{0.8753} & 0.9935 & 0.9314 & 0.9964 & 1.0159 & 1.2814 & 1.0259 & \underline{0.8953} & 1.1804 \\
     & 28 & \textbf{1.1809} & 1.2712 & 1.2761 & 1.3163 & 1.2920 & 1.3973 & 1.3405 & \underline{1.2660} & 1.3926 \\
     & 120 & \textbf{1.5093} & 1.6400 & 1.5321 & 1.6370 & 1.6032 & 1.7110 & 1.5684 & \underline{1.5196} & 1.6268 \\
     \midrule
     \multirow{4}{*}{\rotatebox{90}{WEA-LD}} & 4 & \textbf{1.2448} & 1.4040 & 1.4175 & 1.5651 & 1.9055 & 2.2429 & 1.5485 & \underline{1.2493} & 1.8231 \\
     & 12 & \underline{2.2391} & 2.3942 & 2.3661 & 2.4950 & 2.5732 & 2.9561 & 2.5287 & \textbf{2.2368} & 2.7131 \\
     & 28 & \textbf{2.8627} & 3.0055 & 2.9744 & 3.0418 & 3.0354 & 3.2615 & 3.1209 & \underline{2.8913} & 3.1500 \\
     & 120 & \textbf{3.2539} & \underline{3.3166} & 3.6610 & 3.5329 & 3.5188 & 3.7507 & 3.6614 & 3.3583 & 3.6064 \\
     \midrule
     \multirow{4}{*}{\rotatebox{90}{WEA-NY}} & 4 & \textbf{0.9584} & 1.1266 & 1.0886 & 1.2500 & 1.5107 & 1.7682 & 1.2372 & \underline{0.9673} & 1.4818 \\
     & 12 & \textbf{1.7458} & 1.8459 & 1.8486 & 1.9756 & 2.1140 & 2.3498 & 1.9589 & \underline{1.7531} & 2.2809 \\
     & 28 & \textbf{2.2952} & 2.4040 & 2.4314 & 2.5138 & 2.5193 & 2.6537 & 2.5108 & \underline{2.3223} & 2.7166 \\
     & 120 & \textbf{2.6635} & 2.8076 & 2.9735 & 2.9731 & 2.9984 & 2.9692 & 2.7759 & \underline{2.7372} & 3.1808 \\
     \midrule
     \multirow{4}{*}{\rotatebox{90}{WEA-SG}} & 4 & \textbf{0.3109} & 0.3229 & 0.3243 & 0.3622 & 0.3596 & 0.4010 & 0.3542 & \underline{0.3132} & 0.3809 \\
     & 12 & \textbf{0.3664} & 0.3849 & 0.3899 & 0.4189 & 0.4020 & 0.4487 & 0.4099 & \underline{0.3765} & 0.4285 \\
     & 28 & \textbf{0.4213} & 0.4540 & 0.4565 & 0.4713 & 0.4475 & 0.4866 & 0.4655 & \underline{0.4414} & 0.4719 \\
     & 120 & \textbf{0.4747} & 0.5128 & 0.5101 & 0.5018 & 0.5023 & 0.5191 & \underline{0.4936} & 0.4980 & 0.5151 \\
     
\bottomrule
\end{tabular}
% }
\caption{Mean Absolute Error (MAE) evaluated over the full predicted sequence for all forecasting horizons ($H$), tasks, and models. In consistency with the main experimental setup, error metrics are averaged over $4$ test seasons for ILI tasks, $3$ years for the WEA tasks, and $2$ for ELEC. \bestResults}
\label{tab:over_sequence}
\end{table*}

Prior work has shown that modelling inter-variable dependencies in time series problems may occasionally cause overfitting~\citep{nie2023time}. While our results for the ILI and WEA forecasting tasks do not corroborate this (Section~\ref{subsec:forecasting_results_Sonnet}), we conduct further experiments on a different task (ELEC) that displays a different pattern of performance across most models. In these experiments, we explore different values for the parameter $\alpha$ in Sonnet. $\alpha$ controls the relative contribution of exogenous variables in the learned representations (section~\ref{subsec:joint_embedding}). For $\alpha=0$, Sonnet only uses historical values of the target (endogenous) variable. We use the ELEC data set for evaluation because for this specific task, PatchTST---that does not capture inter-variable dependencies---outperforms some models that do (\eg ModernTCN and Crossformer). In addition, $7$ out of $9$ evaluated models cannot surpass the accuracy of the seasonal persistence model (both in terms of MAE and sMAPE), indicating that for ELEC, seasonal patterns are consistent over time. Sonnet's hyperparameters are re-tuned (see Section~\ref{subsec:forecasting_results_Sonnet} and Appendix~\ref{appsec:supplementary_setting}), except for $\alpha$. Results for $\alpha=\{0,0.25,0.5,0.75\}$ are enumerated in ~\ref{tab:sonnet-alpha-comparison}. The last row contains the performance of the seasonal persistence model (denoted as ``Na\"ive'') with a seasonality set to $168$ time steps ($1$ week).

We observe that although the prediction difference across different values of $\alpha$ is relatively small, Sonnet is consistently more accurate when setting $\alpha = 0$, \ie when only historical values of the target variable are used and no exogenous variables are included. This indicates that for ELEC the exogenous inputs are not very informative. The performance of the seasonal persistence model also supports this conclusion given its strong performance for both years 2020 and 2021 ($r > 0.94$). This suggests that when strong seasonality is present, Sonnet benefits from following an autoregressive forecasting design without incorporating inter-variable dependencies (as $\alpha$ is a learnable parameter).

\subsection{Computational cost of Sonnet}
\label{appsubsec:cost}

We analyse Sonnet's computational cost \wrt $4$ tunable hyperparameters: the number of atoms $K$, the size of the hidden dimension $d$, the length of the look-back window $H$, and the number of exogenous variables $C$. We analyse the influence of each variable independently by keeping the other three parameters fixed. Specifically, we set $d\in\{32,64,128,256\}$ while having $K=32$, $L=112$, $C=128$; similarly, we evaluate $K\in\{16,32,64,128\}$, $L\in\{56,84,112,224\}$, and $C\in\{64,128,256,512\}$, while the other parameters are fixed to $d=64$, $K=32$, $L=112$, and $C=128$, respectively. The computational cost of each setup is obtained by averaging memory consumption over $100$ epochs.

The GPU memory consumption of Sonnet during training is depicted in Figure~\ref{appfig:gpu}. Sonnet's GPU memory consumption stays constant as $C$ increases. This is because the input is mapped to a fixed embedding space for modelling. For the other three parameters, the method shows a linear increase in GPU memory consumption.

\subsection{Note on performance evaluation}
\label{appsubsec:performance_evalution}

We acknowledge that most prior work in time series forecasting evaluates model performance by averaging errors across the entire prediction horizon. While this provides a general view of forecasting accuracy, it may obscure performance at the target time step.
As the prediction horizon extends, forecasting becomes more challenging~\citep{jhyndman2006another}. Model performance can thus vary significantly between forecasting the closer / further time steps. 
For example, consider two models whose predictions over a three-step sequence yield MAEs of $\{0.1, 0.2, 0.6\}$ and $\{0.2, 0.3, 0.4\}$, respectively. While both models have the same averaged MAE when evaluated over the sequence, the second model performs better on the final time step, which is the main forecasting target.
Averaging errors over all time steps may benefit models that excel in early predictions but struggle near the target horizon. We further demonstrate this point in the coming section with empirical results.

\subsection{Evaluating forecasts over the entire output sequence}
\label{appsubsec:over_sequence}

To provide a more comprehensive understanding of model performance, we also provide the forecasting accuracy by averaging the error across the entire output sequence, as opposed to just using the target time step (Table~\ref{tab:results_averaged}). The results are presented in Table~\ref{tab:over_sequence}. It can be observed that Sonnet ranks first in $31$ over $47$ tasks and second in $9$ tasks. The averaged MAE reduction is $1.0\%$. Additionally, Sonnet reduces the MAE of DeformTime, which is the best-performing baseline, by $4.7\%$.

For the ILI and WEA data sets, the best-performing baseline is DeformTime when evaluated at the target time step, with Sonnet reducing the MAE by $3.3\%$ on average. However, Crossformer is the best performing model when evaluated over the sequence, with Sonnet reducing the MAE by $5.4\%$. This observation indicates that DeformTime performs better than Crossformer at the target horizon, while Crossformer yields better performance when considering the entire series of forecasts. This change in the ranking of baselines supports our argument in Section~\ref{appsubsec:performance_evalution} that averaging errors across the sequence, rather than evaluating the error at the target forecasting horizon, can lead to a misleading assessment of how the model performs at different forecasting horizons.

\section{Limitations}
\label{appsec:limitations}
The main limitation is that the validity and predictiveness of the proposed method are only verified empirically (through experiments) but not theoretically. To mitigate against this, we have conducted extensive experiments over a wide range of data sets and tasks, covering different application fields. Many ($8$) of the data sets involve multiple test seasons (hence, multiple training / testing trials can be conducted), and have training sets that encompass time spans longer than a decade. Nevertheless, we acknowledge that additional evaluation across a broader range of forecasting problems could have provided an even more comprehensive assessment. 

As far as the MVCA module is concerned (a key contribution of Sonnet), we argue that it is a direct and effective replacement for na\"ive attention in time series forecasting models. Although our current experiments support this claim, we nevertheless acknowledge that further analysis is needed to better understand its impact across a wider range of applications. Such insights could help guide more informed replacements of attention mechanisms in future work.

Finally, Sonnet---and many other SOTA forecasting models---cannot be considered as an interpretable model. We cannot know why Sonnet makes a certain forecast and we cannot directly attribute this to specific parts of the endogenous input. This can only be approximated through meta-analysis.

\section{Supplementary ILI rate forecasting figures}
\label{appsec:ILI_forecasting_figures}

In this section, a series of figures is shown below for the ILI rate prediction task, where each figure corresponds to a particular forecasting window and location, showcasing outputs from all models in parallel for comparison.

\begin{figure*}[!t]
    \centering
    \begin{minipage}{0.45\linewidth}
        \centering
        \includegraphics[width=\linewidth]{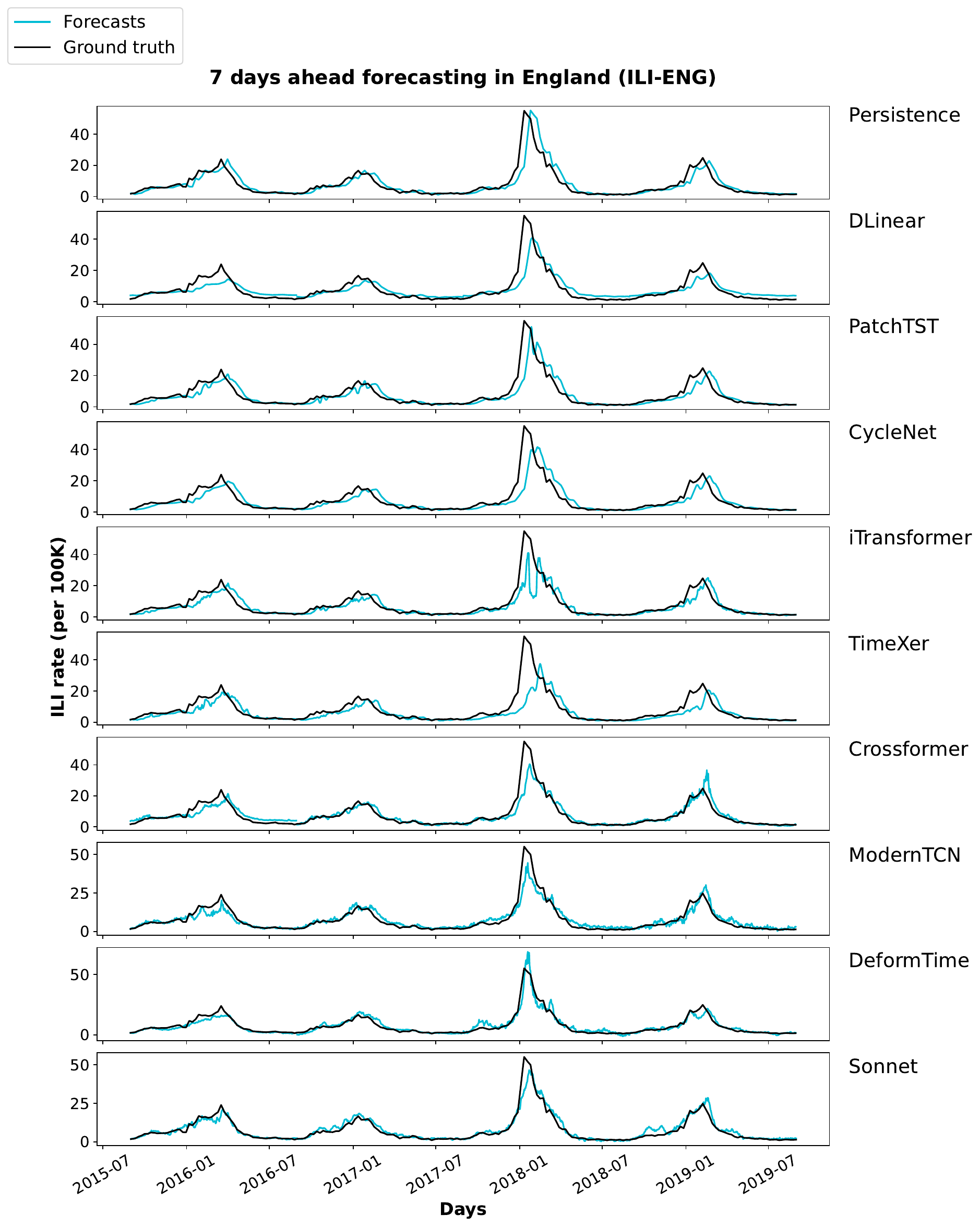}
        \caption{\ILIFigCaptionAppendix{7}{England}{ILI-ENG}}
        \label{fig:uk_7_days_forecasting}
    \end{minipage}%
    \hfill
    \begin{minipage}{0.45\linewidth}
        \centering
        \includegraphics[width=\linewidth]{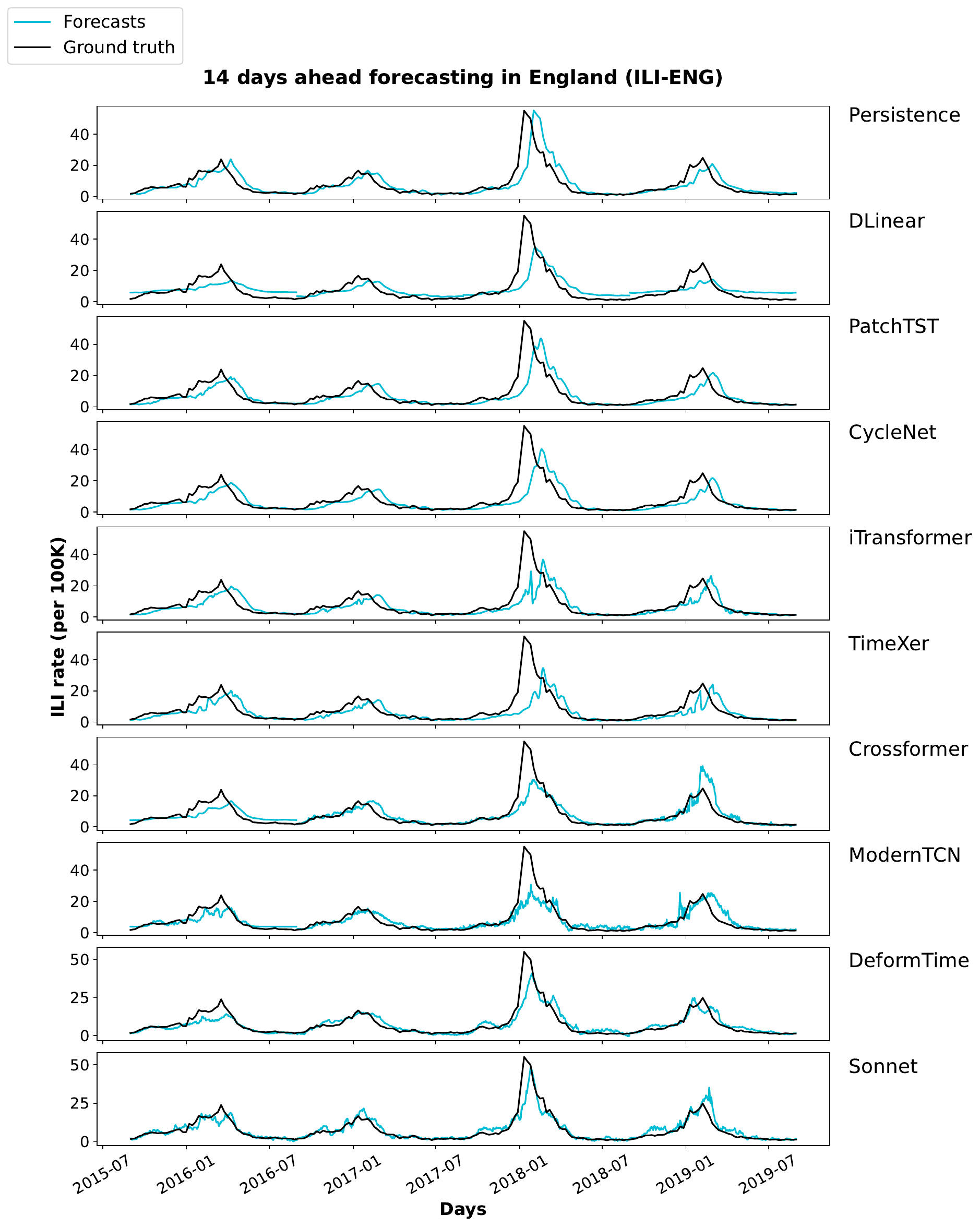}
        \caption{\ILIFigCaptionAppendix{14}{England}{ILI-ENG}}
        \label{fig:uk_14_days_forecasting}
    \end{minipage}
\end{figure*}

\begin{figure*}[!t]
    \centering
    \begin{minipage}{0.45\linewidth}
        \centering
        \includegraphics[width=\linewidth]{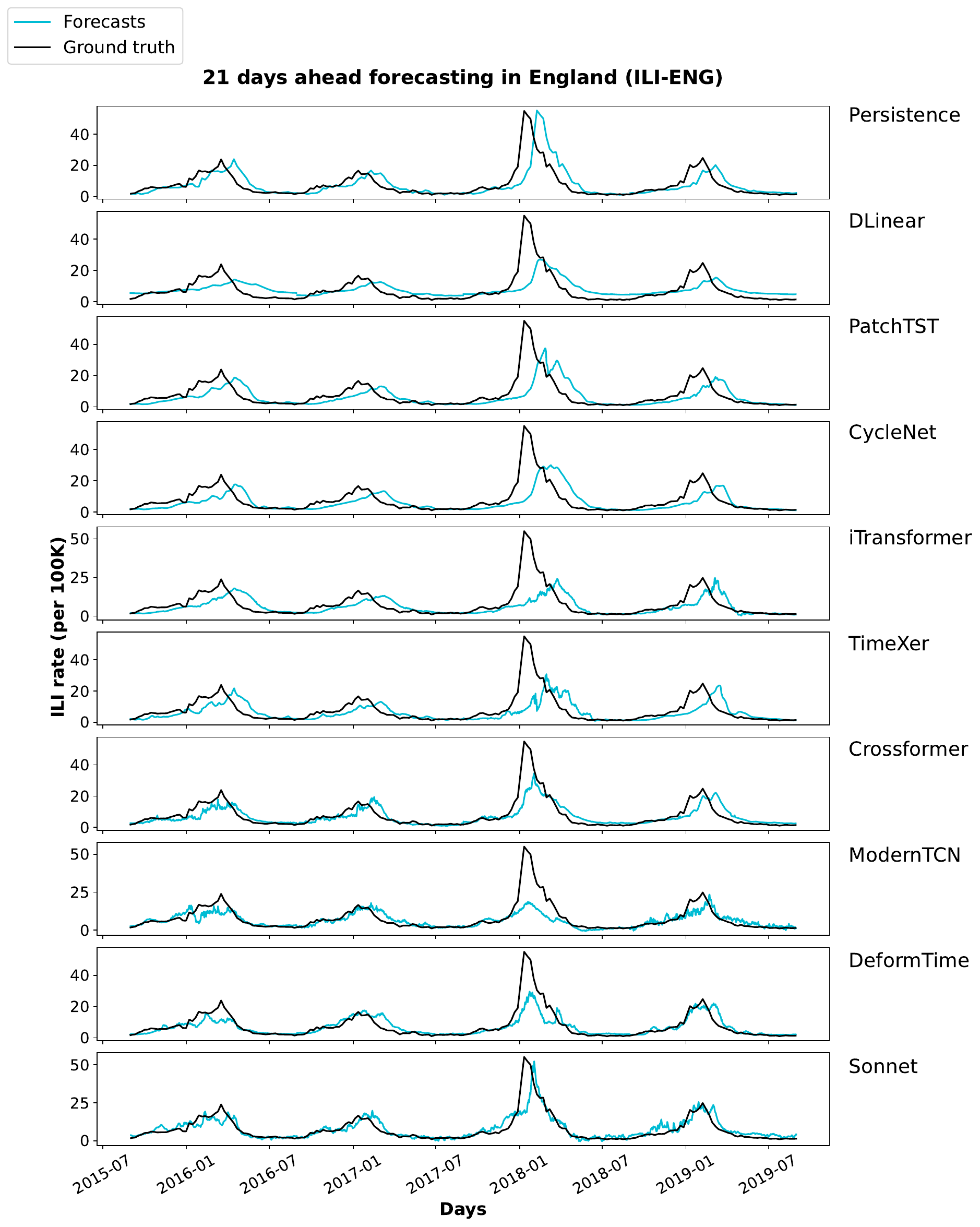}
        \caption{\ILIFigCaptionAppendix{21}{England}{ILI-ENG}}
        \label{fig:uk_21_days_forecasting}
    \end{minipage}%
    \hfill
    \begin{minipage}{0.45\linewidth}
        \centering
        \includegraphics[width=\linewidth]{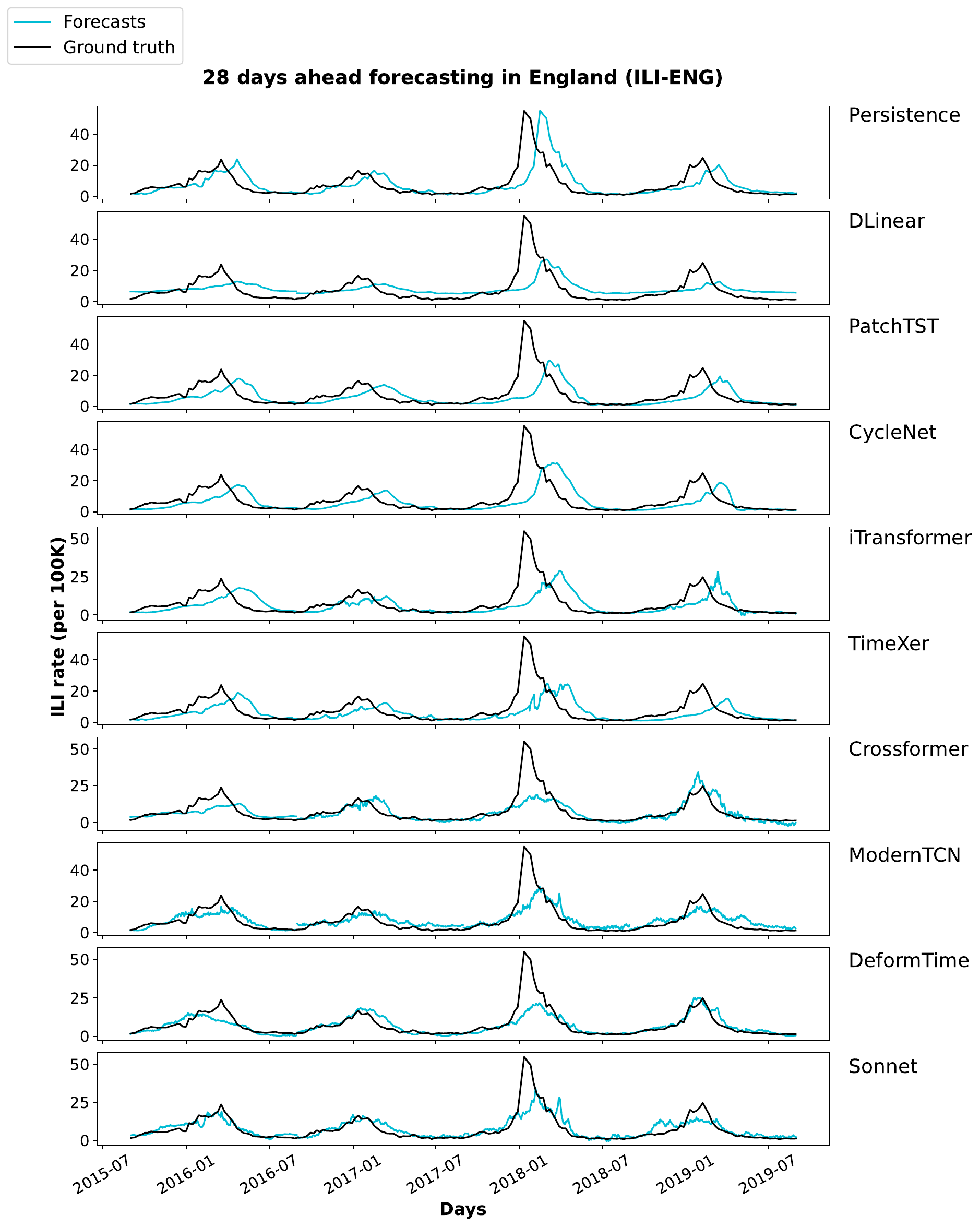}
        \caption{\ILIFigCaptionAppendix{28}{England}{ILI-ENG}}
        \label{fig:uk_28_days_forecasting}
    \end{minipage}
\end{figure*}

\begin{figure*}[!t]
    \centering
    \begin{minipage}{0.45\linewidth}
        \centering
        \includegraphics[width=\linewidth]{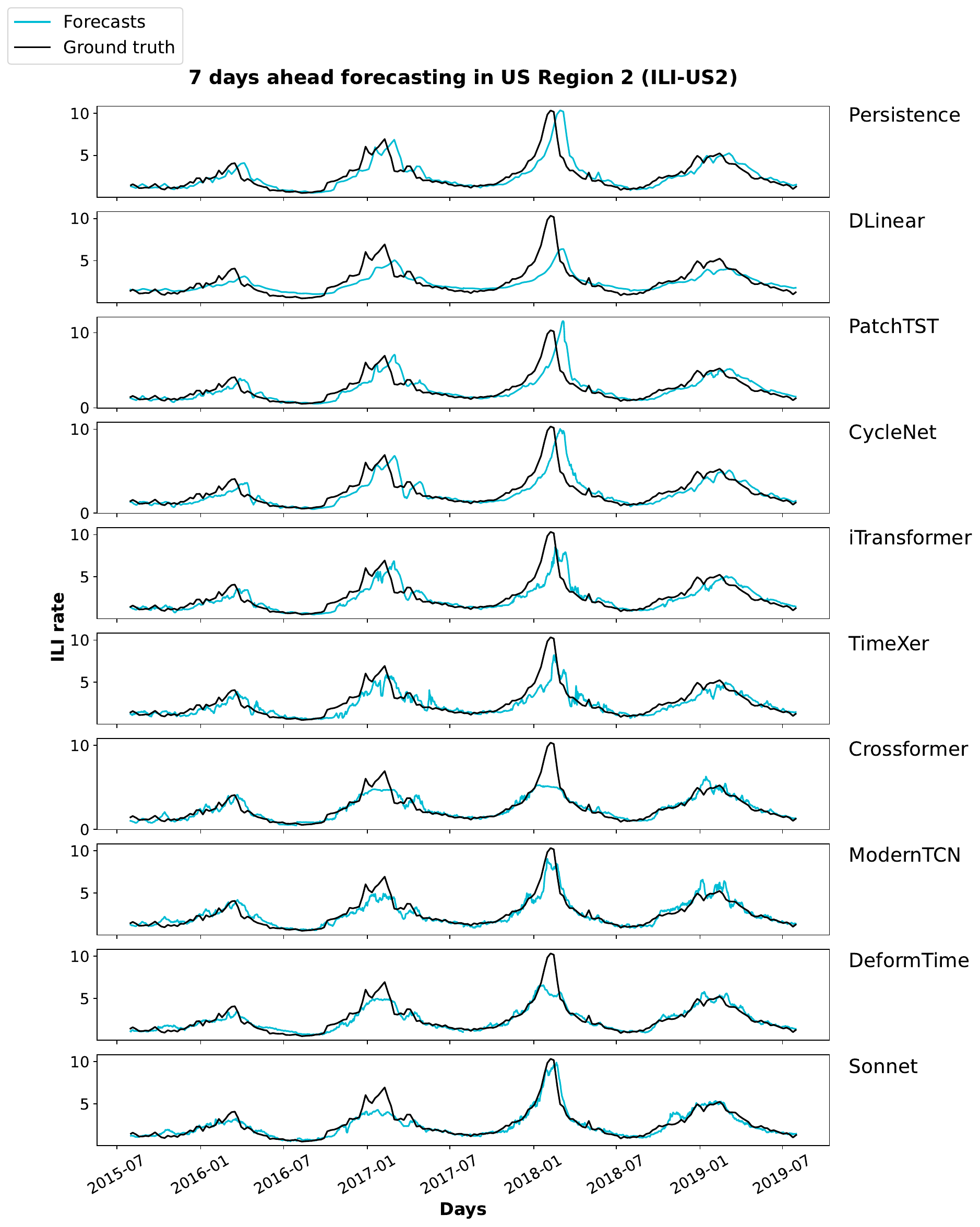}
        \caption{\ILIFigCaptionAppendix{7}{US Region 2}{ILI-US2}}
        \label{fig:us2_7_days_forecasting}
    \end{minipage}%
    \hfill
    \begin{minipage}{0.45\linewidth}
        \centering
        \includegraphics[width=\linewidth]{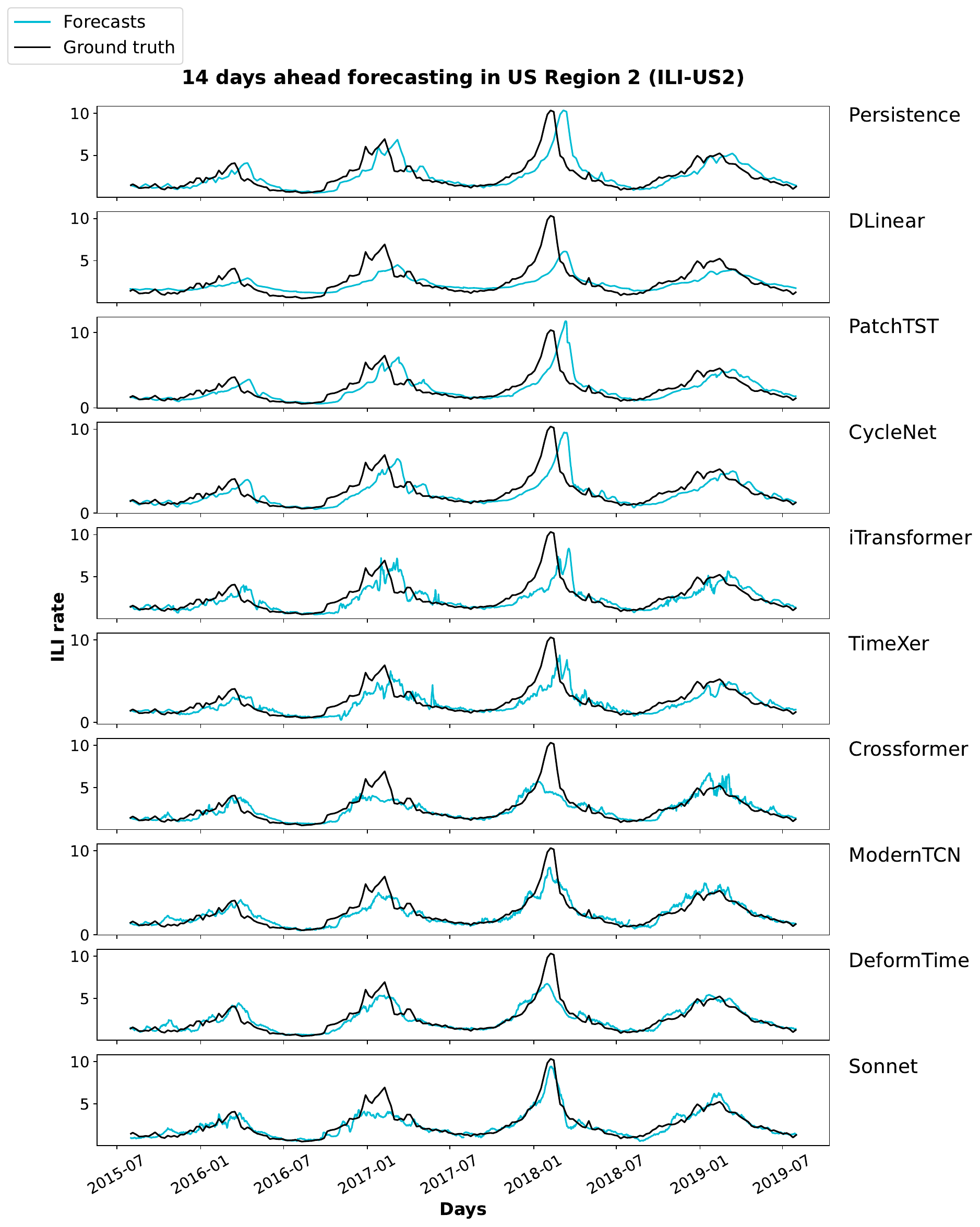}
        \caption{\ILIFigCaptionAppendix{14}{US Region 2}{ILI-US2}}
        \label{fig:us2_14_days_forecasting}
    \end{minipage}
\end{figure*}

\begin{figure*}[!t]
    \centering
    \begin{minipage}{0.45\linewidth}
        \centering
        \includegraphics[width=\linewidth]{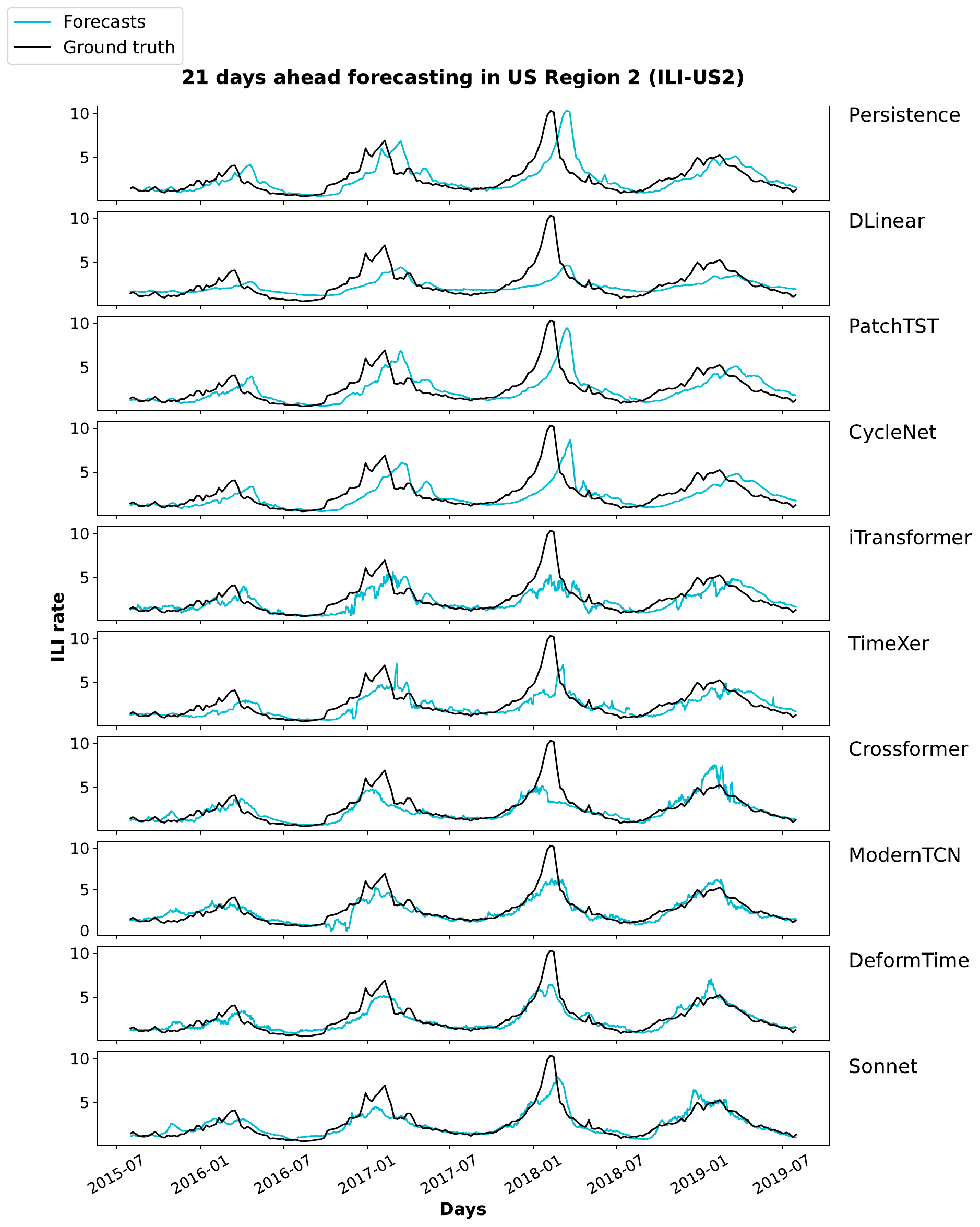}
        \caption{\ILIFigCaptionAppendix{21}{US Region 2}{ILI-US2}}
        \label{fig:us2_21_days_forecasting}
    \end{minipage}%
    \hfill
    \begin{minipage}{0.45\linewidth}
        \centering
        \includegraphics[width=\linewidth]{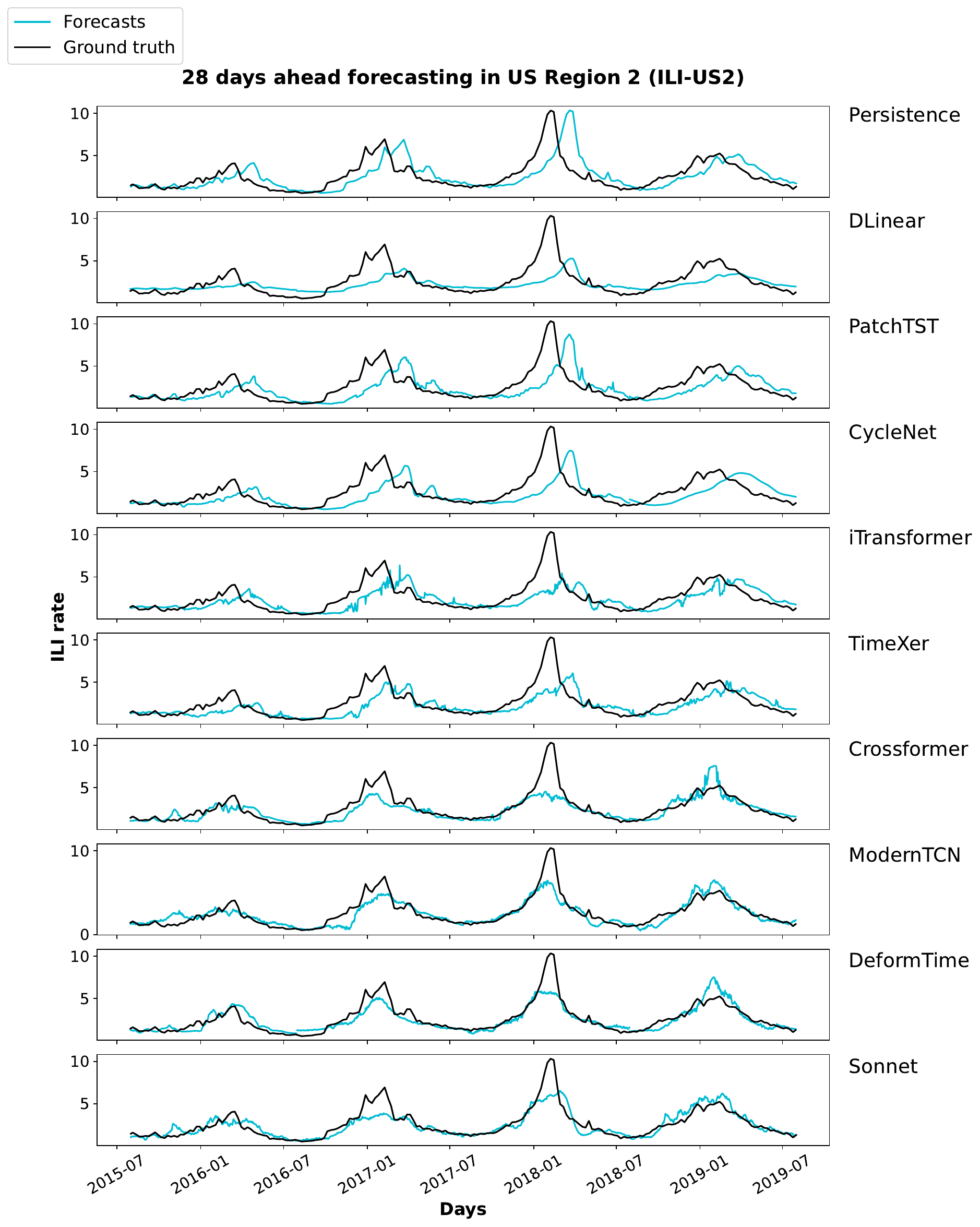}
        \caption{\ILIFigCaptionAppendix{28}{US Region 2}{ILI-US2}}
        \label{fig:us2_28_days_forecasting}
    \end{minipage}
\end{figure*}

\begin{figure*}[!t]
    \centering
    \begin{minipage}{0.45\linewidth}
        \centering
        \includegraphics[width=\linewidth]{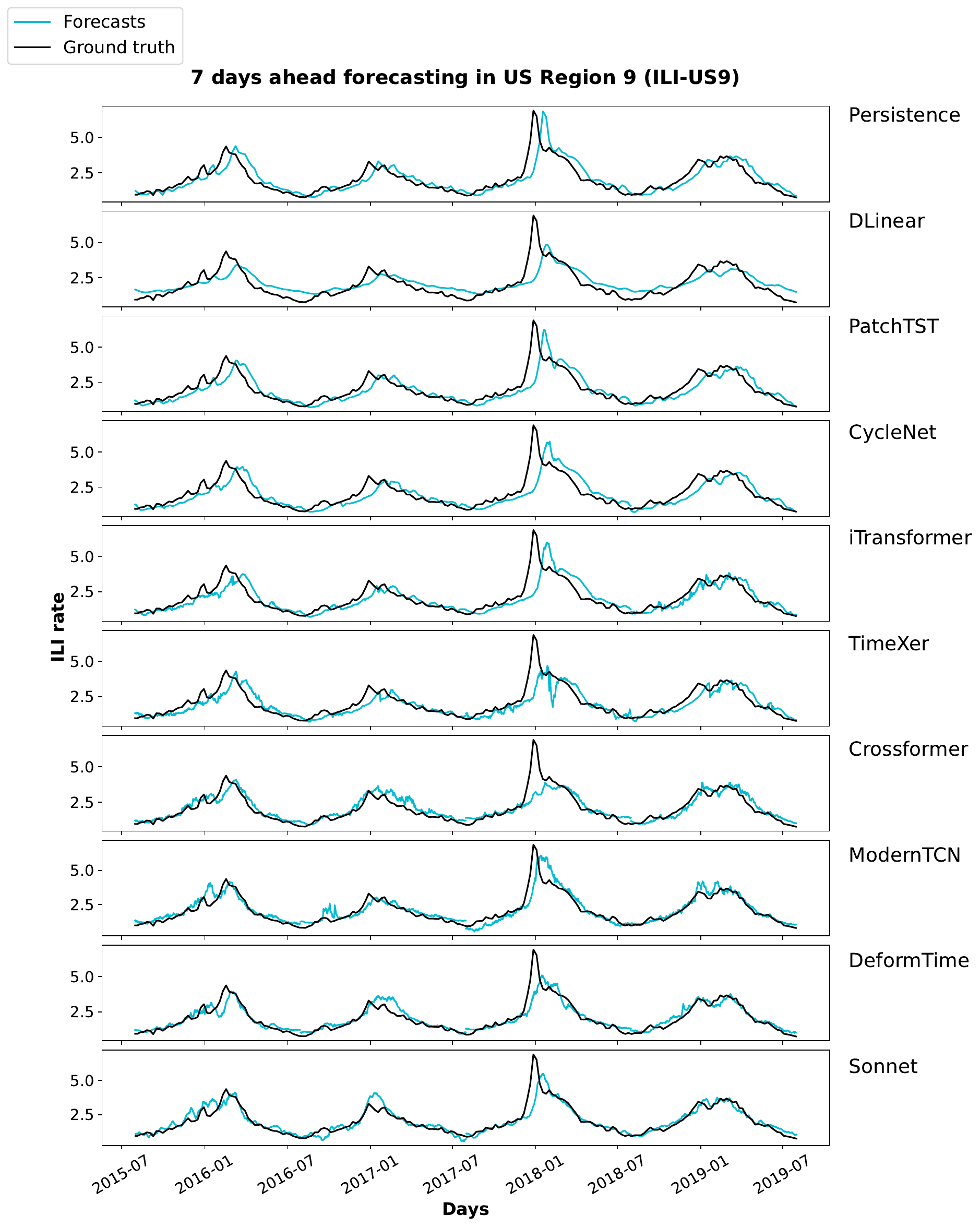}
        \caption{\ILIFigCaptionAppendix{7}{US Region 9}{ILI-US9}}
        \label{fig:us9_7_days_forecasting}
    \end{minipage}%
    \hfill
    \begin{minipage}{0.45\linewidth}
        \centering
        \includegraphics[width=\linewidth]{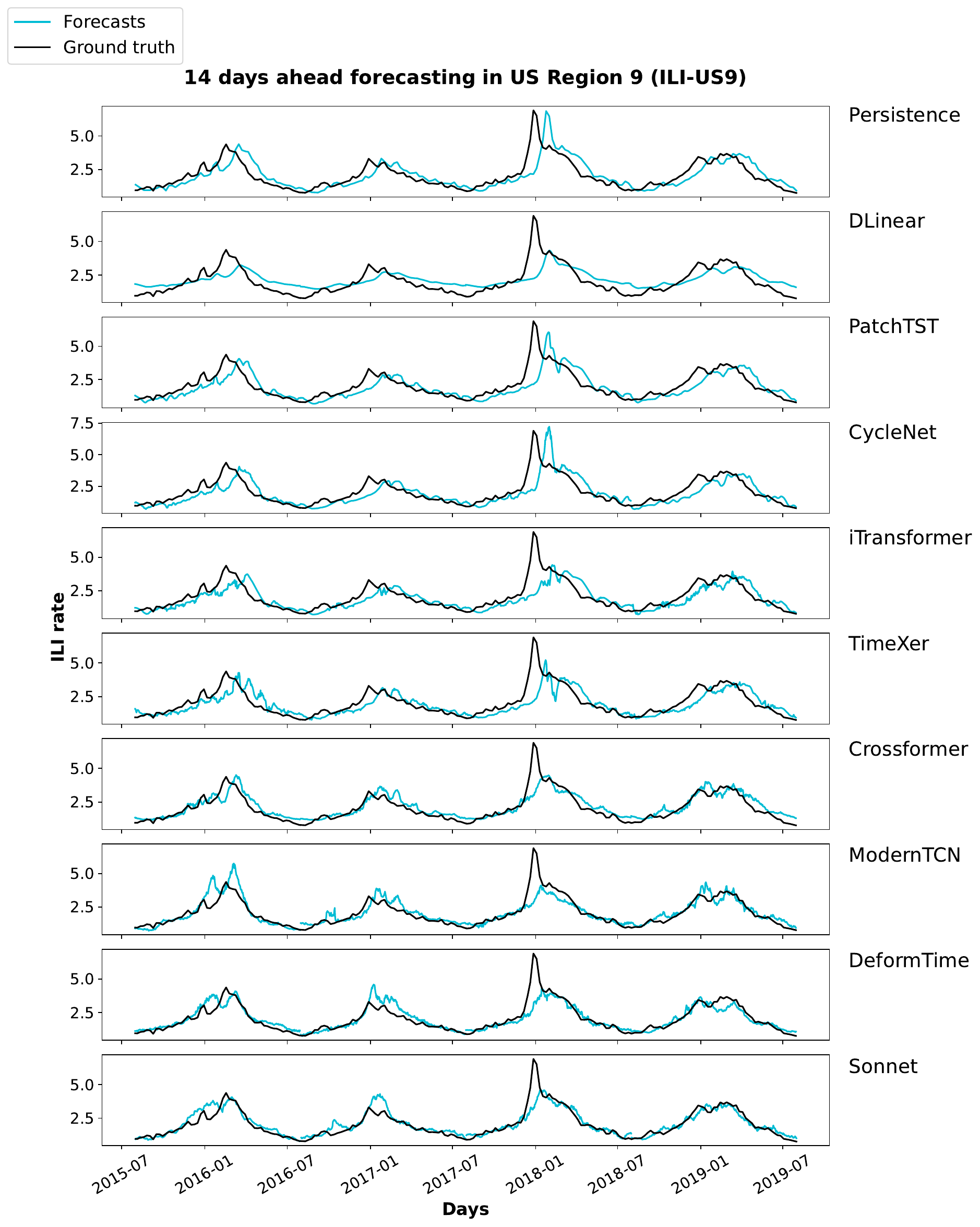}
        \caption{\ILIFigCaptionAppendix{14}{US Region 9}{ILI-US9}}
        \label{fig:us9_14_days_forecasting}
    \end{minipage}
\end{figure*}

\begin{figure*}[!t]
    \centering
    \begin{minipage}{0.45\linewidth}
        \centering
        \includegraphics[width=\linewidth]{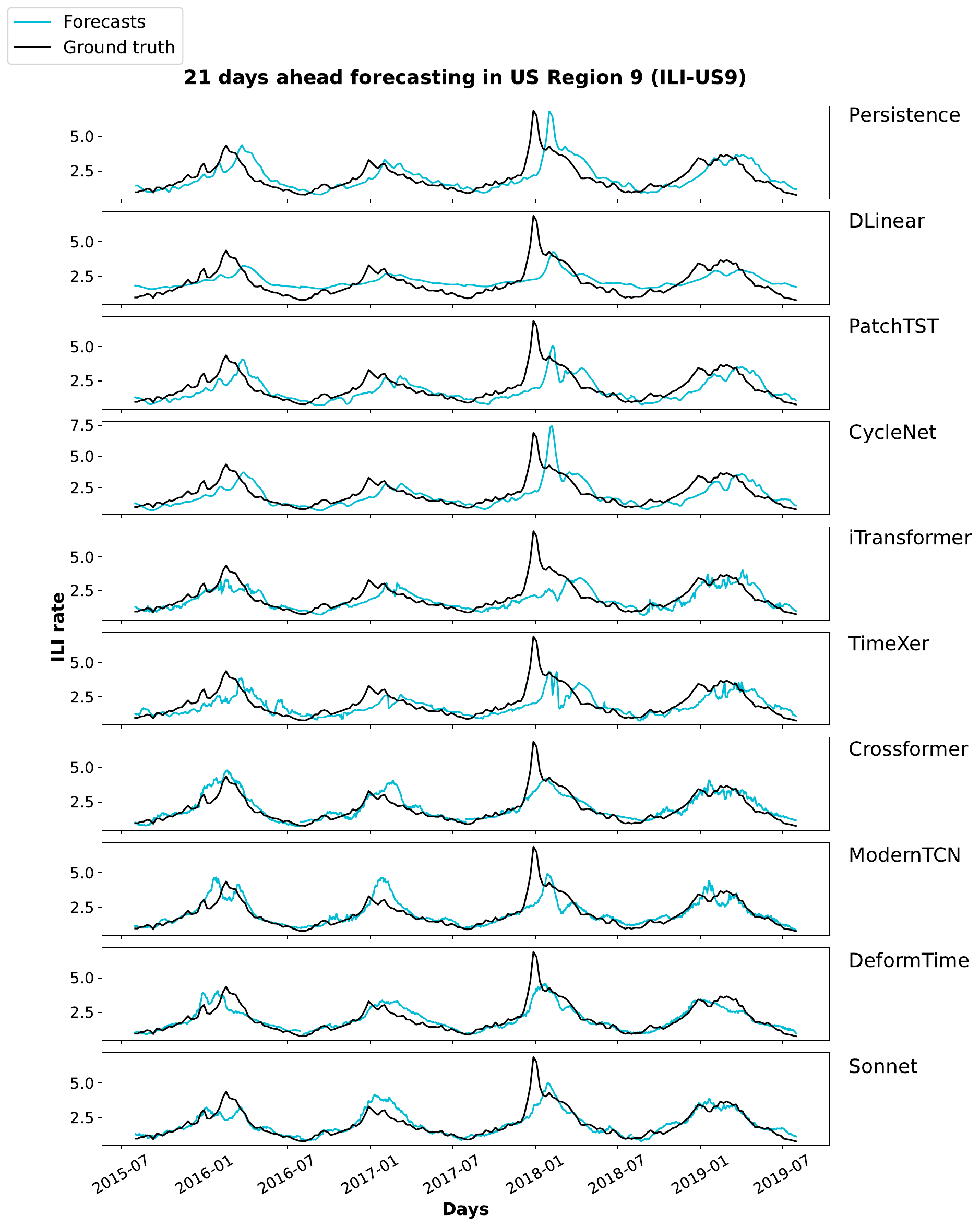}
        \caption{\ILIFigCaptionAppendix{21}{US Region 9}{ILI-US9}}
        \label{fig:us9_21_days_forecasting}
    \end{minipage}%
    \hfill
    \begin{minipage}{0.45\linewidth}
        \centering
        \includegraphics[width=\linewidth]{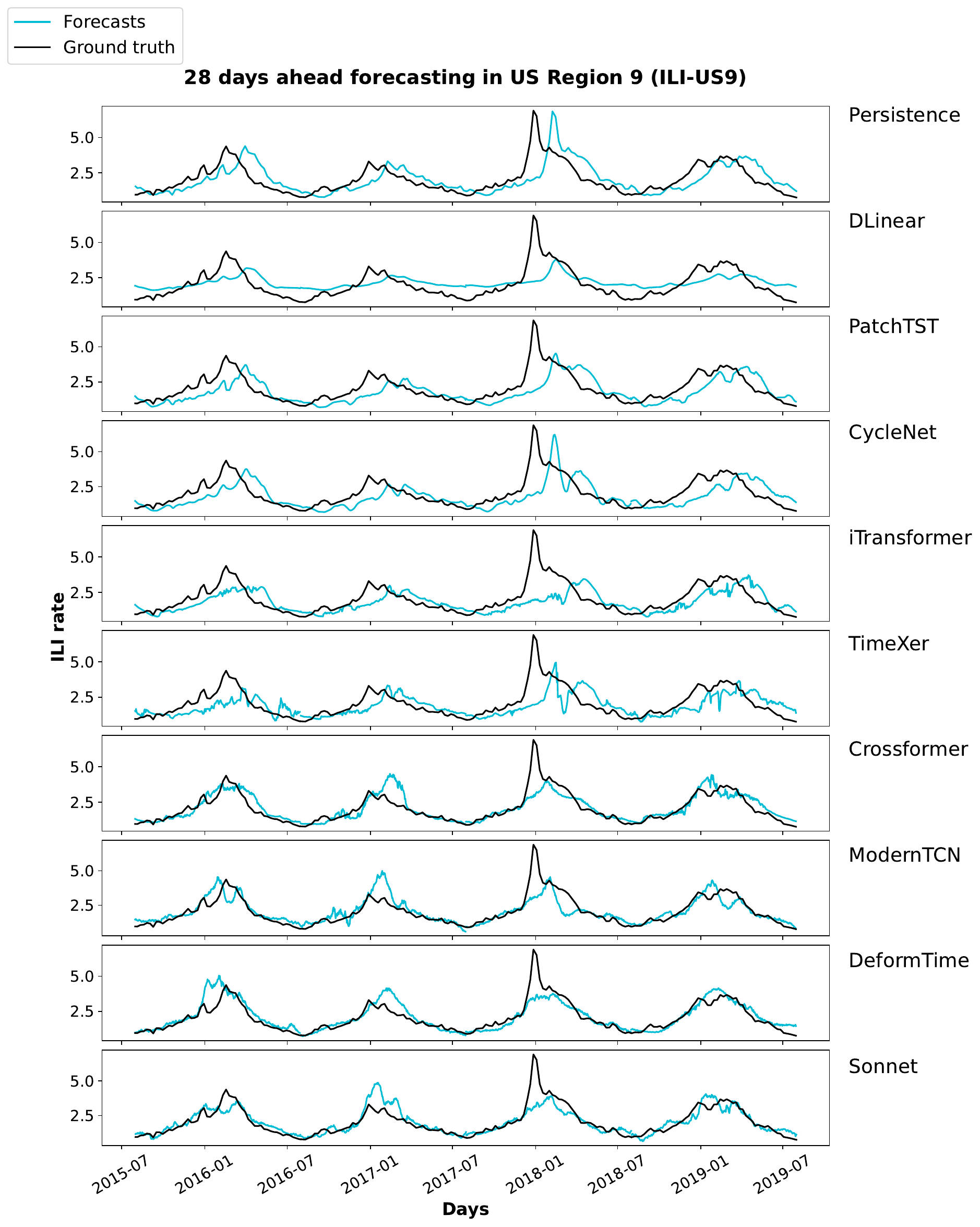}
        \caption{\ILIFigCaptionAppendix{28}{US Region 9}{ILI-US9}}
        \label{fig:us9_28_days_forecasting}
    \end{minipage}
\end{figure*}

\end{document}